\documentclass[10pt,twocolumn,letterpaper]{article}

\usepackage{arxiv}
\usepackage{times}
\usepackage{epsfig}
\usepackage{xcolor}
\usepackage{graphicx}
\usepackage{subfigure}
\usepackage{booktabs} 
\usepackage{amsmath}
\usepackage{amssymb}
\usepackage{dsfont}
\usepackage{algorithm}
\usepackage{algorithmic}

\usepackage{arydshln}
\usepackage{multirow}
\usepackage{subfigure}
\usepackage{enumitem}
\DeclareMathOperator*{\argmax}{arg\,max}

\usepackage{amsthm}
\usepackage{enumitem}
\usepackage[symbol]{footmisc}

\theoremstyle{definition}
\newtheorem{definition}{Definition}[section]

\usepackage[pagebackref=true,breaklinks=true,letterpaper=true,colorlinks,bookmarks=false]{hyperref}

\arxivfinalcopy 

\def\arxivPaperID{1758} 

\ifarxivfinal\fi

\begin{document}

\title{Prototype Guided Federated Learning of Visual Feature Representations}

\author{Umberto Michieli\textsuperscript{1,2}\thanks{Researched during internship at Samsung Research UK.} \quad \quad \quad Mete Ozay\textsuperscript{1}\\
\textsuperscript{1}Samsung Research UK \quad  \textsuperscript{2}University of Padova\\
{\tt\small \{u.michieli, m.ozay\}@samsung.com}
}

\maketitle
\ifarxivfinal\thispagestyle{empty}\fi

\begin{abstract}
Federated Learning (FL) is a framework which enables distributed model training using a large corpus of decentralized training data. Existing methods aggregate models disregarding their internal representations, which are crucial for training models in vision tasks. System and statistical heterogeneity (e.g., highly imbalanced and non-i.i.d.\ data) further harm model training. To this end, we introduce a method, called FedProto, which computes client deviations using margins of prototypical representations learned on distributed data, and applies them to drive federated optimization via an attention mechanism. In addition, we propose three methods to analyse statistical properties of feature representations learned in FL, in order to elucidate the relationship between accuracy, margins and feature discrepancy of FL models. In experimental analyses, FedProto demonstrates state-of-the-art accuracy and convergence rate across image classification and semantic segmentation benchmarks by enabling maximum margin training of FL models. Moreover, FedProto reduces uncertainty of predictions of FL models compared to the baseline. To our knowledge, this is the first work evaluating FL models in dense prediction tasks, such as semantic segmentation.
\end{abstract}


\section{Introduction}
\label{sec:introduction}

Federated Learning (FL) is a framework proposed for distributing training of machine learning models in a network of clients (devices) with local data processed only at clients \cite{bonawitz2019towards,kairouz2021advances,li2020federated_b,yang2019federated}. 
In FL, models are trained across multiple rounds. At the beginning of each round, every participating client receives an initial model from a central server, optimizes the model on its local training data and sends  the updated model back to the server. The server then aggregates all the models and updates the aggregate model \cite{mcmahan2017communication}. 
%

\begin{figure}
\centering
	\includegraphics[trim=0cm 12.3cm 16cm 0cm, clip, width=\linewidth]{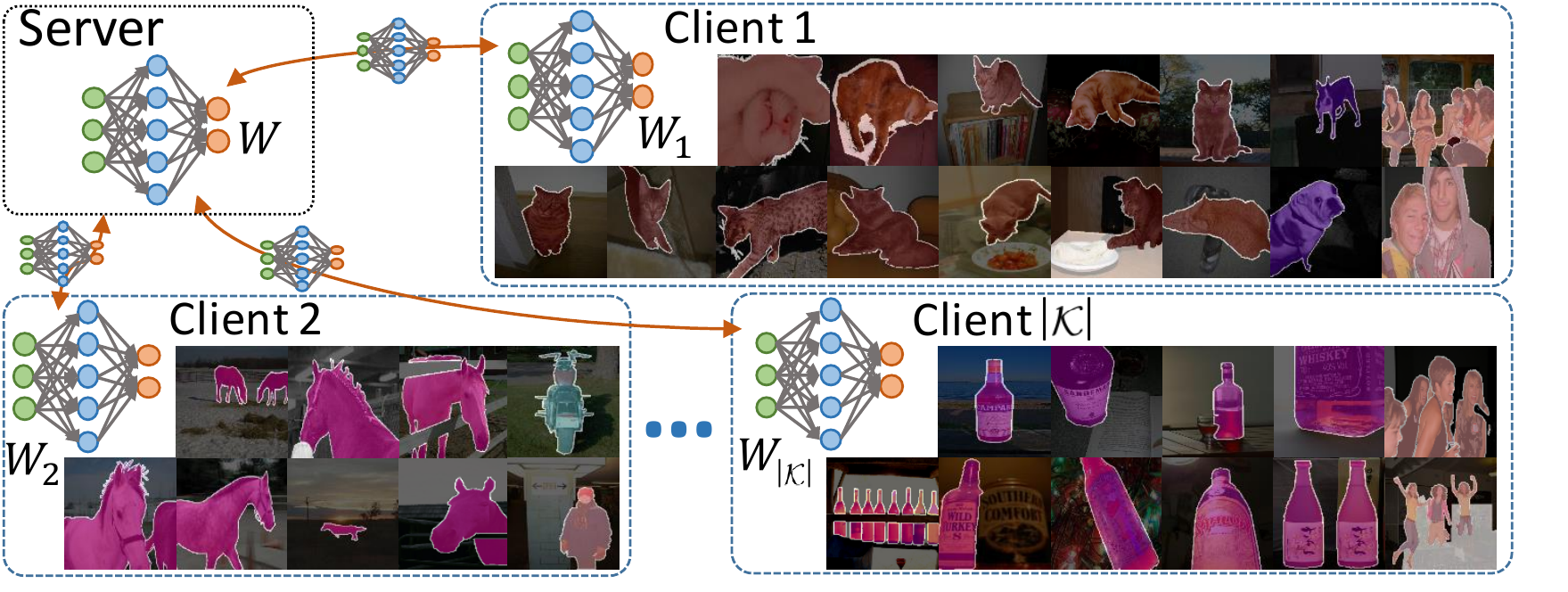}
	\label{fig:graphical_abstract_datadistribution}
	\vspace{-0.3cm}
	\caption{Visual data observed at distributed clients $k \in \mathcal{K}$ are non-i.i.d.\ and imbalanced. This represents a challenge for federated learning of vision models with parameters $W_k, \forall k$.}
	\vspace{-0.35cm}
\end{figure}

\textbf{Challenges of FL:}  Training models in FL systems introduces several novel challenges \cite{kairouz2021advances,li2020federated_b}. In this work, we  address problems caused by system and statistical heterogeneity. System heterogeneity refers to variable computational (\eg, CPU, memory, battery level) and communication (\eg, wifi) capabilities of each device \cite{rapp2020distributed,wang2020towards}. Early approaches suggest to drop devices that fail to compute pre-determined workloads within a time window \cite{bonawitz2019towards,mcmahan2017communication}. However, Li \etal \cite{li2020federated} showed that this has negative effects on convergence as it limits the number of effective devices contributing to training and may induce bias, if dropped devices have specific data characteristics. Hence, we tolerate partial workload on clients following recent works \cite{li2020federated,reddi2020adaptive}. 

Statistical heterogeneity reflects another major challenge for convergence: whilst in centralized training, data can be assumed independent and identically distributed (i.i.d.), decentralized data is generally highly imbalanced (\eg, local data may contain different number of samples for different classes on each device) and non-i.i.d.\ (\eg, samples in remote clients may have large correlation due to user-specific habits or preferences) \cite{zhao2018federated}, as depicted in Fig.~\ref{fig:graphical_abstract_datadistribution}. 

\textbf{Challenges of FL of visual feature representations:} Representation learning has been a prosperous technique used to perform complex computer vision tasks, such as image classification and segmentation \cite{bengio2013representation,girshick2014rich}. In this paradigm, a model is trained to learn  \textit{rich} feature representations of its inputs, and learned representations are employed by task specific predictors (\eg, classifiers or detectors).  Current FL approaches focus on learning features by considering only statistical properties of data, such as joint distribution of samples and their class labels \cite{hsu2020federated}, and weights of models \cite{li2020federated}.  In FedAvg \cite{mcmahan2017communication}, weights are aggregated with importance scores proportional to size of local datasets, ignoring the learning dynamics. A similar approach has been followed by many subsequent methods \cite{hamer2020fedboost,hsu2020federated,karimireddy2020scaffold,li2020federated}. More recently, increasing interest has been devoted toward elucidating aggregation procedures. Attention methods  \cite{huang2020personalized,ji2019learning,peng2019federated,wu2020fast,yu2019federateddetection} were proposed using functions of difference between parameters of local and aggregate models. However, these works disregard relationship between statistical properties of the learned  representations.

Here, instead, we propose a prototype guided federated optimization method (FedProto), which leverages the model aggregation procedure by prioritizing distributed models on basis of their learned prototypical representations of object categories. In particular, FedProto consists of three steps: 

\textbf{ (i) Prototypical representations:} First, we compute prototypical representations using local and aggregate models motivated by their success in meta-learning \cite{allen2019infinite,cermelli2020few,kim2019variational,li2020boosting,liu2020meta,snell2017prototypical}, domain adaptation \cite{pan2019transferrable,toldo2020unsupervised, toldo2021unsupervised}, semantic segmentation \cite{dong2018few,wang2019panet} and continual learning \cite{michieli2021continual,zhang2019variational}.

\textbf{ (ii) Confidence of local and aggregate models:} Second, we compute confidence of local and aggregate models with respect to their decision on local data using prototypical (hypothesis) margins (PMs). PMs have been explored for developing learning vector quantization (LVQ) methods \cite{hammer2005generalization,jin2010regularized,nova2014review,sato1995generalized,schneider2009adaptive}. In \cite{crammer2002margin}, PMs are shown to lower bound sample margins and provide a rigorous upper bound of generalization error. 
Unlike PMs proposed for individual models, we aim to measure the change in semantic representations of FL models learned at different clients and over different rounds considering their generalization properties. Therefore, we first define a novel semantic PM. Then, motivated by these theoretical results, we drive the model aggregation process combining a confidence measure computed between local prototypes at the beginning and at the end of the local optimization (\ie, \textit{Local PM}), as well as a measure computed between aggregate and local prototypes at the server-level (\ie, \textit{Aggregate PM}). Although margins between features and prototypes have been used to solve other vision tasks (\eg, few-shot learning \cite{li2020boosting}), to our knowledge, our work is the first to compute margins among sets of prototypes and employ them for federated optimization. 

\textbf{ (iii) Prototype-based weight attention:} Finally, we propose a weight attention mechanism during global aggregation of local models  using non-linear functions (\eg, sigmoid) of  prototypical margins. In FL, state-of-the-art attention methods \cite{ji2019learning,wu2020fast} consider only statistics of local models ignoring their effect on decision boundary. Instead, our attention mechanism quantifies this information by margins and employs it for aggregation. We conjecture that our mechanism enables maximization of latent-level margins in FL, which is experimentally justified in Sec.~\ref{sec:results}.

Intuitively, driving the model to focus on class prototypes, we achieve a better shaping of the inner space (thus acting as regularization constraint) that eventually eases the classifier task, which is a harder task than feature extraction \cite{wojna2019devil}. Therefore, FedProto shows better convergence rate and accuracy, ultimately achieving a closer latent space organization to the one centralized training would produce.\\
The main contributions of this paper are as follows:
\setlist{nolistsep}
\begin{itemize}[leftmargin=*,noitemsep]
    \item We propose a novel FL algorithm (FedProto) which applies a \textit{prototypical margin-based model attention mechanism} to drive FL optimization in heterogeneous systems.
    \item We achieve state-of-the-art results on a variety of image classification and semantic segmentation benchmarks. To the best of our knowledge, this is the first work  exploring federated learning of semantic segmentation models.
    \item We propose two quantitative metrics and a qualitative method based on entropy maps to analyse statistical properties of feature representations learned in FL systems.
\end{itemize}





\vspace{-0.2cm}
\section{Federated Learning}
\label{sec:problem}
\vspace{-0.2cm}
In an FL system consisting of a set of clients ${\mathcal{K}=\{ 1,2,\dots,K\}}$, parameters ${W_k \in \mathcal{W}_k}$  of models ${M_k: \mathcal{W}_k \times \mathcal{X}_k \to \mathcal{Y}_k}$, are optimized at each client $k \in \mathcal{K}$ using its local dataset  to learn feature representations, where $\mathcal{X}_k = \{ \mathbf{x}_{k,j} \}_{j=1} ^{n_k}$  and $\mathcal{Y}_k  = \{ \mathbf{y}_{k,j} \}_{j=1} ^{n_k}$ denote respectively the set of samples and their ground truth labels (\eg, one-hot encoded vectors of category labels for image classification, and vectors of segmentation maps for image  segmentation) observed at the client $k$. In centralized FL systems, a central server coordinates the optimization of a set of parameters $\mathcal{W}$ of an aggregated model $M(\mathcal{W}, \cdot)$ by minimizing a global learning objective $L(W)$ \cite{mcmahan2017communication}  without sharing local datasets ${\mathcal{S}_k = \{ s_{k,j} = (\mathbf{x}_{k,j} , \mathbf{y}_{k,j}) \}_{j=1} ^{n_k}} $ by solving 
\vspace{-0.2cm}
\begin{equation}
    \min_{W \in \mathcal{W}} L(W) = \min_{W\in \mathcal{W}} \sum_{k\in\mathcal{K}} p_k L_k (W; \mathcal{S}_k),
    \label{eq:fedavg}
    \vspace{-0.2cm}
\end{equation}
where  the local objective is computed by
\vspace{-0.2cm}
\begin{equation}
\label{eq:L_k}
    L_k(W; \mathcal{S}_k) = \frac{1}{n_k} \sum_{j=1}^{n_k} l_k (W; s_{k,j} \in \mathcal{S}_k),
    \vspace{-0.2cm}
\end{equation}
with $l_k (\cdot;\cdot)$ being a user-specific loss function, $p_k \geq 0$ is the weight of the objective $L_k(\cdot ; \cdot)$ of the $k^{th}$ client and $\sum_{k\in\mathcal{K}} p_k = 1$. McMahan \etal \cite{mcmahan2017communication} proposed to use ${p_k = \frac{n_k}{n}}$, where ${n=\sum_{k\in\mathcal{K}} n_k}$. Thereby,  $L(W)$ coincides with the training objective of the centralized setting. 

Federated averaging  (FedAvg) \cite{Li2020On,mcmahan2017communication} is a benchmark federated optimization algorithm  widely used to solve the problem \eqref{eq:fedavg}.
%
In FedAvg, a subset $\mathcal{K}^t \subseteq \mathcal{K}$ of $K'$ clients are selected according to $p_k$ at each federated round $t$. 
Selected clients $k \in \mathcal{K}^t $ download the aggregated model $W^t \in \mathcal{W}^t$ from a central server, perform local optimization minimizing an empirical objective $L_k(W^t;\mathcal{S}_k)$ with learning rate $\eta$ for $F$ epochs using a local optimizer such as SGD, and then send the final solution $W^{t+1}_k$ back to the server. The server averages the solutions obtained from the clients with weights proportional to the size of the local datasets by 
\begin{equation}
W^{t+1} = \sum_{k\in \mathcal{K}^t} \frac{n_k}{\sum \limits _{j \in \mathcal{K}^t} n_j} W_k ^{t+1} .    
\end{equation}
The procedure is iterated for $T-1$ federated rounds and the final aggregate model is then identified by $W^T$.

Optimizing local models with the same number of local epochs at all clients is unfeasible for real-world applications \cite{bonawitz2019towards, hsu2020federated,li2020federated}. A more natural approach is to allow the epochs to vary according to the characteristics of the FL system, and to properly merge solutions accounting for heterogeneity of the system, as we formalize next.


\vspace{-0.1cm}
\section{Prototype Guided Federated Learning}
\label{sec:method}
\vspace{-0.1cm}
	
Prototypical representations have been successfully employed in various computer vision tasks \cite{kim2019variational,michieli2021continual,pan2019transferrable,snell2017prototypical}.
In this work, we employ prototypes for federated optimization of vision models. Our prototype guided federated optimizer (FedProto) is motivated by the results obtained from the recent theoretical and experimental analyses of generalization capacity of latent class-conditional prototypes \cite{kim2019variational,snell2017prototypical}.
%
%

\textbf{Partial workload toleration:}
First of all, we observe that different clients in FL systems are likely to have very different resource constraints (causing system heterogeneity), such as different data resources, hardware configurations (\eg, visual sensors, cameras or processors), network connections and battery levels \cite{kairouz2021advances}. Therefore, we allow partial amount of work to be conducted locally by each client prior to the aggregation stage, as utilized in FedProx \cite{li2020federated}. In other words, at each round, instead of dropping $\delta\%$ of clients that performed less epochs than the total number $F$ in a predetermined amount of time, we aggregate all the solutions sent from local clients tolerating partial workload, \ie, even if the completed number of local epochs is $F'<F$. Following \cite{li2020federated}, we mimic this behavior by uniformly sampling $F' \sim \mathcal{U}([0, F))$ on each client.

At each round $t$ and client $k$,  a local model $M_k^t (\mathcal{W}_k^t;\mathcal{X}_k) = C_k^t (\mathcal{W}_{c,k}^t) \circ E_{k}^t (\mathcal{W}_{e,k}^t;\mathcal{X}_k)$ is computed, where $\circ$ denotes function composition, and $\mathcal{W}_{c,k}^t \subset \mathcal{W}_k^t$ and $\mathcal{W}_{e,k}^t \subset \mathcal{W}_k^t$ denotes sets of parameters of classifiers and encoders embodied in the model $M_k^t$, respectively. For each input $\mathbf{x}_{k,j} \in \mathcal{X}_k$, its latent representation $\mathbf{e}_{k,j}^t = E_k^t(\mathcal{W}_{e,k}^t;\mathbf{x}_{k,j} )$ is computed and then fed to a classifier $C_k^t(\mathcal{W}_{c,k}^t;\mathbf{e}_{k,j}^t)$ to retrieve class-wise probability scores. 
Features corresponding to the same class are then averaged to construct local latent class-conditional prototypes. 
 \vspace{-0.2cm}
\subsection{Computation of Prototypes}
\vspace{-0.15cm}
 At each  round $t>0$, class $c \in \mathcal{C}$ and client $k \in \mathcal{K}^t$, the $c^{th}$ element of prototypes $\mathbf{p}_{k}^t$ is computed  by
 \vspace{-0.2cm}
\begin{equation}
\label{eq:local_protos}
    \mathbf{p}_{k}^t [c] = 
    \sum_{\mathbf{e}_{k,j,c}^t \in \mathcal{F}_{k,c}^t} 
    \frac{\mathbf{e}_{k,j,c}^t} 
    { \mathbf{n}_k^t[c] }, \quad \forall c \in \mathcal{C}, \forall k \in \mathcal{K}^t,
    \vspace{-0.15cm}
\end{equation}
where 
$\mathcal{F} _{k,c} ^t$ is the set of feature vectors $\mathbf{e}_{k,j,c}^t $ extracted from the sample $\mathbf{x}_{k,j} \in \mathcal{X}_k$ belonging to the class $c$, 
and ${\mathbf{n}_k^t[c] = |\mathcal{F} _{k,c} ^t |}$ is the cardinality of $\mathcal{F} _{k,c} ^t $. At $t=0$, we initialize prototypes as ${\mathbf{p}_{k}^0 [c]=\mathbf{0}}$, $\forall c, k$. Since features representing different classes have variable norm \cite{xu2019larger}, we employ min-max normalization over the channels and denote the normalized prototypes by $\hat{\mathbf{p}}_{k}^t$. 


\vspace{-0.1cm}
\subsection{Local and Aggregate Prototype Margins}
\vspace{-0.1cm}
To guide the optimization, we rely on a combination of two clues derived from displacement of prototypes:
\setlist{nolistsep}
\begin{enumerate}[leftmargin=*,noitemsep]
\item \textit{Local Prototype Margin (LPM)} measures deviation of on-client prototypes before and after local training. 
\item \textit{Aggregate Prototype Margin (APM)} measures deviation of aggregate prototypes from local prototypes.
\end{enumerate}
As a measure for displacement, we embraced the margin theory  \cite{crammer2002margin,hammer2005generalization,li2020boosting,nova2014review,rosset2003margin,sato1995generalized}, in which PMs measure the distance between features and  class decision boundaries. In our work, instead, we aim to measure change of semantic representations among clients over different rounds for FL. Therefore, we propose a novel semantic PM next.
\vspace{-0.2cm}
\begin{definition}[Semantic PM - SPM]
Given two prototype vectors $\mathbf{p}_i$ and $\mathbf{p}_j$ defined on the same class space $\mathcal{C}$, we restrict to $\mathcal{C}'\subset \mathcal{C}$ such that $\mathbf{n}_i [c] >0$ and $\mathbf{n}_j[c]>0$, $\forall c\in\mathcal{C}'$, the distance between prototypes corresponding to the same semantic label $c$ is computed by
\vspace{-0.10cm}
\begin{equation}
\label{eq:d_plus}
    \mathbf{d}^+_{i,j}[c] =
    d(\mathbf{p}_i[c], \mathbf{p}_j[c]),
    \quad \forall c\in\mathcal{C}'
    ,
    \vspace{-0.1cm}
\end{equation}
and the average distance between prototype of a certain class $c$ and prototypes of different classes is computed by
\begin{equation}
    \mathbf{d}^-_{i,j}[c] =
    \sum_{\substack{c' \neq c\\ c'\in\mathcal{C}'}}
    \frac{d(\mathbf{p}_i[c], \mathbf{p}_j[c'])}
    { |   c' \neq c \land c'\in\mathcal{C}'     | }, \quad \forall c\in\mathcal{C}'
    .
    \vspace{-0.05cm}
\end{equation}
Then, the SPM for class $c$ is defined by
\vspace{-0.05cm}
\begin{equation}
   \mu (\mathbf{p}_i[c], \mathbf{p}_j) \overset{def}{=} \frac{\mathbf{d}^-_{i,j}[c] - \mathbf{d}^+_{i,j}[c]}
    {\mathbf{d}^-_{i,j}[c] + \mathbf{d}^+_{i,j}[c]}, \quad \forall c\in\mathcal{C}'
    .
    \vspace{-0.05cm}
\end{equation}
%
\end{definition}
%
In FL, we employ SPMs in two cases, LPM and APM, which are defined in Def.~\ref{def:LPM} and~\ref{def:APM}. In the analyses, we identify $d(\cdot, \cdot)$ by the Euclidean distance, since it has been shown to outperform cosine similarity \cite{dong2018few,snell2017prototypical} or to achieve comparable performance \cite{oreshkin2018tadam,wang2019panet}.
\vspace{-0.1cm}
\begin{definition}[LPM]\label{def:LPM}
The LPM is defined by 
\vspace{-0.1cm}
\begin{equation}
\label{eq:LPM}
    \boldsymbol\mu_{\mathrm{loc},k}^t [c] \overset{def}{=}
    \mu
    ( \hat{\mathbf{p}}_k^{t-1}[c] ,
    \hat{\mathbf{p}}_k^t), 
    \quad \forall k \in \mathcal{K}^t ,
    \forall c \in \mathcal{C}
    \vspace{-0.1cm}
\end{equation}
and it measures  change of local prototypes obtained from local models before and after their local training.
\end{definition}
\vspace{-0.1cm}
\begin{definition}[APM]\label{def:APM}
The APM is defined to measure discrepancy between local and aggregate set of prototypes by
\vspace{-0.15cm}
\begin{equation}
\label{eq:APM}
    \boldsymbol\mu_{\mathrm{agg},k}^t[c] \overset{def}{=}
    \mu
    ( \hat{\mathbf{p}}_k^{t}[c] , \hat{\mathbf{p}}_{\mathrm{agg}}^{t-1} ) ,
    \quad \forall k \in \mathcal{K}^t , \forall c \in \mathcal{C}
    \vspace{-0cm}
\end{equation}
where the aggregate set of prototypes is defined by
\vspace{-0.05cm}
\begin{equation}
\label{eq:aggregate_protos}
    \hat{\mathbf{p}}_{\mathrm{agg}}^t [c] \overset{def}{=} \sum_{k\in\mathcal{K}^t} \frac{\mathbf{n}_k^t [c]}
    {\mathbf{n}_{\mathrm{agg}}^t [c]}
    \hat{\mathbf{p}}_{k}^t [c],
    \quad \forall c \in \mathcal{C},
    \vspace{-0.05cm}
\end{equation}
with aggregate number of features ${\mathbf{n}_{\mathrm{agg}}^t [c] = \sum_{k \in \mathcal{K}^t} \mathbf{n}_k^t [c]}$, $\forall c \in \mathcal{C}$ and initialization $\hat{\mathbf{p}}_{agg}^0 [c]=\mathbf{0}$.
\end{definition}

We remark that APM requires transmission of prototypes from clients to server. However, this does not raise privacy issues since prototypes represent only an averaged statistic over all local data of already compressed feature representations, nor large communication overhead, as the size of prototypes is negligible compared to the model size. While local deviation measured by LPM gives a hint of how much a model adapts its inner representation for each class, server-side deviation measured by APM tells how much a local model changes its inner representations with respect to the prototypical representations aggregated over previous rounds and clients. The effect of distributed versus centralized calculation of margins is analysed in Sec.~\ref{subsec:results:classification}.
 
 \subsection{Federated Attention using Prototype Margins}
Client deviations are computed by summing over all the classes and applying a sigmoid function $\sigma$ \cite{nova2014review,sato1995generalized} by
\begin{equation}
    \mathbf{v}_{\iota}^t[k] = \sigma \!\left(
    \sum_{c \in \mathcal{C}} \boldsymbol\mu_{\iota,k}^t [c] \right)\!,
    \ \ 
    \forall k \in \mathcal{K}^t, 
    \iota \in 
    \left\lbrace \mathrm{loc,\ agg}  \right\rbrace\!.
\end{equation}
\begin{definition}[Local, aggregate and federated attention]
A local (aggregate) weight attention vector $\mathbf{a}_{\mathrm{loc}}^t$ ($\mathbf{a}_{\mathrm{agg}}^t$) is computed normalizing the client deviations by 
\begin{equation}
    \mathbf{a}_{\iota}^t[k] \overset{def}{=} 
    \frac{\mathbf{v}_{\iota}^t[k]}
    {\sum_{j\in\mathcal{K}^t} \mathbf{v}_{\iota}^t[j]},
    \quad 
    \forall k \in \mathcal{K}^t,
    \iota \in 
    \left\lbrace \mathrm{loc,\ agg}  \right\rbrace.
\end{equation}
The federated weight attention vector $\mathbf{a}^t$ is defined by
\begin{equation}
\label{eq:attention_vector}
\displaystyle
    \mathbf{a}^{t}[k] \overset{def}{=}  
    \begin{cases}
       \frac{n_k}{\sum_{j\in\mathcal{K}^t} n_j}, & \text{if}\ t=0 \\
      \frac{\mathbf{a}_{\mathrm{agg}}^{t}[k] + \mathbf{a}_{\mathrm{loc}}^{t}[k]}{2}, & \text{if}\ t>0
    \end{cases}
\end{equation}
\end{definition}
%
%
%
%
Intuitively, each $\mathbf{a}^t[k]$ represents a measure of client drift: as prototypes computed using weights $W_k \in \mathcal{W}_k$ of a model of a client $k$ deviate from  reference prototypes in terms of margin (either locally or on server), higher attention is applied on the weights $W_k \in \mathcal{W}_k$, and vice-versa. 
We remark that, according to our definition, if a client is not able to build reliable latent representations (low margin), then its model is considered less during aggregation.

Finally, federated attention vectors  $\mathbf{a}^t$ are used to aggregate local weights at each $t^{th}$ round by
\begin{equation}
    W^{t+1} = \sum_{k \in \mathcal{K}^t} \mathbf{a}^t[k] W_k^t.
\label{eq:fedproto}
\end{equation}

\begin{algorithm}[tb]
  \caption{FedProto.}
  \label{alg:fedproto}
\begin{algorithmic}
  \STATE {\bfseries Input:} $\mathcal{K}, T, F, W^0, \eta, N$.
  \FOR{$t=0$ {\bfseries to} $T-1$}
  \STATE A server samples $\mathcal{K}^t \subseteq \mathcal{K}$ clients $\propto p_k$, and sends $W^t$.
      \FOR{$k \in \mathcal{K}^t$}
      \STATE Compute local prototypes \eqref{eq:local_protos}.
      \STATE Update $W^t_k$ with $L_k$ \eqref{eq:L_k} and step size $\eta$ to $W^{t+1}_k$.
      \STATE Compute local prototypes \eqref{eq:local_protos} and LPM \eqref{eq:LPM}.
      \STATE Send $W^{t+1}_k$, LPM and $\hat{\mathbf{p}}_k^t$ back to the server.
      \ENDFOR
  \STATE The server computes APM \eqref{eq:APM}, $\mathbf{a}^t$ \eqref{eq:attention_vector} and  $W^{t+1}$ \eqref{eq:fedproto}.
  \ENDFOR
\end{algorithmic}
\end{algorithm}

Our proposed FL method which employs \eqref{eq:fedproto} to solve \eqref{eq:fedavg} is called FedProto and is summarized in Algorithm~\ref{alg:fedproto}.


\section{Experimental Setup}
\label{sec:implementation}

\begin{table*}[]
\centering
\caption{Statistics of the employed datasets (left) and hyper-parameters (right). In segmentation datasets, image background is excluded, and the accuracy refers to the mIoU. DeepLab-V3+ \cite{chen2018encoder} uses MobileNet-V2 \cite{howard2017mobilenets,sandler2018mobilenetv2} as the backbone pre-trained on ImageNet \cite{krizhevsky2012imagenet}.}
\footnotesize
\setlength{\tabcolsep}{3.5pt}
\begin{tabular}{lcccccccc:ccccc}
\hline
\textbf{Dataset} & \textbf{\# Classes} & \textbf{Clients} & \textbf{Samples} & \multicolumn{2}{c}{\textbf{Samples/Client}} & \textbf{Model} & \textbf{Distribution} & \textbf{Central.} & \textbf{Start lr}  & \textbf{Solver} & $\mathbf{F}$ & \textbf{Rounds} & \textbf{Batch} \\
& & & & Mean & Std. & & & \textbf{Acc. (\%)} & & & & & \textbf{size} \\\hline
Synthetic & $10$ & $30$ & $9,600$ & $320.0$ & $1051.6$ & $2$ dense layers & Power-law & $78.5$ & $0.01$ & SGD & $20$ & $200$ & $10$ \\\hdashline 
MNIST & $10$ & $1,000$ & $61,676$ & $61.7$ & $164.7$ & $2$-layer CNN & Power-law & $99.0$ & $0.01$ & SGD & $20$ & $200$ & $10$ \\
FEMNIST & $10$ & $200$ & $16,421$ & $82.1$ & $143.0$ & $2$-layer CNN & Power-law & $99.0$ & $0.001$ & SGD & $20$ & $400$ & $10$ \\
CelebA & $2$ & $9343$ & $177,457$ & $19.0$ & $7.0$ & $4$-layer CNN & Power-law & $92.6$ & $0.1$ & SGD & $20$ & $200$ & $10$\\\hdashline
FPascal Macro & $4$ & $100$ & $6,665$ & $66.7$ & $25.7$ & DeepLab-V3+ & Power-law & $79.7$ & $10^{-4}$ & Adam & $2$ & $400$ & $16$\\
FPascal & $20$ & $100$ & $6,665$ & $66.7$ & $25.7$ & DeepLab-V3+ & Power-law & $66.3$  & $10^{-4}$ & Adam & $2$ & $400$ & $16$ \\\hline
\end{tabular}
\label{tab:datasets_summary}
\end{table*}

\begin{figure*}
\centering
	\includegraphics[trim=4.75cm 0.35cm 4.75cm 1.4cm, clip, width=\linewidth]{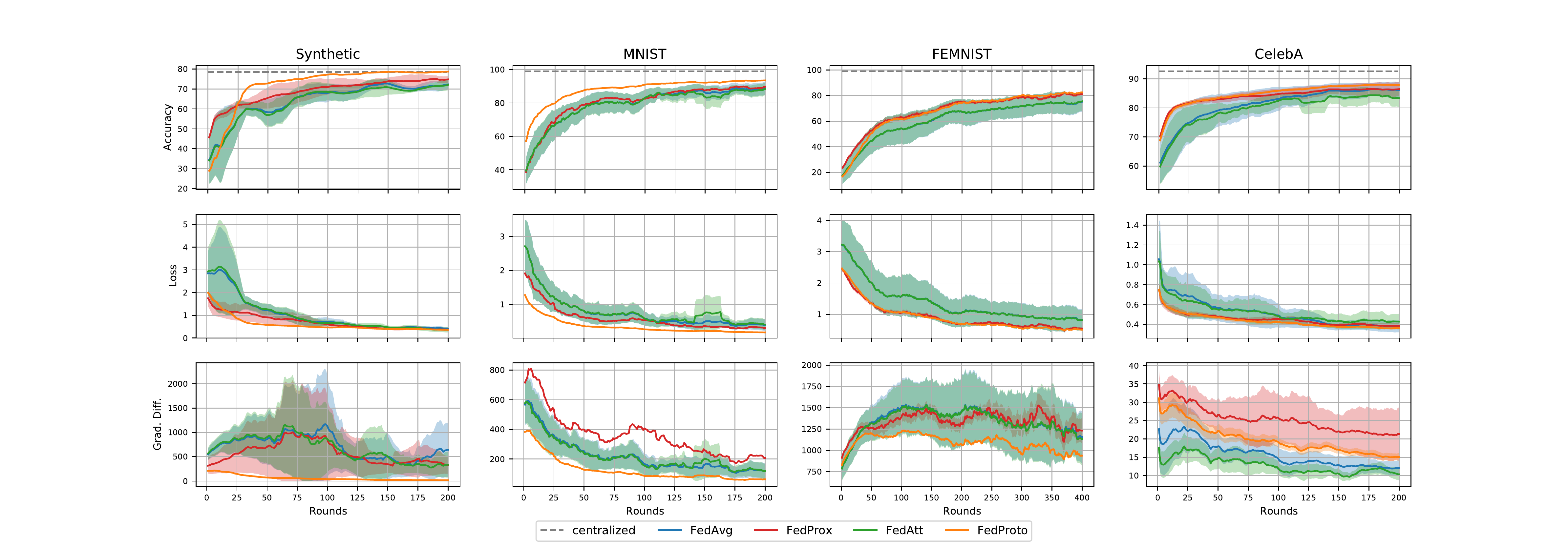}
	\caption{Experimental results for the classification task. Evaluation is performed across $\delta \in \{ 0\%, 50\%, 80\% \}$ and a moving average window of $10\%$ rounds is applied for visualization. Solid lines and shaded regions represent the mean and standard deviation, respectively.}
	\vspace{-0.2cm}
	\label{fig:classification_complete}
\end{figure*}

We evaluate on various tasks, models and real-world federated vision datasets. Full details on the experimental setup are given in Suppl. Mat. and are summarized in Table~\ref{tab:datasets_summary}.


\textbf{Classification Data.}  We evaluate FedProto on four classification datasets adopted from the related work \cite{caldas2018leaf,hsu2020federated,mcmahan2017communication}. 
First, we generate \textbf{synthetic} data following \cite{li2020federated,shamir2014communication}, with addition of heterogeneity among clients. We sample from a logistic regression model with two parameters: $\phi_1$, controlling how much local models differ from each other, and $\phi_2$ controlling how much local data distribution at each client differs from that of other clients. To obtain highly non-i.i.d.\ data, we set $\phi_1=\phi_2=1$ being the most heterogeneous, yet challenging, scenario. Other analyses have been carried out in \cite{li2020federated}. Assuming that the  data generation model is agnostic, we use a cascade of $2$ dense and a softmax layer. 

Then, we employ some real-world classification data. We distribute \textbf{MNIST} \cite{lecun1998gradient} data among $1,000$ clients such that each client has samples of only $2$ digits (out of $10$) and the number of samples per client follows a power-law \cite{caldas2018leaf}. We use the federated version of EMNIST \cite{cohen2017emnist,caldas2018leaf} (\textbf{FEMNIST}) proposed in \cite{li2020federated} where $10$ lower-case letters are subsampled and only 5 classes are distributed to each client. 
Finally, we generate non-i.i.d.\ \textbf{CelebA} \cite{liu2015faceattributes} data (for smile classification), such that the underlying data distribution for each user is consistent with the raw data. 

\textbf{Image Segmentation Data.} We use the VOC2012 \cite{everingham2010pascal} semantic segmentation dataset. We restrict to images with one single class inside (and the background) to mimic classification splits. We devise two class sets: macro ($4$ classes) and standard ($20$ classes). We distribute data to each client according to a Dirichlet distribution over the number of classes on each client with concentration parameter $\alpha > 0$, where low $\alpha$ values mean high non-i.i.d.\ data among clients, and vice-versa \cite{hsu2019measuring,hsu2020federated}. The number of samples per client follows a power-law distribution with parameter $\gamma=3$. 


\textbf{Implementation Details.} Utilized hyper-parameters are reported on the right side of Table~\ref{tab:datasets_summary}. We tuned learning parameters of each dataset on FedAvg (with $F=1$ and no system heterogeneity) and, for fair comparison, we use the same parameters on all experiments for that dataset. 
We set $|\mathcal{K}^t| = 10, \forall t$ for all datasets. Randomly selected clients 
and mini-batch orders are kept fixed across all runs for comparative experiments. 
For simplicity, we use a constant learning rate on classification tasks, and polynomially decaying learning rate with power $0.9$ and weight decay $4\cdot 10^{-5}$ \cite{chen2017deeplab,chen2018encoder} for segmentation tasks.
For classification data, we measure accuracy as the percentage of correct predictions, whilst for segmentation data we use the mean Intersection over the Union (mIoU). All simulations are performed for $10\%$ rounds more, and metrics are moving averaged over a window of $10\%$ rounds in the visualization. The algorithms are implemented in Tensorflow \cite{abadi2016tensorflow} and are trained on a single NVIDIA RTX 2080Ti GPU.



\section{Experimental Analyses for Federated Vision}
\label{sec:results}

\subsection{Federated Image Classification}
\label{subsec:results:classification}

In this section, we report an extensive evaluation of our approach on image classification tasks. We compare FedProto with the baseline FedAvg, with the state-of-the-art regularization-based FedProx \cite{li2020federated} and with FedAtt \cite{ji2019learning}, which employs weight-based attention. Fig.~\ref{fig:classification_complete} shows per-round aggregate accuracy, training loss and gradients difference on the four classification datasets introduced in Sec.~\ref{sec:implementation}.
%
%
From the first row of Fig.~\ref{fig:classification_complete}, we observe that FedProto robustly outperforms FedAvg in terms of accuracy on every dataset. FedAtt brings minimal improvement compared to FedAvg, proving that a simple weight-based attentive mechanism is not very useful in vision tasks. FedProx, instead, leverages accuracy thanks to the toleration of partial workload and presence of the proximal term. However, our approach can effectively match or surpass the accuracy of FedProx. Additionally, we observe that both FedProto and FedProx show much lower variance (narrower shaded region) than competing approaches by tolerating partial results. 
Similar considerations are also reflected on the training loss (second row).
The third row reports the average of squared $\ell_2$ norm of difference of gradients over all clients, \ie, $ \frac{1}{|\mathcal{K}|}\sum _{k \in \mathcal{K}} || \nabla L_k (W^t;\mathcal{S}_k) - \nabla L(W^t) ||_2^2$. As in \cite{li2020federated}, we interpret this dissimilarity measure as a proxy of accuracy. In particular, we observe how FedProto shows smaller dissimilarity (\ie, better convergence \cite{li2020federated}) compared to FedProx, thanks to the regularization effects brought by the proposed modules.
To better appreciate accuracy and loss gaps observed in Fig.~\ref{fig:classification_complete}, we give results obtained using the aggregate models at the final round in Table~\ref{tab:classification_complete} where FedProto shows significant improvements across all the datasets.

\textbf{Ablation Studies.} To explore the effect of the components of our approach on accuracy, we report a comparative ablation study in Table~\ref{tab:mnist_ablation}. 
First, we noticed that our approach tends to produce weights $\mathbf{a}^t$ deviating less from a fairness policy (\ie, aggregation of weights $W_k^t$ by ${\mathbf{f}^t[k] = 1 / K'}, \forall k, \forall t$) than FedAvg, as also observed in other contemporary approaches \cite{wu2020fast,zhang2020hierarchically}. 
A fair policy (row 2), indeed, outperforms FedAvg by a small margin. However, we argue that this is an implicit effect of the weighted sampling scheme of the active clients at each round introduced in Sec.~\ref{sec:problem}. As a matter of fact, sampling active clients i.i.d.\ (row 3) brings results comparable to FedAvg. 
Second, we analyse the toleration of partial workload. Adding it on top of na\"ive implementations of FedAvg and Fairness (rows 4 and 5) improves the accuracy and the robustness over different amounts of $\delta$. At the same time, we remark that our approach can achieve competitive performance even without tolerating partial workload (row 6). 

Analyses of our model design are given in the last block of Table~\ref{tab:mnist_ablation}. Margin-based deviation can be viewed as an enhanced measure rather than just using the distance between prototypes belonging to the same class, \ie, using only $d^+$ (row 7) from \eqref{eq:d_plus} to accommodate the class-wise probability distribution obtained from the distributed clients during aggregation. Although providing considerable improvements compared to FedAvg, we found margin-based deviation to be generally more stable. Finally, employing only one of the two proposed clues (LPM and APM in rows 8 and 9) still improves accuracy, and the combination of the two (last row) outperforms the effect of the singular components.

\begin{table}[]
\centering
\caption[Caption]{Final mean and std of accuracy (\%) and loss from Fig.~\ref{fig:classification_complete}. Centralized accuracy are $78.5$, $99.0$, $99.4$, $92.6$, and losses are $0.33$, $0.00$, $0.00$, $0.15$ for Synth., MNIST, FEMNIST and CelebA.}
\label{tab:classification_complete}
\footnotesize
\setlength{\tabcolsep}{2pt}
\begin{tabular}{clcccc}
 & & FedAvg & FedProx & FedAtt & FedProto \\\hline
\multirow{4}{*}{\rotatebox[origin=c]{90}{Accuracy}} &  \multicolumn{1}{|l}{Synthetic} & $72.3 \pm 2.6$ & $74.8 \pm 1.6$ & $72.1 \pm 2.7$ & $\mathbf{78.7 \pm 0.2}$ \\
& \multicolumn{1}{|l}{MNIST} & $88.8 \pm 3.8$ & $91.7 \pm \mathbf{0.2}$ & $88.4 \pm 3.7$ & $\mathbf{93.3 \pm 0.2}$ \\
& \multicolumn{1}{|l}{FEMNIST} & $75.1 \pm 7.7$ & $81.1 \pm 1.0$ & $75.5 \pm 7.5$ & $\mathbf{82.5 \pm 0.3}$ \\
& \multicolumn{1}{|l}{CelebA} & $86.2 \pm 2.8$ & $86.4 \pm 2.4$ & $83.4 \pm 3.0$ & $\mathbf{87.8 \pm 0.4}$\\\hline
\multirow{4}{*}{\rotatebox[origin=c]{90}{Loss}} & \multicolumn{1}{|l}{Synthetic} & $0.41 \pm 0.06$ & $0.37 \pm 0.07$ & $\mathbf{0.36} \pm 0.12$ & $\mathbf{0.36 \pm 0.02}$ \\
& \multicolumn{1}{|l}{MNIST} & $0.39 \pm 0.17$ & $0.30 \pm \mathbf{0.02}$ & $0.41 \pm 0.16$ & $\mathbf{0.18 \pm 0.02}$ \\
& \multicolumn{1}{|l}{FEMNIST} & $0.83 \pm 0.35$ & $0.55 \pm 0.04$ & $0.81 \pm 0.34$ & $\mathbf{0.51 \pm 0.01}$ \\
& \multicolumn{1}{|l}{CelebA} & $0.38 \pm 0.06$ & $0.39 \pm 0.03$ & $0.43 \pm 0.08$ & $\mathbf{0.36 \pm 0.02}$ \\\hline
\end{tabular}
\end{table}

\begin{table}[]
\centering
\caption{MNIST classification accuracy ($\%$) of different strategies.}
\footnotesize
\label{tab:mnist_ablation}
\begin{tabular}{|l|c|c|c|c|}
\cline{2-4}
\multicolumn{1}{c}{} & \multicolumn{3}{|c|}{$\delta$} \\\hline
Method & $0\%$ & $50\%$ & $80\%$ & Avg. $\pm$ Std. \\\hline
FedAvg & $92.7$ & $88.7$ & $85.1$ & $88.8 \pm 3.8$ \\  
Fairness & $92.8$ & $89.9$ & $86.5$ & $89.7 \pm 3.2$\\
Fairness sampling i.i.d.\ & $92.5$ & $88.4$ & $84.9$ & $88.6 \pm 3.8$ \\\hdashline
FedAvg $+$ toleration & $92.7$ & $90.2$ & $89.1$ & $90.7 \pm 1.8$  \\
Fairness $+$ toleration & $92.8$ & $91.2$ & $90.6$ & $91.5 \pm 1.1$ \\
FedProto (no toleration) & $\mathbf{93.5}$ & $90.8$ & $88.1$ & $90.8 \pm 2.7$ \\\hdashline
FedProto ($d^+$) only & $92.8$ & $92.4$ & $92.1$ & $92.4 \pm 0.4$ \\
FedProto (APM only) & $93.0$ & $92.7$ & $92.6$ & $92.8 \pm \mathbf{0.2}$\\
FedProto (LPM only) & $91.9$ & $91.0$ & $90.6$ & $91.2 \pm 0.7$ \\
FedProto & $\mathbf{93.5}$ & $\mathbf{93.4}$ & $\mathbf{93.1}$ & $\mathbf{93.3 \pm 0.2}$ \\\hline 
\end{tabular}
\end{table}

\begin{figure}[htbp]
\centering
\setlength{\tabcolsep}{0.5pt} 
\renewcommand{\arraystretch}{0.4}
\centering
\footnotesize
\begin{tabular}{cccc}
  
   & \tiny \textbf{Synthetic} & & \tiny \textbf{MNIST} \\
  
   \tiny \rotatebox{90}{\quad \quad \quad \quad \ \ \ $\bar{\boldsymbol{\mu}}[t]$} &
   \includegraphics[trim=3.7cm 1.3cm 4.5cm 2.3cm, clip, width=0.485\linewidth]{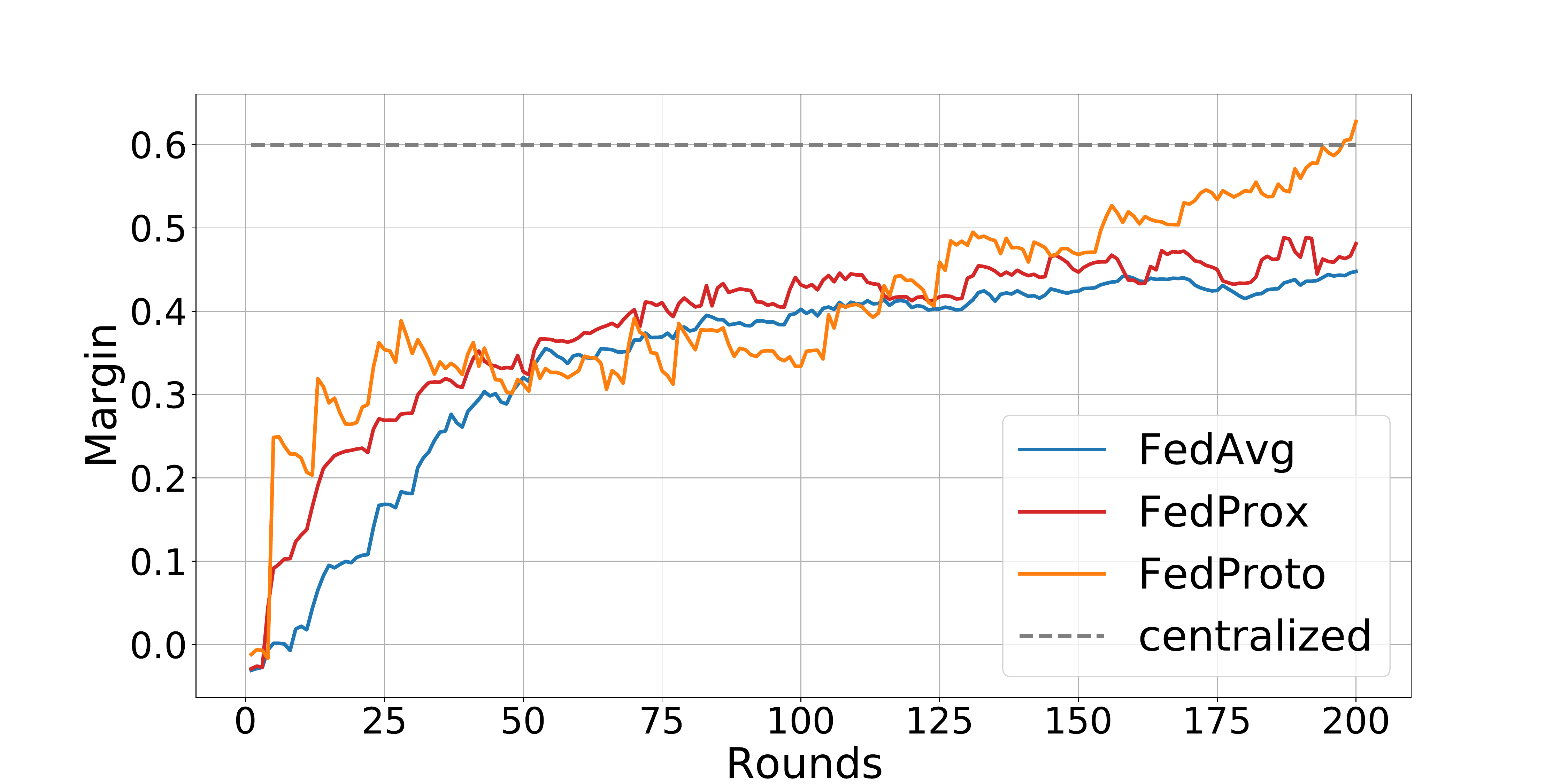} & &
   \includegraphics[trim=3.7cm 1.3cm 4.5cm 2.3cm, clip, width=0.485\linewidth]{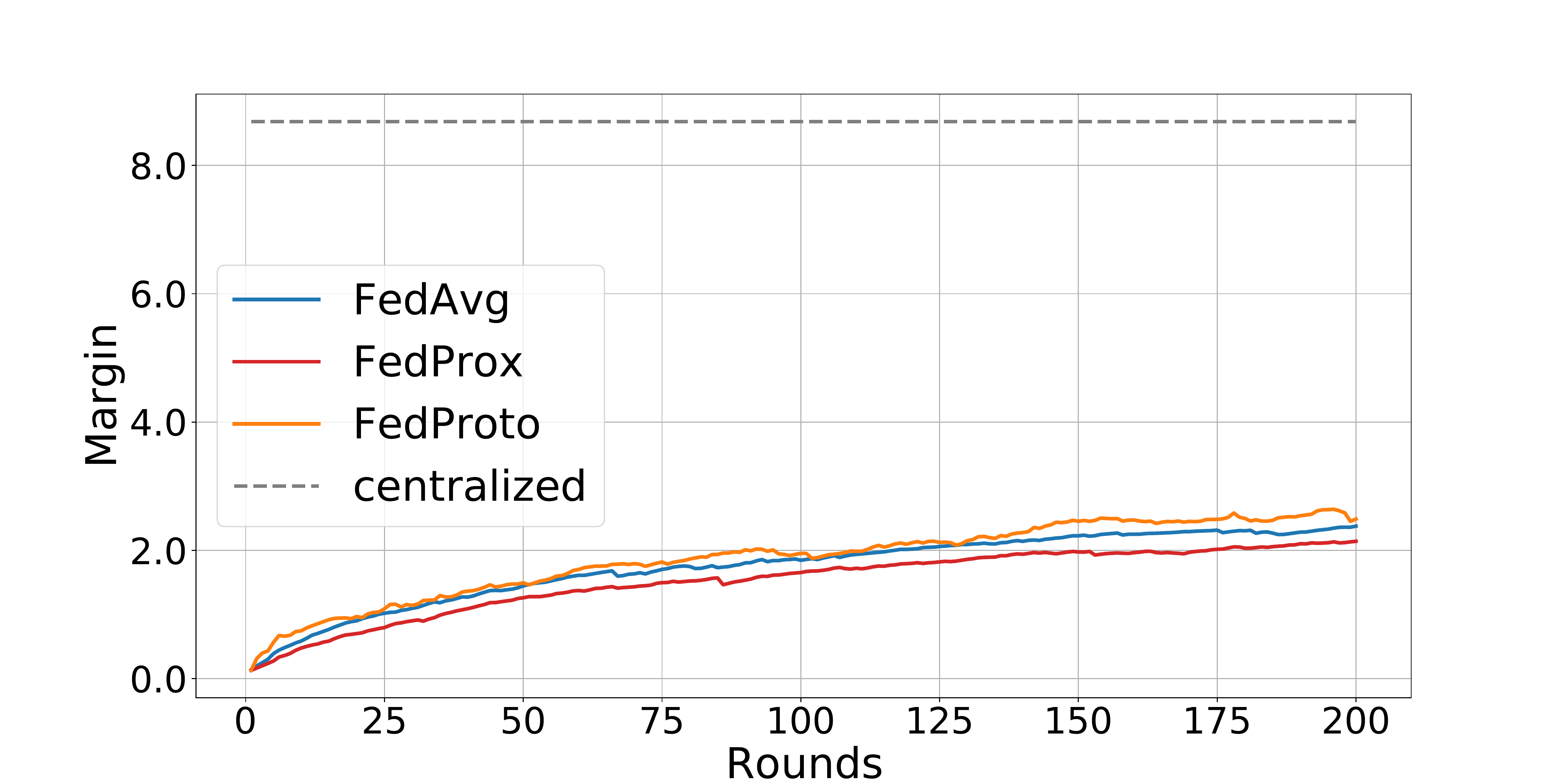} \\
   
   & \hspace{0.1cm} \tiny Rounds & & \hspace{0.1cm} \tiny Rounds \\
   & & & \\
   & \tiny \textbf{FEMNIST} & & \tiny \textbf{CelebA} \\
   
   \tiny \rotatebox{90}{\quad \quad \quad \quad \ \ \ $\bar{\boldsymbol{\mu}}[t]$} &
   \includegraphics[trim=3.7cm 1.3cm 4.5cm 2.3cm, clip, width=0.485\linewidth]{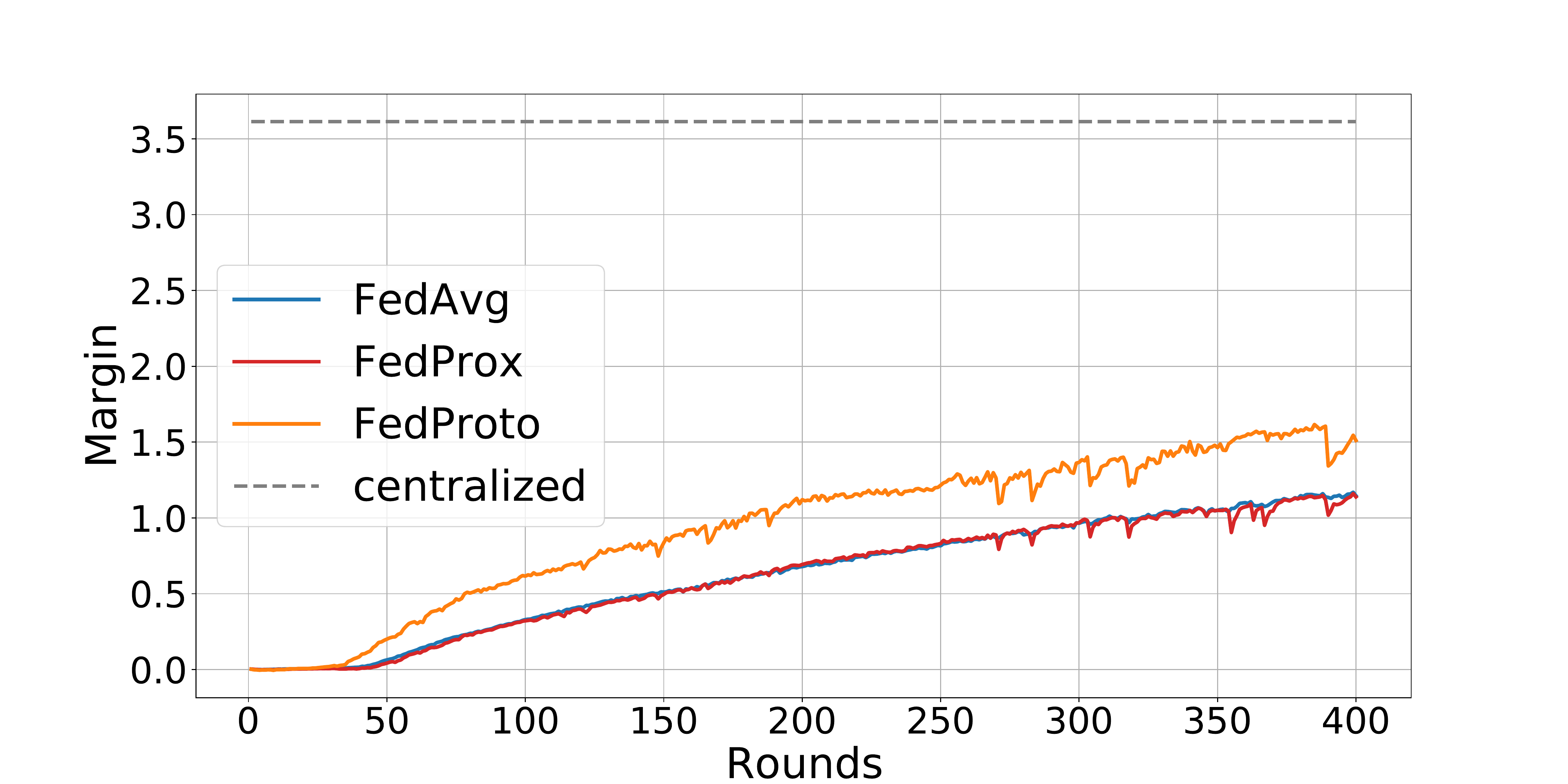} & &
   \includegraphics[trim=3.7cm 1.3cm 4.5cm 2.3cm, clip, width=0.485\linewidth]{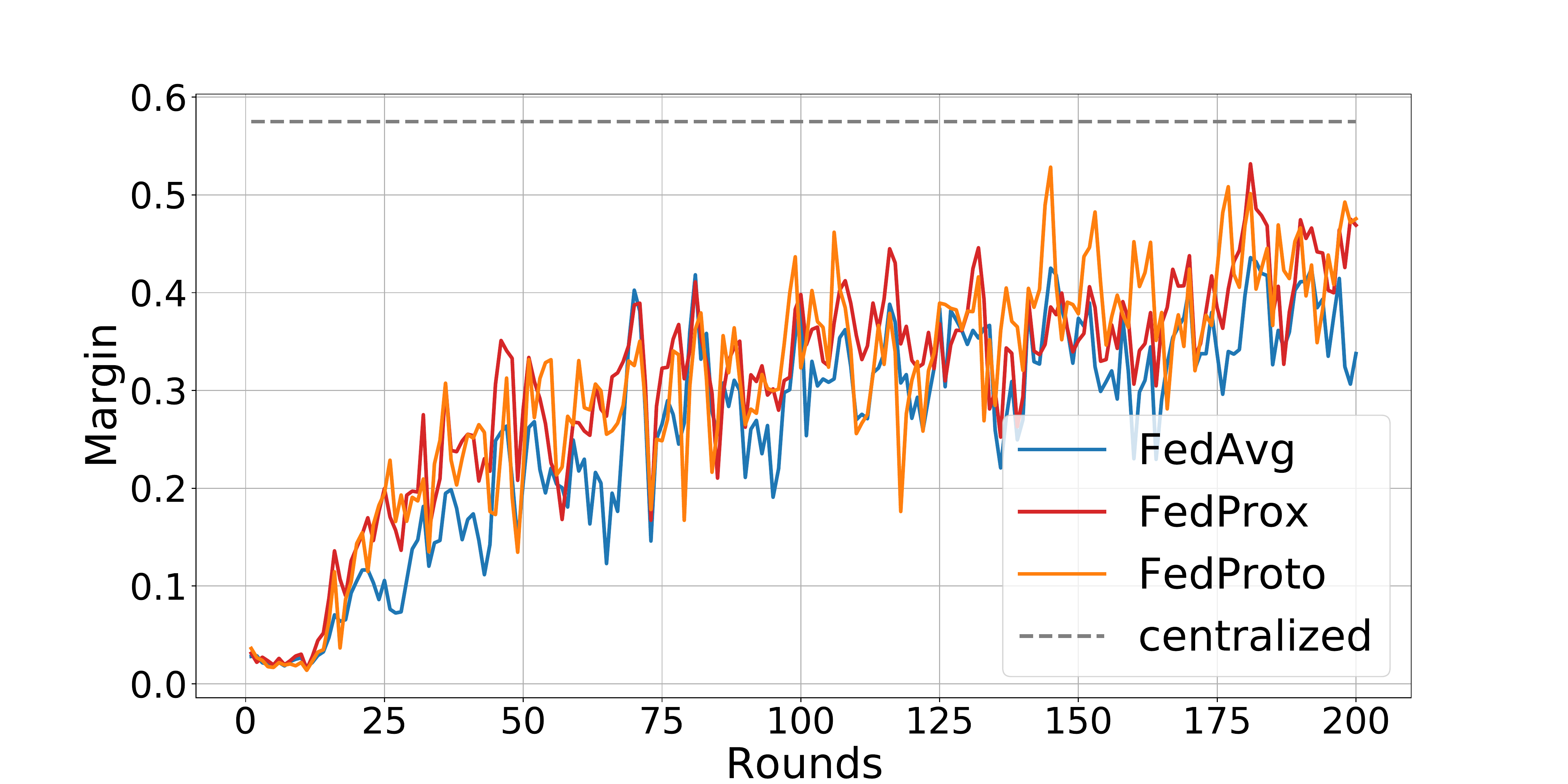} \\
   
   & \hspace{0.1cm} \tiny Rounds & & \hspace{0.1cm} \tiny Rounds \\
  
\end{tabular}
\caption{Per-round AMM ($\bar{\boldsymbol{\mu}}[t]$) values on classification datasets.}
\label{fig:margins_classification}
\end{figure}

\begin{table}[]
\centering
\caption[Caption]{Margin $\bar{\boldsymbol{\mu}}[T]$ of the final aggregate model and FFD ($\%$).}
\label{tab:margin_FFD}
\small
\setlength{\tabcolsep}{4pt}
\begin{tabular}{clcccc}
 & & Synthetic & MNIST & FEMNIST & CelebA \\\hline
 \multirow{4}{*}{\rotatebox[origin=c]{90}{$\bar{\boldsymbol{\mu}}[T]$}} & \multicolumn{1}{|l}{FedAvg} & $0.45$ & $2.38$ & $1.15$ & $0.34$ \\
& \multicolumn{1}{|l}{FedProx} & $0.48$ & $2.14$ & $1.14$ & $0.47$ \\
& \multicolumn{1}{|l}{FedProto} & $\mathbf{0.63}$ & $\mathbf{2.49}$ & $\mathbf{1.51}$ & $\mathbf{0.48}$ \\\cdashline{2-6}
& \multicolumn{1}{|l}{Centralized} & $0.60$ & $8.68$ & $3.61$ & $0.48$ \\\hline
\multirow{2}{*}{\rotatebox[origin=c]{90}{FFD}} &  \multicolumn{1}{|l}{FedProx} & $61.9$ & $5.9$ & $1.5$ & $4.5$  \\
& \multicolumn{1}{|l}{FedProto} & $\mathbf{64.5}$ & $\mathbf{7.7}$ & $\mathbf{5.7}$ & $\mathbf{7.8}$  \\\hline
\end{tabular}
\end{table}

\textbf{Aggregate Mean Margin (AMM).} 
To examine margin maximization properties of federated optimizers, we define a measure called aggregate mean margin (AMM) by
\begin{equation}
\label{eq:mu_bar}
    \bar{\boldsymbol{\mu}}[t] = \frac{1}{|\mathcal{C}|}\sum_{c \in \mathcal{C}}\mu(\mathbf{p}_{agg}^t [c], \mathbf{p}_{agg}^t).
\end{equation}
In Fig.~\ref{fig:margins_classification}, we show change of AMM for different optimizers and datasets during training in FL. FedProto achieves higher $\bar{\boldsymbol{\mu}}[t]$ compared to other optimizers. This is a direct consequence of a better shaping of latent representations with improved class-discrimination acting as regularizer for learning meaningful feature representations similar to centralized training. The AMM for the last round, $\bar{\boldsymbol{\mu}}[T]$, is reported in Table~\ref{tab:margin_FFD}. The results show a positive correlation between AMM and accuracy (given in Table~\ref{tab:classification_complete}) with Pearson's correlation coefficient $\rho=0.68$ (p-value $0.01$). 

\begin{figure}[tbp]
\centering
\setlength{\tabcolsep}{0.3pt} 
\renewcommand{\arraystretch}{0.4}
\centering
\footnotesize
\begin{tabular}{cccc}
  
   & \tiny \textbf{Synthetic} & & \tiny \textbf{MNIST} \\
  
   \tiny \rotatebox{90}{\quad \quad \quad \quad \quad MMD} &
   \includegraphics[trim=3.9cm 1.3cm 4.5cm 2.3cm, clip, width=0.485\linewidth]{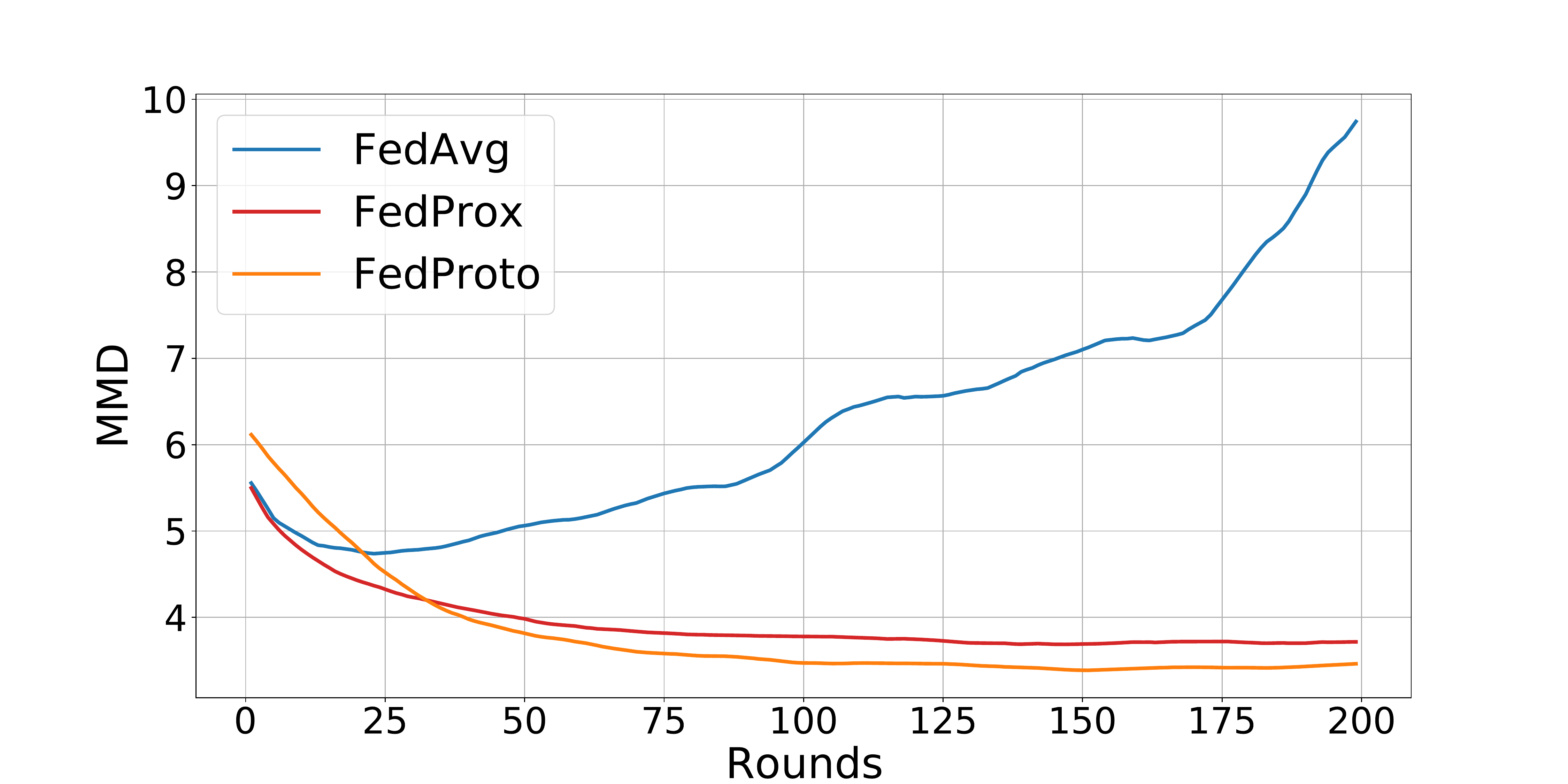} & &
   \hspace{0.1cm}\includegraphics[trim=3.9cm 1.3cm 4.5cm 2.3cm, clip, width=0.485\linewidth]{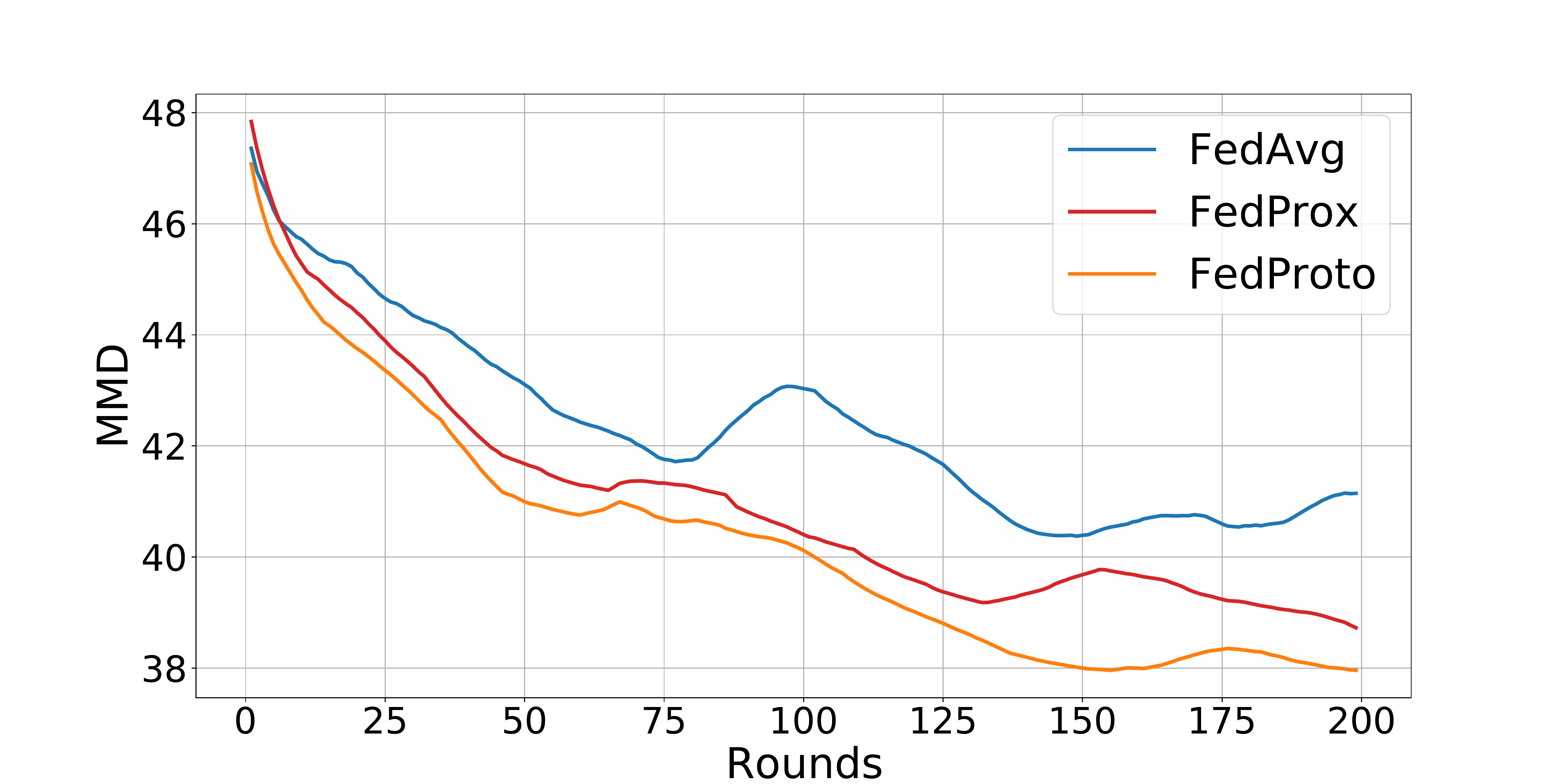} \\
   
   & \hspace{0.1cm} \tiny Rounds & & \hspace{0.2cm} \tiny Rounds \\
   & & & \\
   & \tiny \textbf{FEMNIST} & & \tiny \textbf{CelebA} \\
   
   \tiny \rotatebox{90}{\quad \quad \quad \quad \quad MMD} &
   \includegraphics[trim=3.9cm 1.3cm 4.5cm 2.3cm, clip, width=0.485\linewidth]{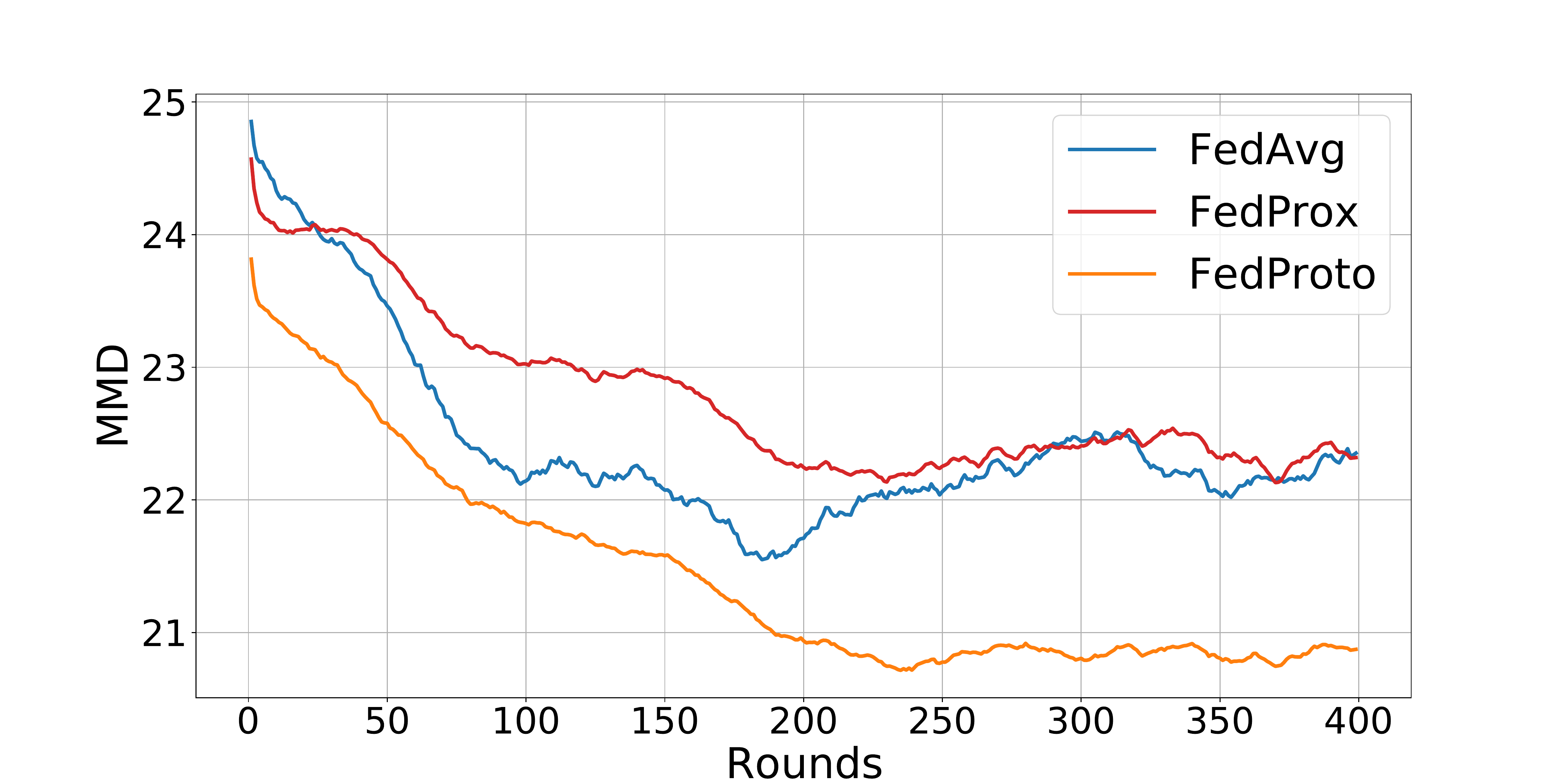} & &
   \includegraphics[trim=3.1cm 1.3cm 4.5cm 2.3cm, clip, width=0.497\linewidth]{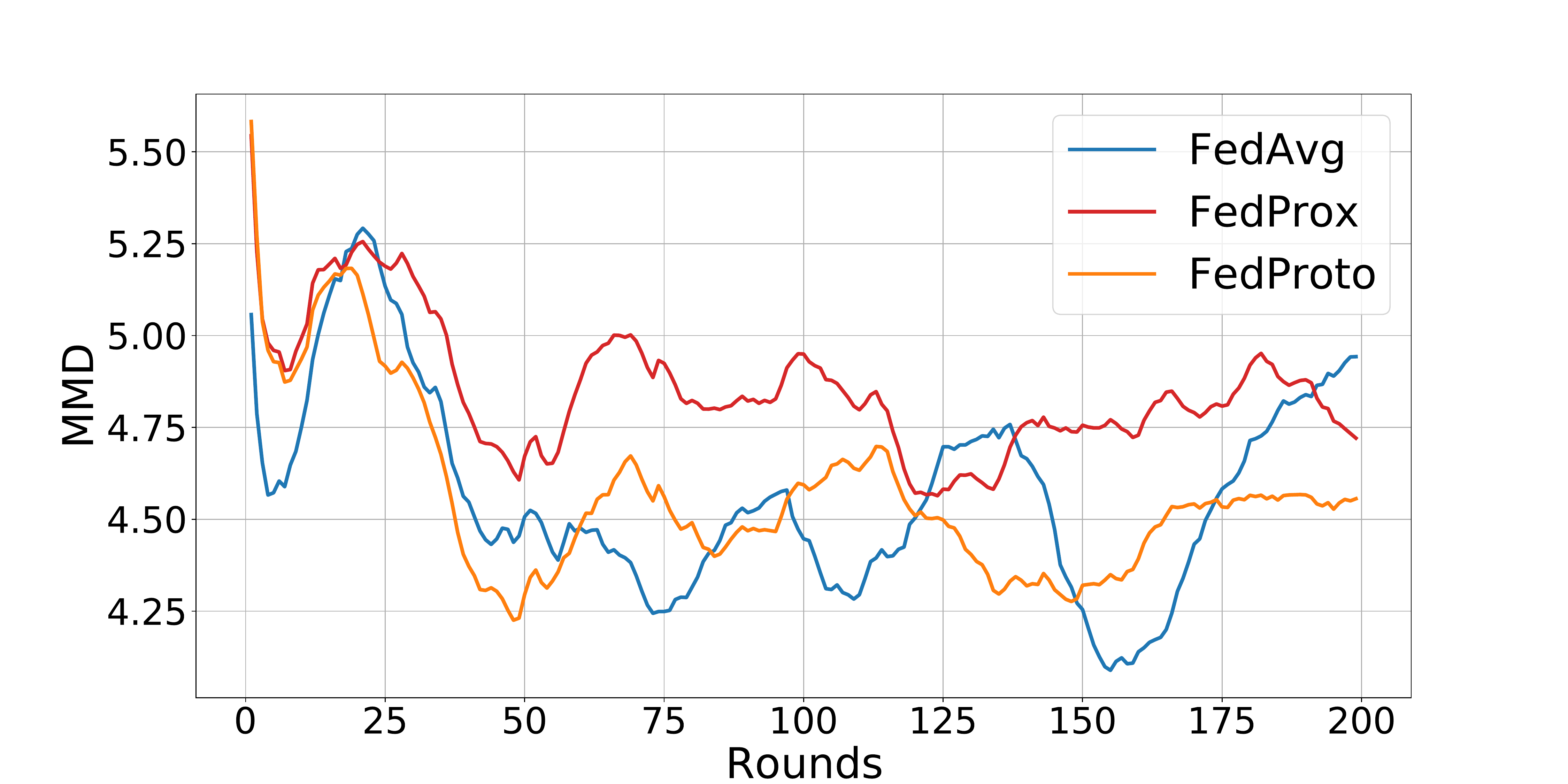} \\
   
   & \hspace{0.1cm} \tiny Rounds & & \hspace{0.2cm} \tiny Rounds \\
  
\end{tabular}
\caption{Per-round MMD~\eqref{eq:MMD} on classification datasets.}
\label{fig:MMD_classification}
\vspace{-0.25cm}
\end{figure}

\textbf{Federated Feature Discrepancy (FFD).} 
FFD is devised to analyze how feature distributions provided by a model $M_A$ trained with a federated optimizer $A$ are closer to those generated by centralized training of a model $M_C$, compared to a baseline optimizer $B$. To this end, we first compute distribution $P_k^t,\forall k$ of features provided by $M_A$. Second, we train a model $M_C$ on the same dataset without any distributed setting, and $Q$ denotes the distributions obtained from $M_C$. Then, we compute the average Maximum Mean Discrepancy (MMD) \cite{gretton2012kernel} between $P_k^t$  and $Q$ by
\begin{equation}
\label{eq:MMD}
    MMD_A^t = \frac{1}{|\mathcal{K}^t|} \sum _{k \in \mathcal{K}^t} MMD(P_k^t, Q).
\end{equation}
We define the FFD ($\%$) between $A$ and $B$ as the relative gain of $MMD_A^T$ over $MMD_B^T$ by
\begin{equation}
\label{eq:FFD}
FFD(A, B) = \frac{MMD_{B}^T - MMD_{A}^T}{MMD_{B}^T} \times 100.
\end{equation}
Since our interest is to give a comparison with respect to the baseline FedAvg, we set $A$ to FedProx or FedProto and $B$ to FedAvg.
The per-round MMD is shown in Fig.~\ref{fig:MMD_classification} and the final FFD values (the higher the better) are reported in the bottom part of Table~\ref{tab:margin_FFD}. Overall, we observe that distributions of features learned using FedProto are consistently more similar to those learned in centralized training than FedAvg. Last, we also note that FedProx can achieve some latent regularization thanks to the proximal term, however it is robustly surpassed by our proposed FedProto.

%


%
%

\begin{figure*}
\centering
\hspace{0.85cm}\includegraphics[trim=0cm 18.1cm 4cm 0cm, clip, width=0.9\linewidth]{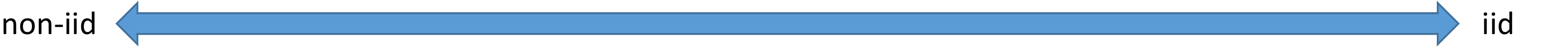}
\includegraphics[trim=4.85cm 0.2cm 4.75cm 1.6cm, clip, width=1\linewidth]{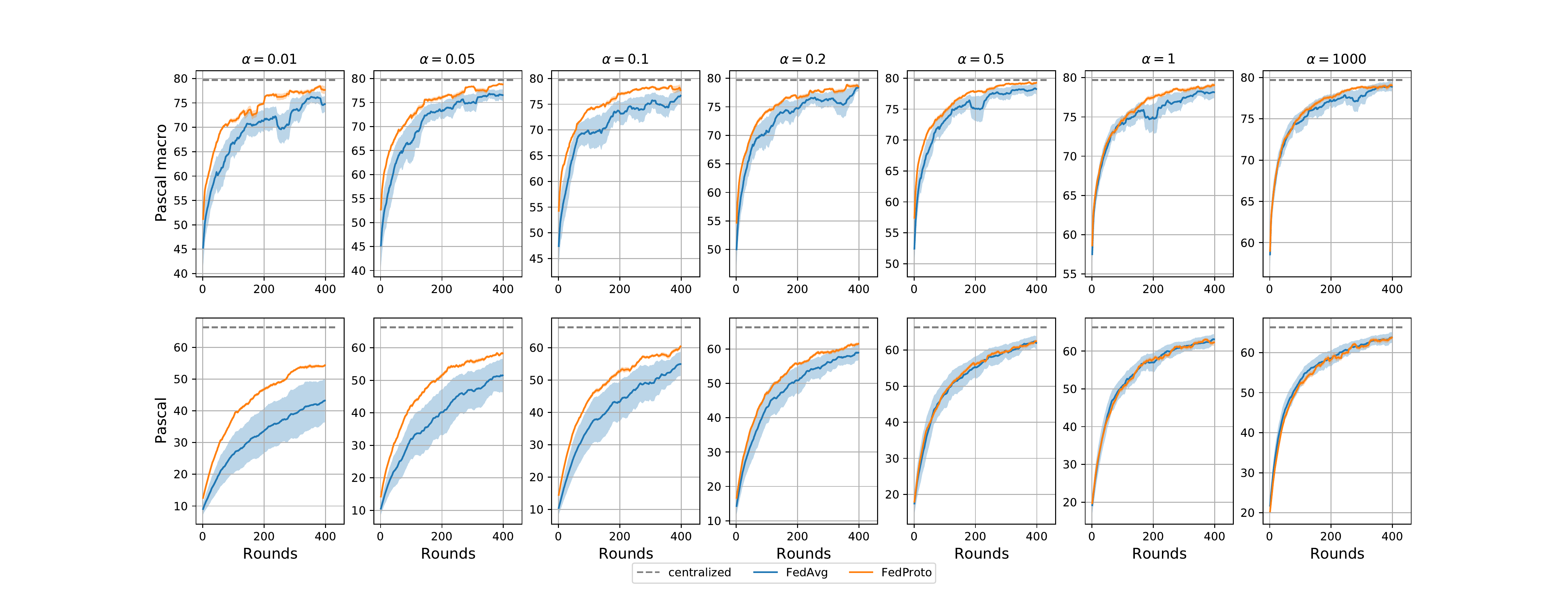}
\caption{Change of mIoU on segmentation data distributed using different $\alpha$ values. Evaluation is performed across $\delta \in \{ 0\%, 50\%, 80\% \}$ and a moving average window of $10\%$ rounds is applied. Solid and shaded lines represent mean and standard deviation.}
\label{fig:segmentation_complete}
\end{figure*}
\newcommand{\imgsize}{18.6mm}
\begin{figure*}[htbp]
\centering
\setlength{\tabcolsep}{0.5pt} 
\renewcommand{\arraystretch}{0.4}
\centering
\footnotesize
\begin{tabular}{cccccccccc}
    
   & &  \multicolumn{6}{c}{\includegraphics[trim=0cm 17.5cm 3.8cm 0cm, clip, width=0.5\linewidth]{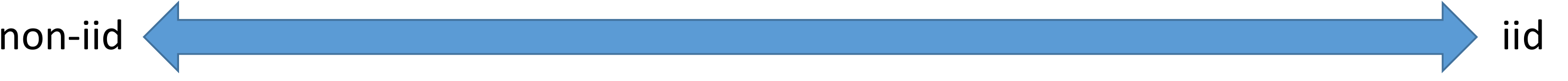}} & & \\
  
   & & \multicolumn{2}{c}{$\alpha=0.01$} & \multicolumn{2}{c}{$\alpha=0.1$} & \multicolumn{2}{c}{$\alpha=1$} & & \\
  
   \rotatebox{90}{\ \ Segmentation} &
   \includegraphics[width=\imgsize]{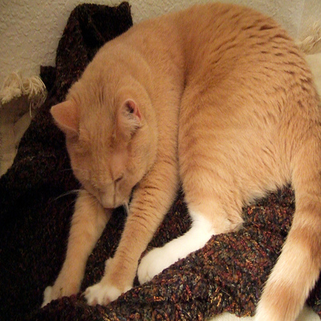} &
   \includegraphics[width=\imgsize]{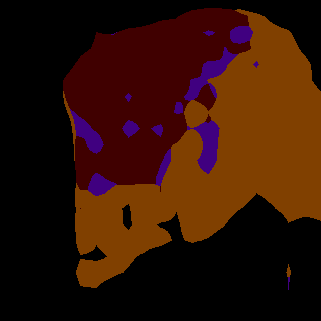} &
   \includegraphics[width=\imgsize]{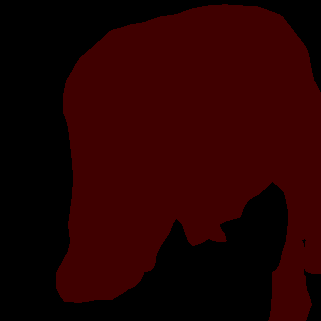} &   
   \includegraphics[width=\imgsize]{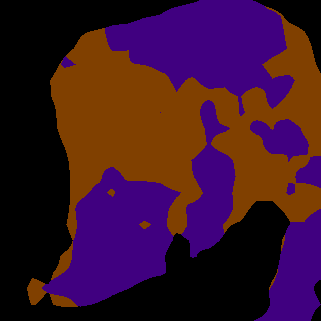} &   
   \includegraphics[width=\imgsize]{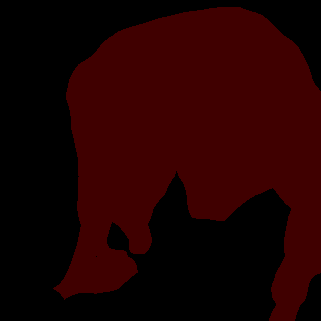} &   
   \includegraphics[width=\imgsize]{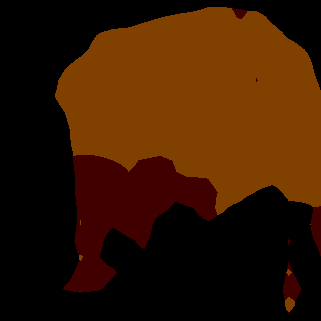} &   \includegraphics[width=\imgsize]{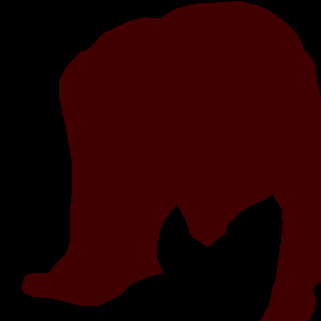} &
   \includegraphics[width=\imgsize]{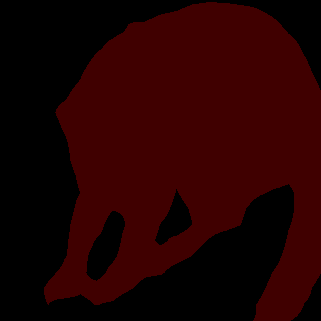} &
   \includegraphics[width=\imgsize]{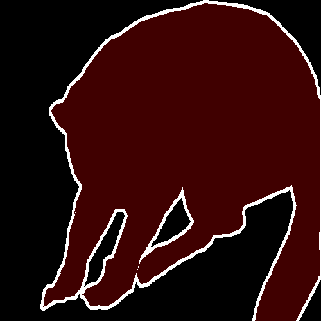} \\

   \rotatebox{90}{\ \ \ Soft.\ Entropy} &
   \includegraphics[width=\imgsize]{img/pascal_qualitative/img337_RGB.png} &
   \includegraphics[width=\imgsize]{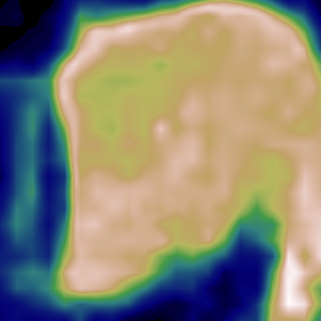} &
   \includegraphics[width=\imgsize]{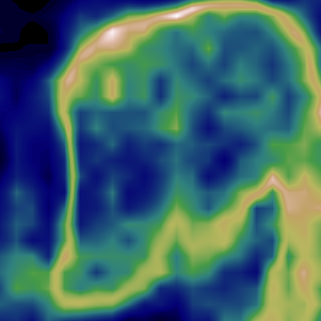} &   
   \includegraphics[width=\imgsize]{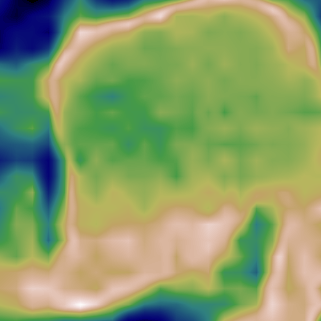} &   
   \includegraphics[width=\imgsize]{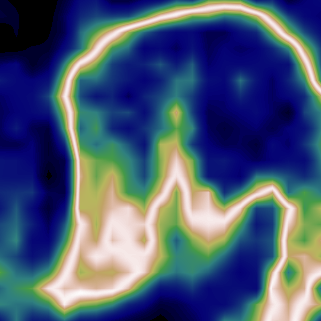} &   
   \includegraphics[width=\imgsize]{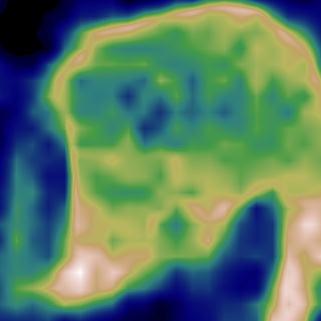} &   \includegraphics[width=\imgsize]{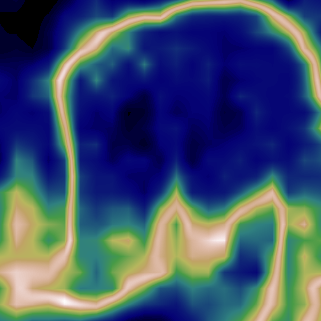} &
   \includegraphics[width=\imgsize]{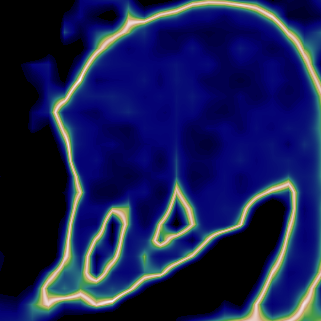} & 
   \hspace{-1cm} \multirow{2}{*}[17mm]{\includegraphics[width=4.7mm]{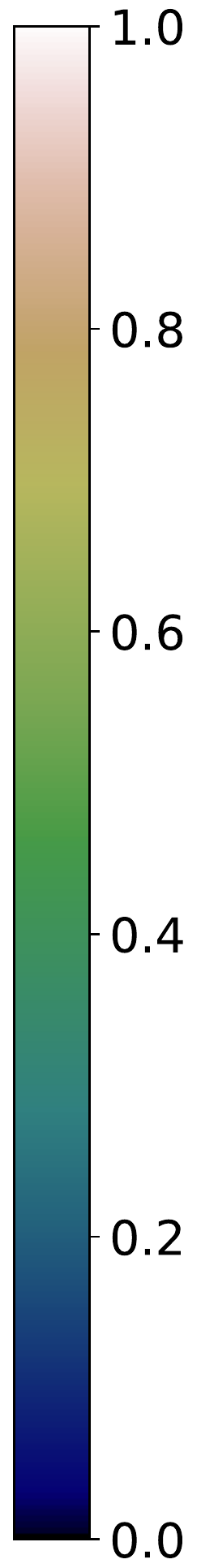}} \\
   
   \rotatebox{90}{\ \ Feat.\ Entropy} &
   \includegraphics[width=\imgsize]{img/pascal_qualitative/img337_RGB.png} &
   \includegraphics[width=\imgsize]{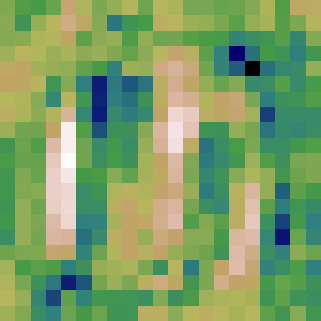} &
   \includegraphics[width=\imgsize]{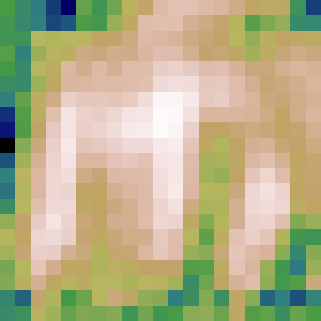} &   
   \includegraphics[width=\imgsize]{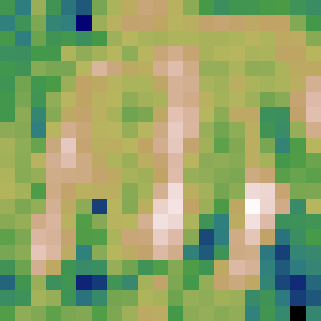} &   
   \includegraphics[width=\imgsize]{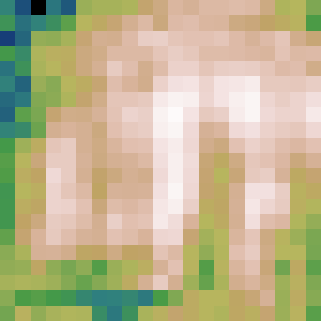} &   
   \includegraphics[width=\imgsize]{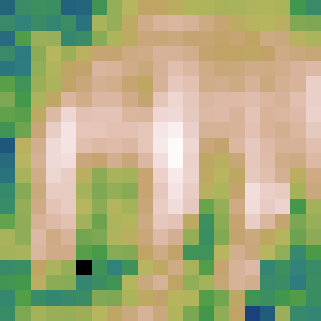} &   \includegraphics[width=\imgsize]{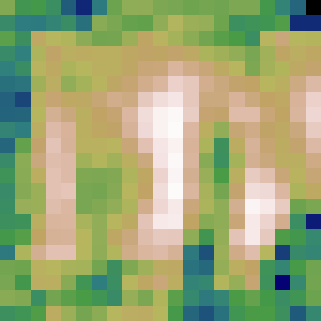} &
   \includegraphics[width=\imgsize]{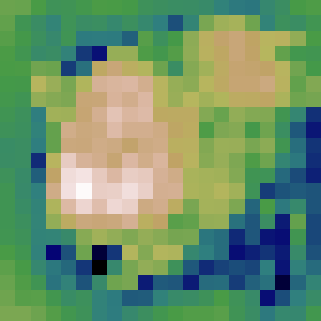} & \\
   
   
   \rotatebox{90}{\ \ Segmentation} &
   \includegraphics[width=\imgsize]{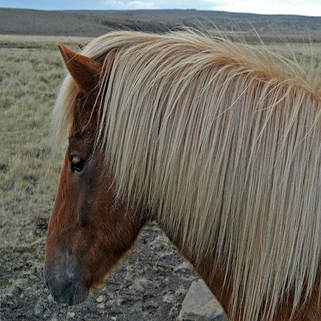} &
   \includegraphics[width=\imgsize]{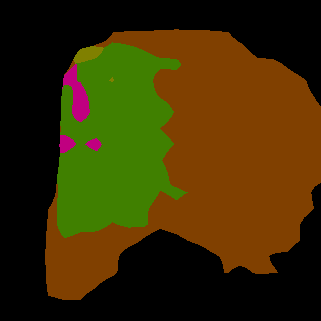} &
   \includegraphics[width=\imgsize]{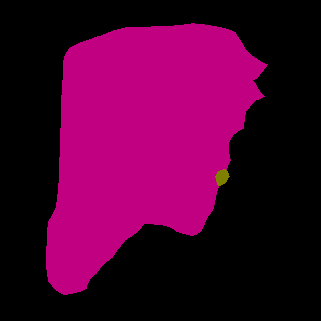} &   
   \includegraphics[width=\imgsize]{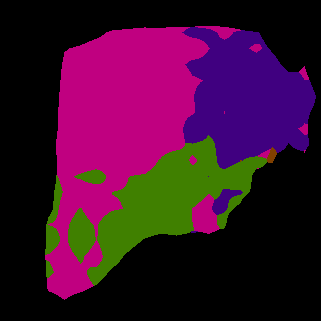} &   
   \includegraphics[width=\imgsize]{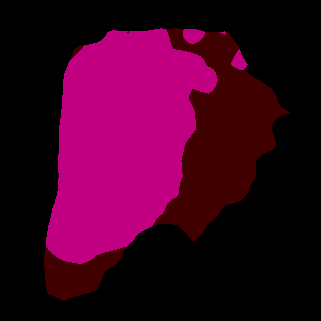} &   
   \includegraphics[width=\imgsize]{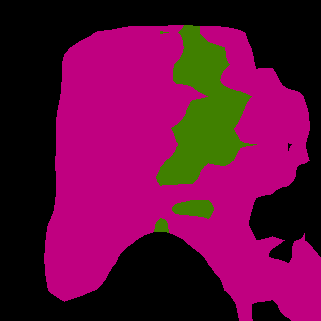} &   \includegraphics[width=\imgsize]{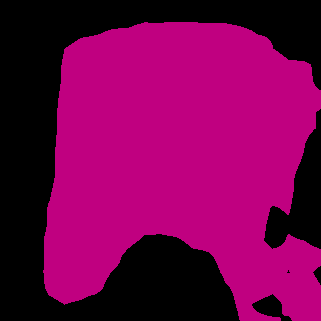} &
   \includegraphics[width=\imgsize]{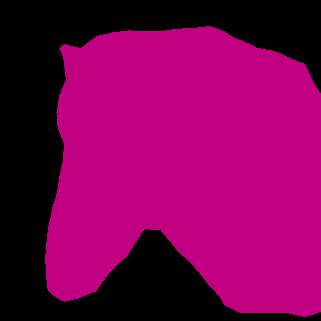} &
   \includegraphics[width=\imgsize]{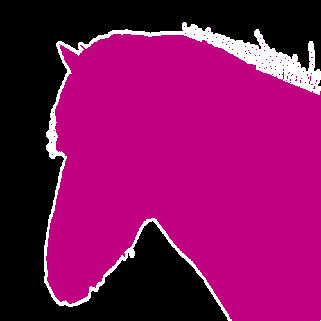} \\

   \rotatebox{90}{\ \ \ Soft.\ Entropy} &
   \includegraphics[width=\imgsize]{img/pascal_qualitative/img121_RGB.png} &
   \includegraphics[width=\imgsize]{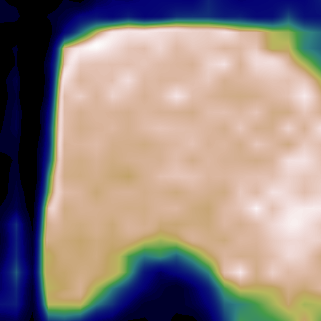} &
   \includegraphics[width=\imgsize]{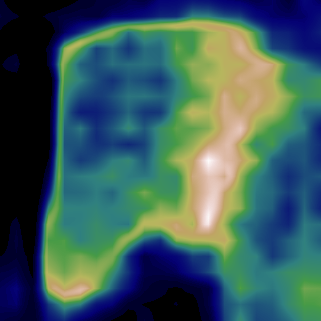} &   
   \includegraphics[width=\imgsize]{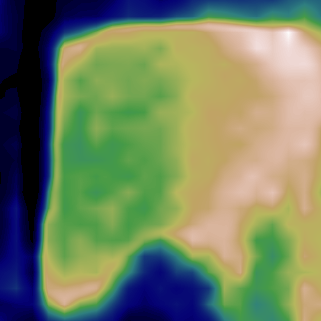} &   
   \includegraphics[width=\imgsize]{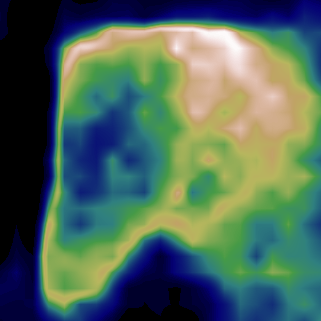} &   
   \includegraphics[width=\imgsize]{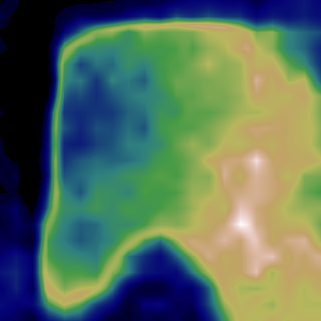} &   \includegraphics[width=\imgsize]{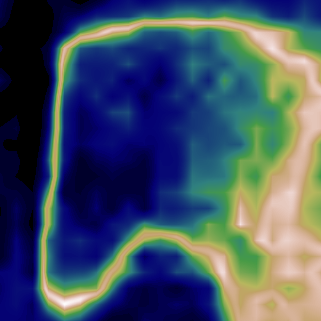} &
   \includegraphics[width=\imgsize]{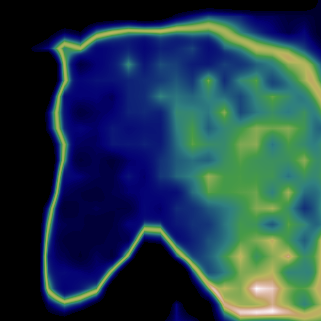} &
   \hspace{0cm} \multirow{2}{*}[17mm]{\includegraphics[trim=0cm 13cm 30.1cm 0cm, clip, width=18mm]{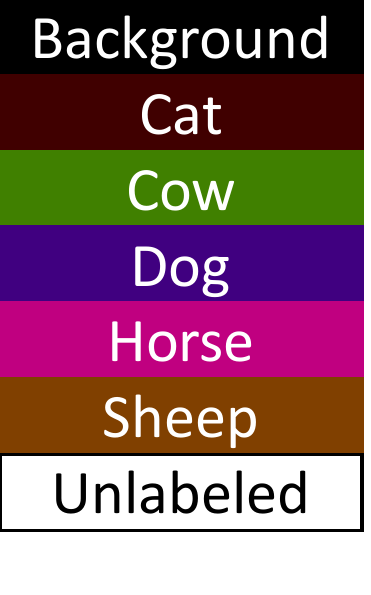}} \\
   
   \rotatebox{90}{\ \ Feat.\ Entropy} &\includegraphics[width=\imgsize]{img/pascal_qualitative/img121_RGB.png} &
   \includegraphics[width=\imgsize]{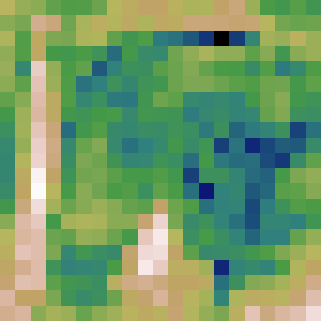} &
   \includegraphics[width=\imgsize]{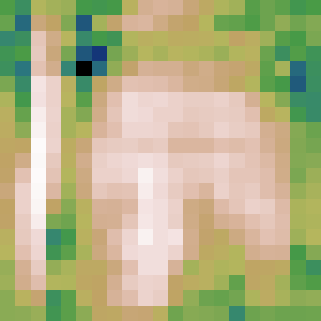} &   
   \includegraphics[width=\imgsize]{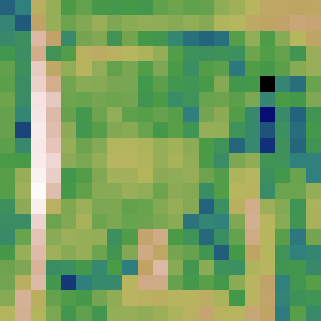} &   
   \includegraphics[width=\imgsize]{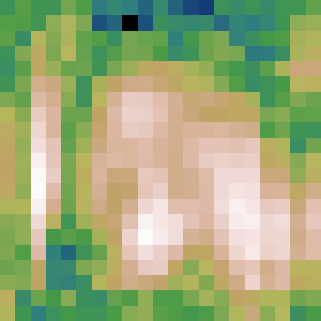} &   
   \includegraphics[width=\imgsize]{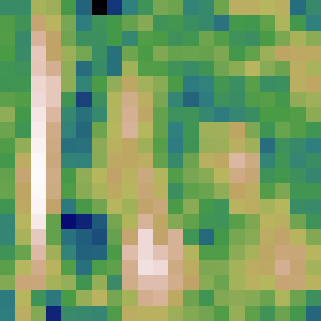} &   \includegraphics[width=\imgsize]{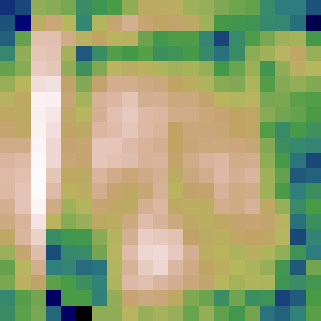} &
   \includegraphics[width=\imgsize]{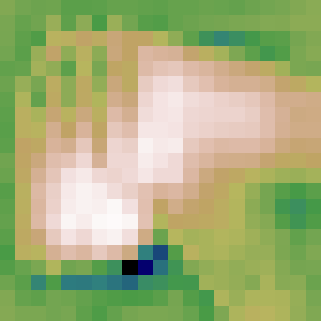} & \\
   
   & RGB &  FedAvg  &  FedProto  & FedAvg & FedProto & FedAvg  &  FedProto  &  Centralized & GT
  
\end{tabular}
\caption{Qualitative analyses of representations learned using FedAvg and FedProto for three non-i.i.d.\ to i.i.d.\ configurations. For each of the two sample images, we show output segmentation maps (rows 1 and 4), softmax-level entropy maps (rows 2 and 5), entropy maps of internal features (rows 3 and 6). As reference, centralized training results are shown at the second last column. \textit{Best viewed in colors}.}
\label{fig:qual_res}
\end{figure*}

\subsection{Federated Semantic Segmentation}
\label{subsec:results:segmentation}

We analyze our FedProto for federated semantic segmentation. Differently from image classification, segmentation task is more challenging as it involves dense predictions and highly class-imbalanced datasets. 
Altogether, these circumstances make aggregating local models even more severe.

We start by analyzing the effect of i.i.d.\ structure (i.i.d.-ness) of data on mIoU of federated segmentation models. For this purpose, we distribute two benchmark datasets among clients using the Dirichlet distribution with concentration parameter $\alpha$ (details are given in Sec.~\ref{sec:implementation} and in Suppl. Mat.). Then, we train models on distributed data using the baseline FedAvg and our FedProto. The results depicted on Fig.~\ref{fig:segmentation_complete} show the relationship between convergence of models and i.i.d.-ness of data.
Note that, as the non-i.i.d.-ness of distributed data increases by lower $\alpha$, data heterogeneity and client drift increase.
Our FedProto improves mIoU and robustness compared to FedAvg on every configuration, and especially on highly non-i.i.d.\ data, where class-conditional representations on certain remote clients could be non-reliable due to the non-i.i.d.\ partitioning (only few samples for particular classes observed on certain clients).

A qualitative analysis on segmentation and entropy maps of two sample images comparing the final aggregate models of FedAvg and FedProto on different data splitting configurations (\ie, setting ${\alpha \in \{0.01, 0.1, 1 \}}$) is reported in Fig.~\ref{fig:qual_res}.\\
%
\textbf{Segmentation maps:} Output segmentation maps (rows 1 and 4) improve when data are more i.i.d., better resembling segmentation maps produced by centralized training. FedProto significantly outperforms FedAvg for more non-i.i.d.\ data (${\alpha=0.01}$): the cat in first row and the horse in fourth row are correctly labeled and well-defined, whilst FedAvg labels them as a mixture of other animals. The ability to distinguish between class ambiguity is the direct consequence of a better latent space organization and regularization that FedProto achieves by maximizing prototype margin.

\textbf{Entropy maps:} Second, we report the entropy map of the softmax probabilities of the final model (rows 2 and 5): \ie,  ${H_S = H(M(\mathcal{W}^T; \mathcal{X}))}$, with $H(\cdot)$ being the pixel-wise Shannon entropy \cite{vu2019advent,wan2019information}. Low entropy (dark blue) indicates a peaked distribution which is the reflection of high confidence of the network on its prediction, and vice-versa.
Ideally, the entropy should be low for every pixel. However, as we can observe from centralized training, contours of objects and certain regions of the images (\eg, the mane of the horse in row 4) have high entropy due to uncertainty on the precise edge localization of the objects or due to intrinsic ambiguity with other classes (all considered animal classes have fur with similar pattern). With these considerations in mind, we observe how FedProto produces generally darker entropy maps than FedAvg, especially on non-i.i.d.\ data.
Last, we analyze the feature-level entropy maps upsampled to match input resolution (rows 3 and 6). To compute it, features $E(\mathcal{W}^T;\mathcal{X})$ are first normalized to $\hat{E}(\mathcal{W}^T;\mathcal{X})$, such that the sum over the channels at each low-resolution pixel location is 1 (\ie, in order for them to be considered as probability vectors), and then we define $H_E= H( \hat{E}(\mathcal{W}^T;\mathcal{X}) )$.
In this case, $H_E$ measures how representative a feature is at each pixel location. Ideally, features corresponding to the desired class should be well activated so that the decoder can discriminate between them and assign the correct label: this is the case of centralized training where features corresponding to (certain parts of) the object class are bright (\ie, high entropy denoting many activated patterns). We observe that FedProto produces a feature-level entropy map which is more similar to centralized training than the map produced by FedAvg (particularly visible for low $\alpha$ values). 




\section{Conclusion}
\label{sec:conclusion}

In this paper, we proposed FedProto, a distributed machine learning paradigm for vision models that can handle clients characterized by system and statistical heterogeneity. 
Previous approaches disregard internal representations to aggregate model weights. 
FedProto, instead, computes client deviations based on the inner class-conditional prototypical representations and uses them to drive federated optimization using an attentive mechanism. 
The experimental analyses 
demonstrated the effectiveness of our framework on both classification and segmentation datasets. In particular, we established a new benchmark on federated semantic segmentation task outlining a new research direction.

{\small
\bibliographystyle{ieee_fullname}
\bibliography{strings,biblio}
}


\clearpage

\onecolumn

\begin{center}
  \textbf{\large Prototype Guided Federated Learning of Visual Feature Representations \\ \vspace{0.1cm} \textit{Supplementary Material}}\\[.2cm]
  Umberto Michieli\textsuperscript{1,2}{\footnote[1]{Researched during internship at Samsung Research UK.}} \quad \quad \quad Mete Ozay\textsuperscript{1}\\[.1cm]
  \textsuperscript{1}Samsung Research UK \quad  \textsuperscript{2}University of Padova\\
  {\tt\small \{u.michieli, m.ozay\}@samsung.com}
\end{center}


\setcounter{equation}{0}
\setcounter{figure}{0}
\setcounter{table}{0}
\setcounter{page}{1}
\setcounter{section}{0}
\renewcommand{\theequation}{S\arabic{equation}}
\renewcommand{\thefigure}{S\arabic{figure}}
\renewcommand{\thetable}{S\arabic{table}}
\renewcommand{\thesection}{S\arabic{section}}

In this document, we present supporting material about task definitions, model designs along with their respective hyper-parameters, and real-world federated datasets analysed in the main paper. 
Furthermore, we report some additional ablation study on image classification and some qualitative results on the semantic segmentation task. 

\section{Federated Vision Datasets and Models}
\label{suppl:sec:datasets}
We evaluated our proposed FedProto on different computer vision tasks, models and real-world federated vision datasets. In this section, we explain the data generation process, data statistics and the models employed in our work. Most of our setups follow prior works \cite{ caldas2018leaf,li2020federated,mcmahan2017communication}. 

\subsection{Synthetic Data Classification Dataset}
First of all, we analyse our approach on highly non-i.i.d.\ synthetic data. For this purpose, we follow a similar setup to that proposed in \cite{li2020federated,shamir2014communication} with the addition of heterogeneity among clients.

\begin{itemize}

    \item \textbf{Synthetic:} For each client $k$, we generate samples $(\mathbf{x}_k,y_k) \in \mathcal{X}_k \times \mathcal{Y}_k$ according to the logistic regression model $y_k=\argmax \limits_{c\in \mathcal{C}}(\mathrm{softmax}(W\mathbf{x}_k+b))$, $\mathbf{x}_k\in \mathbb{R}^{60}$, $W\in \mathbb{R}^{10\times60}$, ${\mathbf{b}_k \in\mathbb{R}^{10}}$. In the model, we first initialize $W_k \sim \mathcal{N}(u_k,1)$, ${\mathbf{b}_k \sim \mathcal{N}(u_k,1)}$, $u_k\sim\mathcal{N}(0,\phi_1)$, ${\mathbf{x}_k \sim \mathcal{N}(\mathbf{v}_k,\boldsymbol{\Sigma})}$, where the covariance matrix $\boldsymbol{\Sigma}$ is diagonal with $\boldsymbol{\Sigma}_{j,j} = j^{-1.2}$. Then, each element of $\mathbf{v}_k$ is drawn from $\mathcal{N}(B_k,1)$, $B_k \sim \mathcal{N}(0,\phi_2)$. Hence, $\phi_1$ controls how much local models differ from each other, $\phi_2$ controls how much local data distribution at each client differs from that of other clients. For our simulations, we set $\phi_1=\phi_2=1$ being the most heterogeneous, yet challenging, scenario. Other analyses have been carried out in \cite{li2020federated}. Assuming that the underlying data generation model is agnostic, we employ a cascade of $2$ dense layers with $128$ and $256$ units respectively, followed by a softmax output layer.
\end{itemize}

\subsection{Real World Image Classification Datasets}
To further investigate accuracy on classification data, we explore real-world image classification datasets, inspired from \cite{caldas2018leaf,li2020federated, mcmahan2017communication}.

\begin{itemize}
    \item \textbf{MNIST:} It is a classification task of $28 \times 28px$ images containing handwritten digits 0-9 \cite{lecun1998gradient}. To simulate a heterogeneous (non-i.i.d.) setting, we distribute data among $1,000$ clients such that each client has samples of only two digits and the number of samples per client follow a power law distribution. To tackle this task we employ a simple custom network with $5\times5$ convolution layers (the first with $32$ channels, the second with $64$, each followed by $2\times2$ max pooling), a fully connected layer with $256$ units and ReLu activation, and a final softmax output layer.
    
    \item \textbf{FEMNIST:} It is a classification task of $28 \times 28px$ images containing 62-class handwritten character digits \cite{lecun1998gradient}. Following \cite{li2020federated}, to generate heterogeneity we first subsample 10 lower case characters (from 'a' to 'j') and we distribute only 5 classes to each client. The number of clients is $200$. To tackle this task we employ a simple custom network with $5\times5$ convolution layers (the first with $32$ channels, the second with $64$, each followed by $2\times2$ max pooling), a fully connected layer with $256$ units and ReLu activation, and a final softmax output layer.
    
    \item \textbf{CelebA:} Finally, we generate non-i.i.d. CelebA \cite{liu2015faceattributes} data (for classification of smiling faces), such that the underlying distribution of data for each user is consistent with the raw data. For this task, we use a $4$-layer CNN each with $32$ channels and followed by $2\times 2$ max pooling and ReLU activation, and a softmax layer.
    
    \end{itemize}

\subsection{Real World Semantic Segmentation Datasets}
Then, we investigate the accuracy on a dense prediction task, such as semantic segmentation. We propose a new benchmark for federated semantic segmentation employing the Pascal VOC2012 dataset \cite{everingham2010pascal} in two different flavours.

\begin{itemize}

\item \textbf{Pascal:} We use the standard Pascal VOC2012 \cite{everingham2010pascal} semantic segmentation dataset. We consider only images with one single class inside (in addition to the background) in order to mimic classification splits. There are a total of $20$ object-level classes (background excluded) and we distribute data to each client according to a Dirichlet distribution with concentration parameter $\alpha > 0$. Low values of $\alpha$ mean that dataset is highly non-i.i.d.\ among clients, and vice-versa for high $\alpha$ values \cite{hsu2019measuring,hsu2020federated}. The number of samples per client follow a power-law distribution with parameter $\gamma=3$. For this task, we use a DeepLab-V3+ \cite{chen2018encoder} architecture with MobileNet \cite{howard2017mobilenets} as the encoder pre-trained on ImageNet \cite{krizhevsky2012imagenet}. \\
To further elucidate on the data splitting mechanism, we report some dataset statistics for different values of $\alpha$ in Figure~\ref{suppl:fig:pascal_dataset_dirichlet}. The first column represents the distribution of the number of classes present on clients, while the second column represents the distribution of clients having a certain amount of classes. In the most non-i.i.d.\ case considered (\ie, $\alpha=0.01$), each client only experiences samples from a few classes, while they progressively observe more and more classes as the i.i.d.-ness improves. The third column of Figure~\ref{suppl:fig:pascal_dataset_dirichlet} illustrates populations drawn from the Dirichlet distribution with different concentration parameters. For visualization purposes, we restrict to $30$ randomly sampled clients and each color refers to a different class (color coding scheme reflects Pascal VOC2012 colormap). As expected, for low values of $\alpha$, the distributions are similar but not identical to a sort-and-partition approach in which each client only sees samples of one class (plus the background) \cite{hsu2019measuring}, since we have a highly imbalanced number of samples per each class and samples are distributed to clients according to a power-law distribution. Experimentally, we verified that a simple extremely i.i.d.\ sort-and-partition approach yields same results as the case with $\alpha=0.01$.

\item \textbf{Pascal macro:} We follow the same exact splitting described for standard Pascal VOC2012 with the only difference that classes are hierarchically grouped according to their semantic meaning into $5$ classes (background included). The coarser set of classes is derived from the notional taxonomy from \cite{everingham2010pascal,everingham2015pascal}. The map from $21$ to $5$ classes is:
\begin{itemize}
    \item Background: \textit{Background};
    \item Person: \textit{Person};
    \item Vehicles: \textit{Aeroplane}, \textit{Bicycle}, \textit{Boat}, \textit{Bus}, \textit{Car}, \textit{Motorbike}, Train;
    \item Household: \textit{Bottle}, \textit{Chair}, \textit{Dining Table}, \textit{Potted Plant}, \textit{Sofa}, \textit{TV/Monitor};
    \item Animals: \textit{Bird}, \textit{Cat}, \textit{Cow}, \textit{Dog}, \textit{Horse}, \textit{Sheep}.
\end{itemize}
\end{itemize}


    

\section{Hyperparameters and Implementation Details}
\label{suppl:sec:hyperparams}

For all the considered datasets, we randomly split the data on each local client into a training and a testing set with a $80/20$ ratio. We fix the number of selected clients to be $10$ for all experiments and most of the hyper-parameters has been reported in Table $1$ of the main paper. 
Unless otherwise stated, we assume that FedAvg does not tolerate partial local solutions (\ie, dropped clients are not aggregated), while FedProto and FedProx do tolerate them. For Synthetic, MNIST, FEMNIST, and CelebA, we set the proximal loss term of FedProx following the guidelines of \cite{li2020federated}, and the best accuracy is obtained respectively for: $0.1$, $1$, $1$, $0.01$.

We developed our framework in Tensorflow \cite{abadi2016tensorflow}. 
We simulate the federated learning setup ($1$ server and $|\mathcal{K}|$ clients) on a single NVIDIA$^{\tiny{\textregistered}}$ GeForce RTX 2080 Ti GPU with 2 Intel$^{\tiny{\textregistered}}$ Xeon$^{\tiny{\textregistered}}$ Gold 5220 CPU at 2.20GHz.

\newcommand{\imgsizee}{53.5mm}
\begin{figure*}[h!]
\centering
\setlength{\tabcolsep}{1pt} 
\renewcommand{\arraystretch}{0}
\begin{tabular}{cccccccccc}
  
  & \textbf{Distribution of \# classes} & \textbf{Distribution of \# clients} & \textbf{Per-client class distribution} \\

   \rotatebox{90}{\quad \quad \ $\boldsymbol{\alpha=0.01}$} \hspace{0.05cm} &
   \rotatebox{90}{\footnotesize \quad \quad \quad \ \ \ \# classes}
   \includegraphics[trim=4cm 1cm 3.4cm 2.4cm, clip, width=\imgsizee]{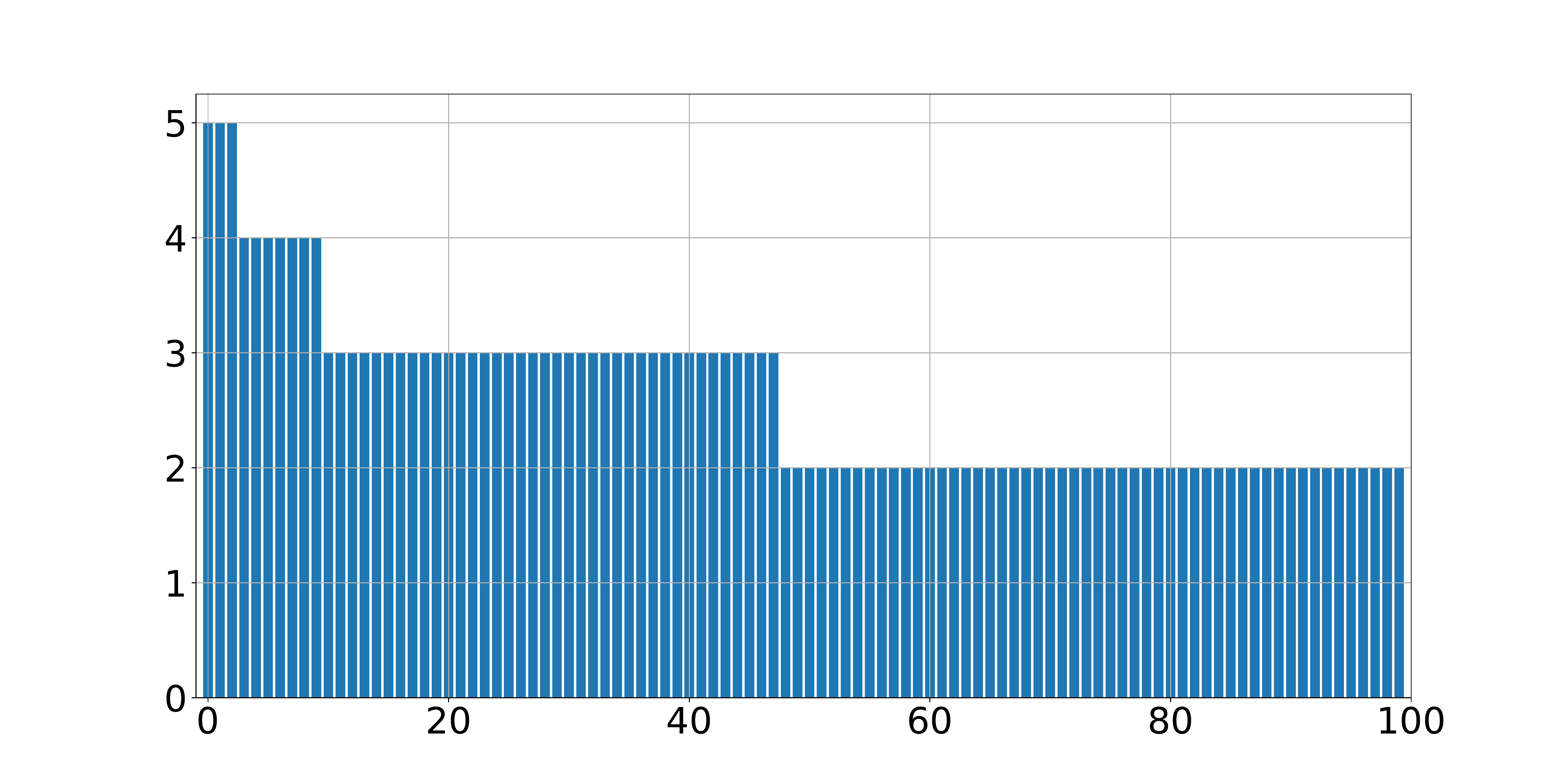} \hspace{0.03cm} &
   \rotatebox{90}{\footnotesize \quad \quad \quad \ \ \ \# clients}
   \includegraphics[trim=4cm 1cm 3.4cm 2.4cm, clip, width=\imgsizee]{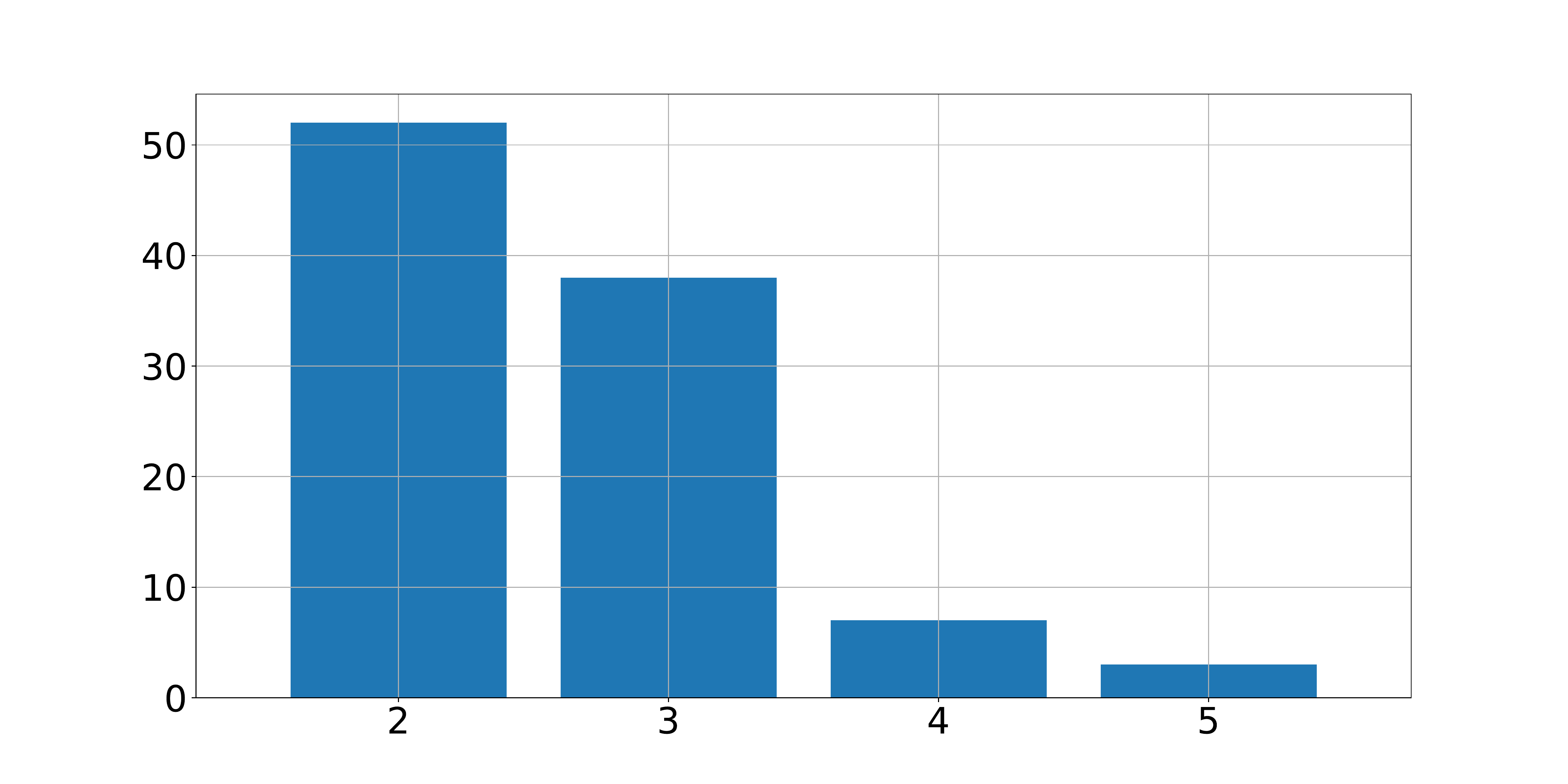} \hspace{0.03cm} &
   \rotatebox{90}{\footnotesize \quad \quad \quad \ \ client ID}
   \includegraphics[trim=4cm 1cm 3.4cm 2.4cm, clip, width=\imgsizee]{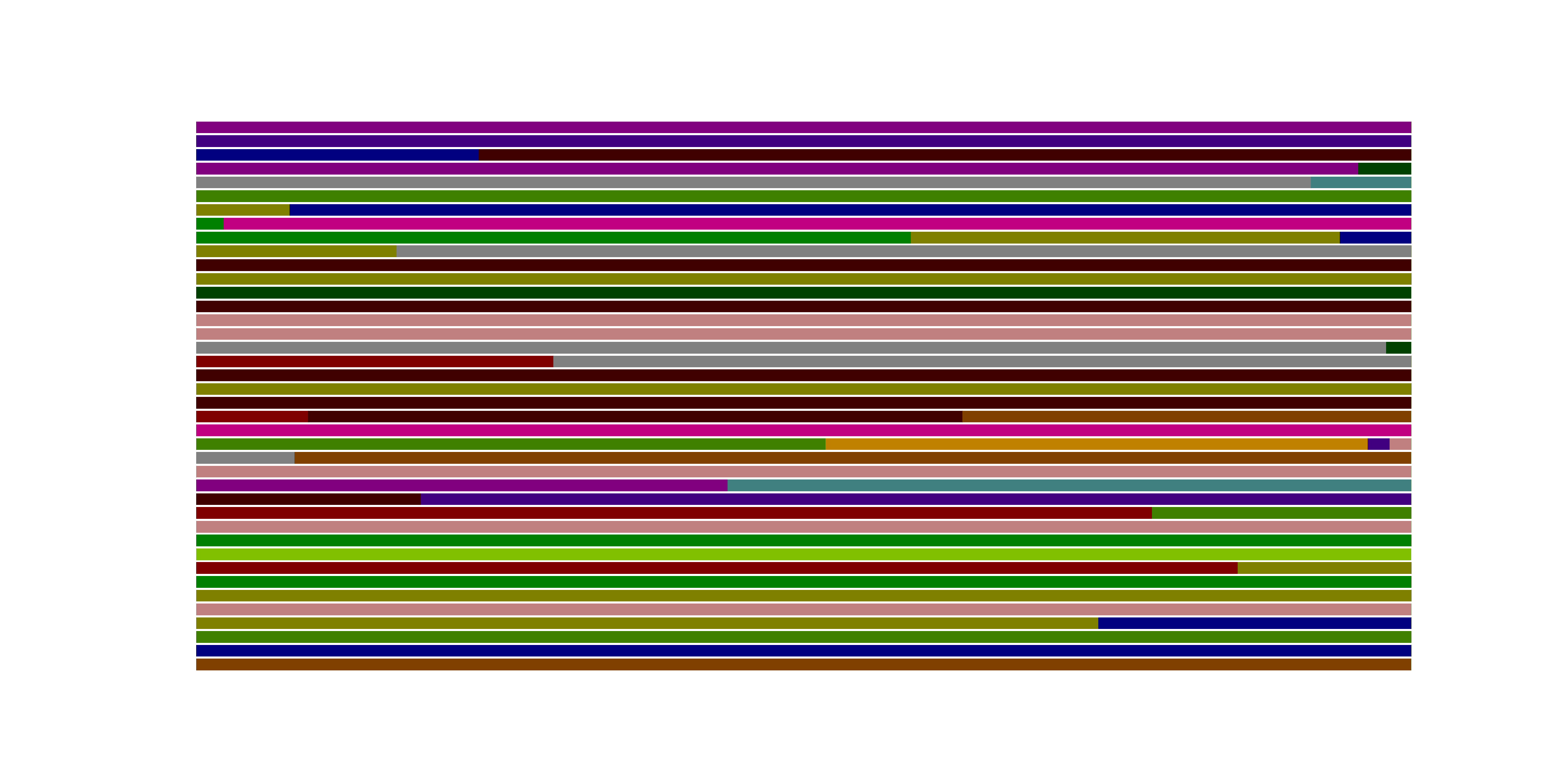} \\
   
   \rotatebox{90}{\quad \quad \ $\boldsymbol{\alpha=0.05}$} \hspace{0.05cm} &
   \rotatebox{90}{\footnotesize \quad \quad \quad \ \ \ \# classes}
   \includegraphics[trim=4cm 1cm 3.4cm 2.4cm, clip, width=\imgsizee]{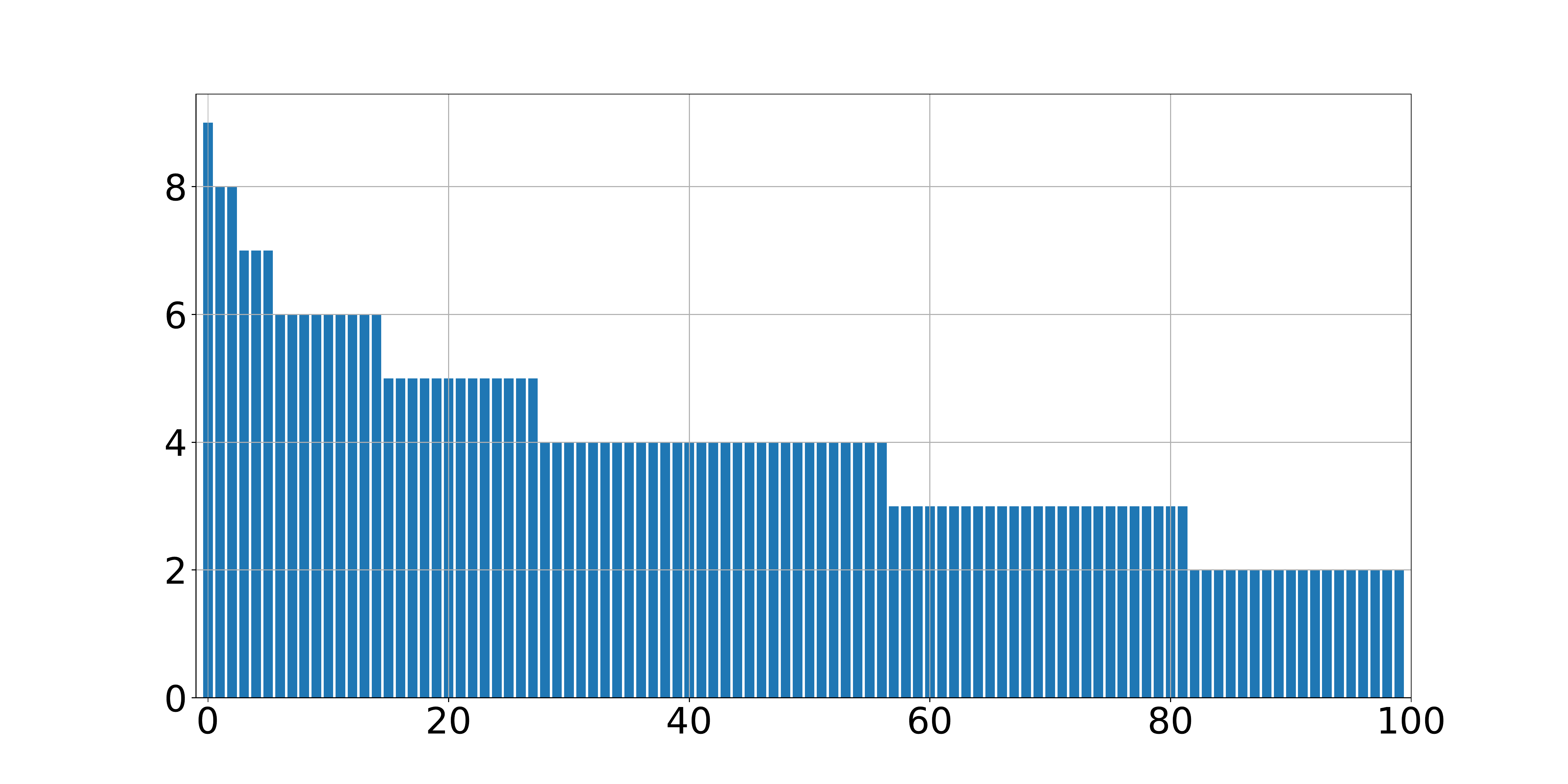} \hspace{0.03cm} &
   \rotatebox{90}{\footnotesize \quad \quad \quad \ \ \ \# clients}
   \includegraphics[trim=4cm 1cm 3.4cm 2.4cm, clip, width=\imgsizee]{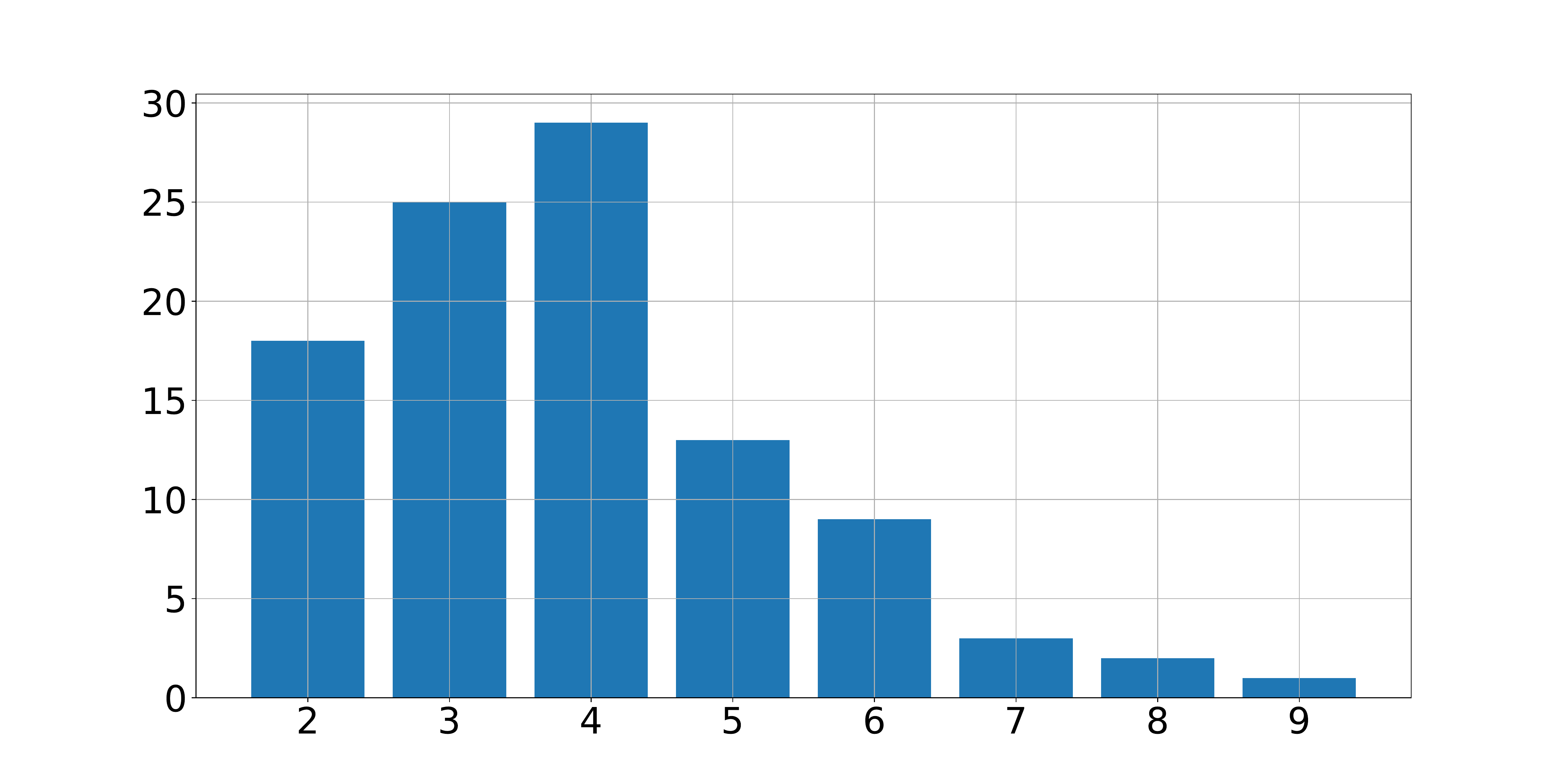} \hspace{0.03cm} &
   \rotatebox{90}{\footnotesize \quad \quad \quad \ \ client ID}
   \includegraphics[trim=4cm 1cm 3.4cm 2.4cm, clip, width=\imgsizee]{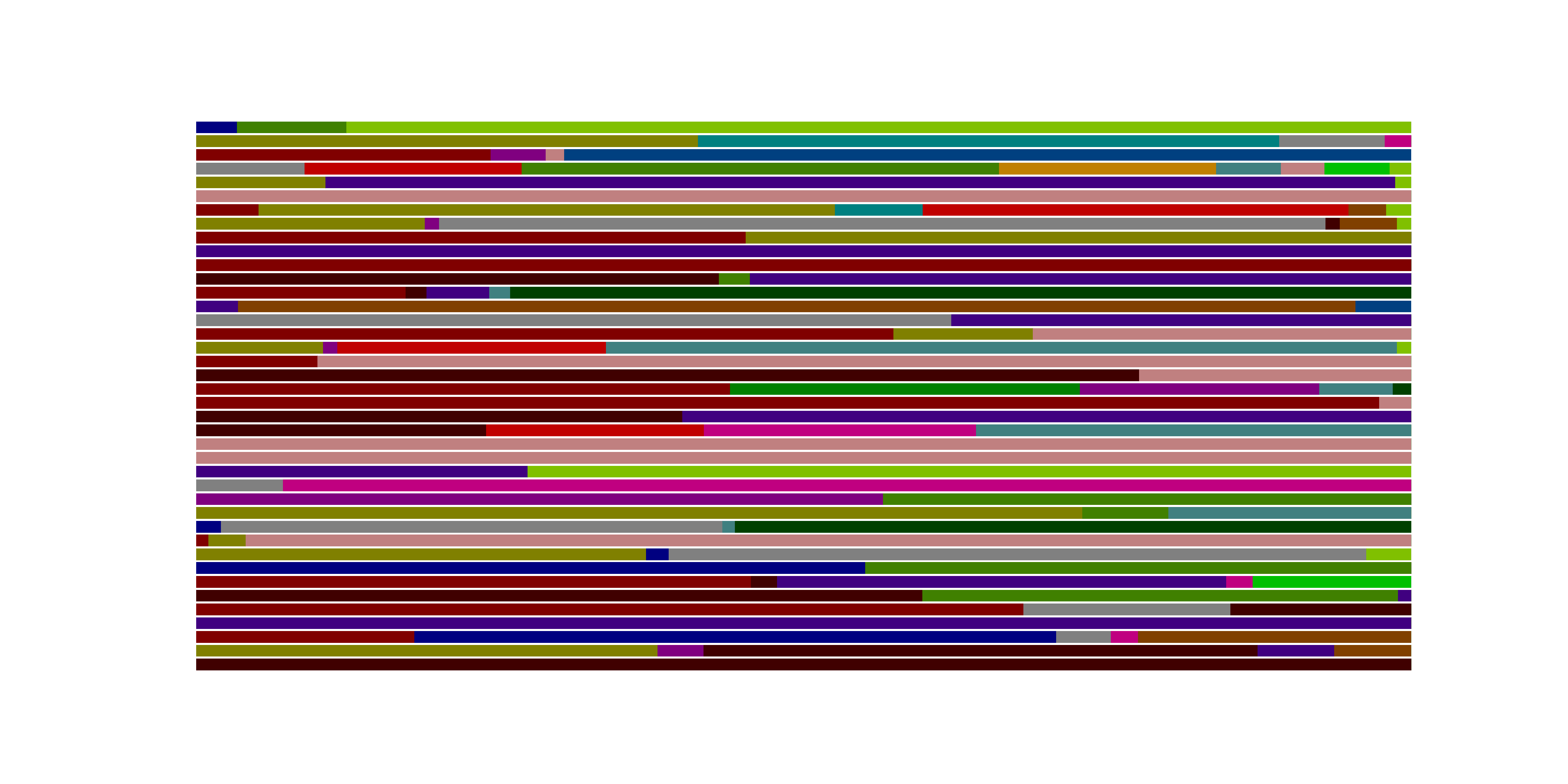} \\
   
   \rotatebox{90}{\quad \quad \ $\boldsymbol{\alpha=0.1}$} \hspace{0.05cm} &
   \rotatebox{90}{\footnotesize \quad \quad \quad \ \ \ \# classes}
   \includegraphics[trim=4cm 1cm 3.4cm 2.4cm, clip, width=\imgsizee]{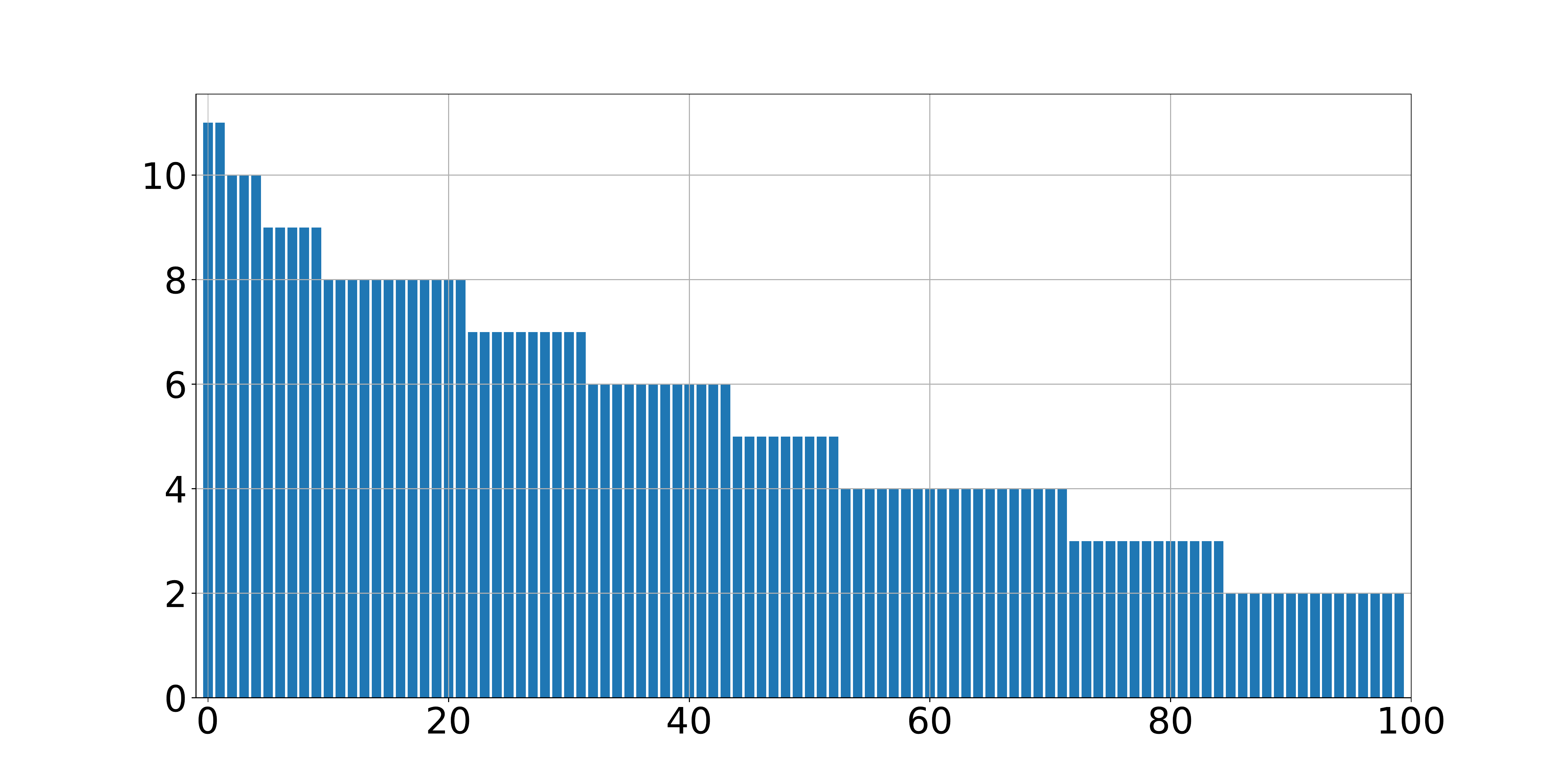} \hspace{0.03cm} &
   \rotatebox{90}{\footnotesize \quad \quad \quad \ \ \ \# clients}
   \includegraphics[trim=4cm 1cm 3.4cm 2.4cm, clip, width=\imgsizee]{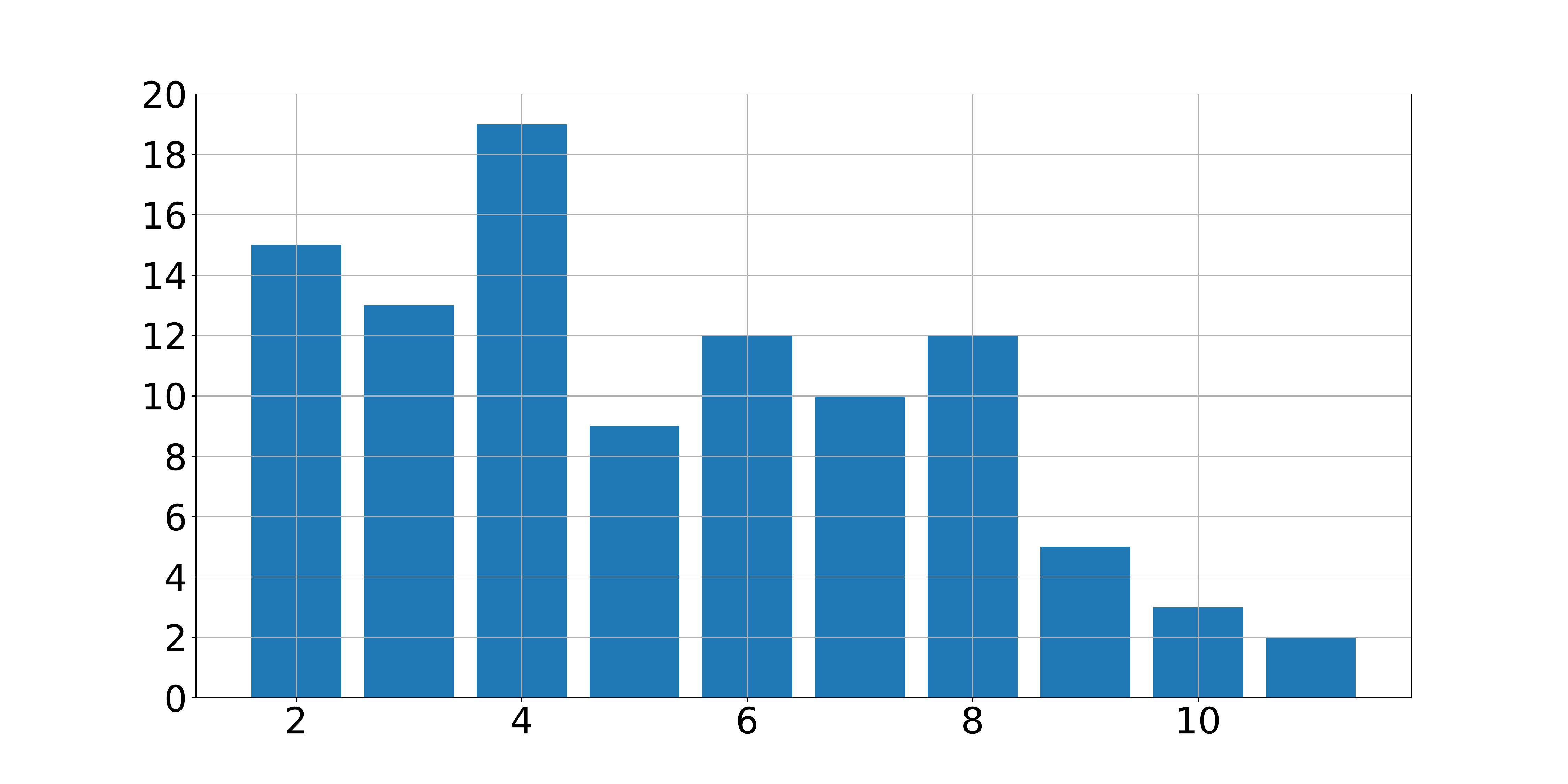} \hspace{0.03cm} &
   \rotatebox{90}{\footnotesize \quad \quad \quad \ \ client ID}
   \includegraphics[trim=4cm 1cm 3.4cm 2.4cm, clip, width=\imgsizee]{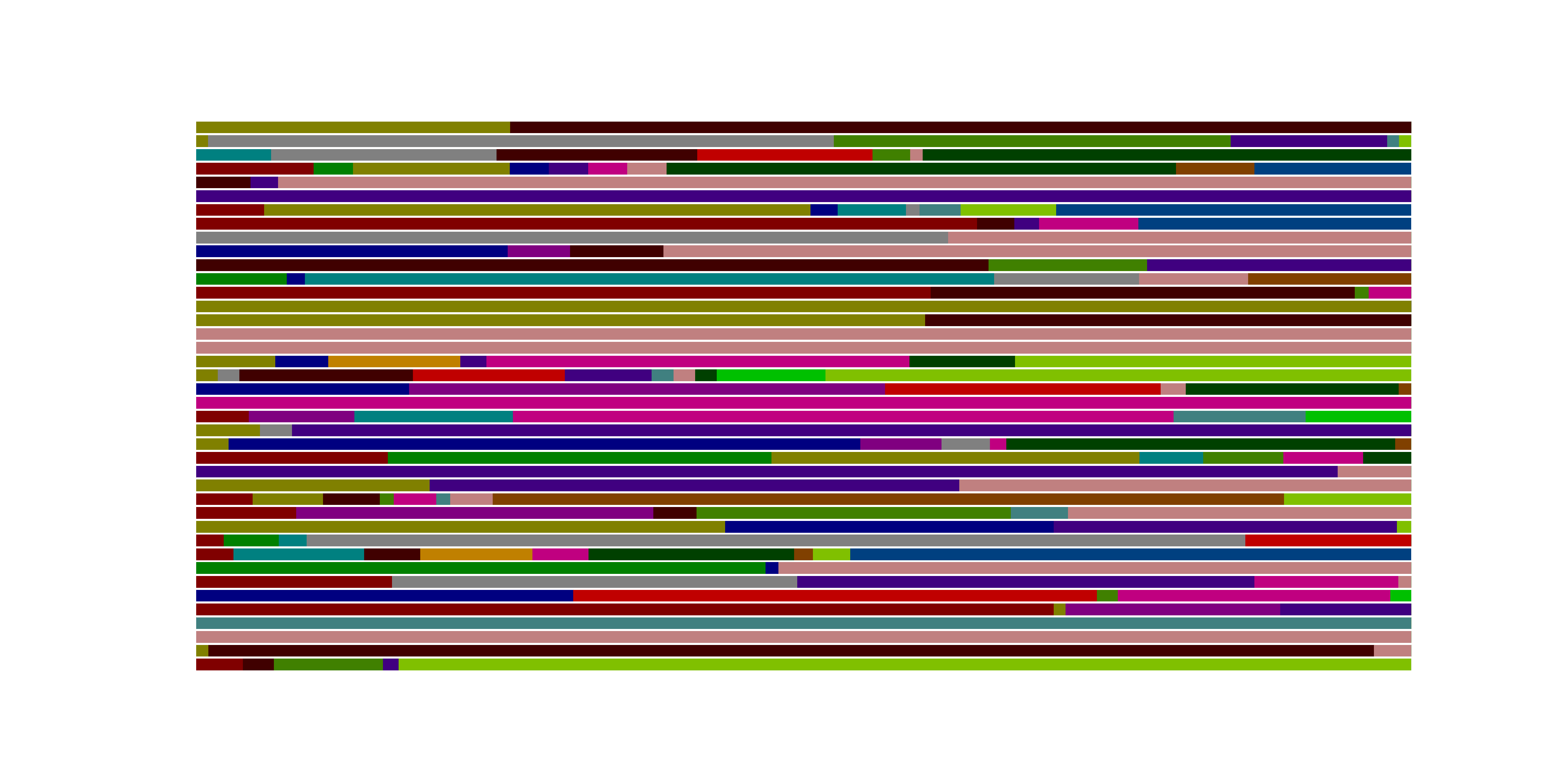} \\

      \rotatebox{90}{\quad \quad \ $\boldsymbol{\alpha=0.2}$} \hspace{0.05cm} &
   \rotatebox{90}{\footnotesize \quad \quad \quad \ \ \ \# classes}
   \includegraphics[trim=4cm 1cm 3.4cm 2.4cm, clip, width=\imgsizee]{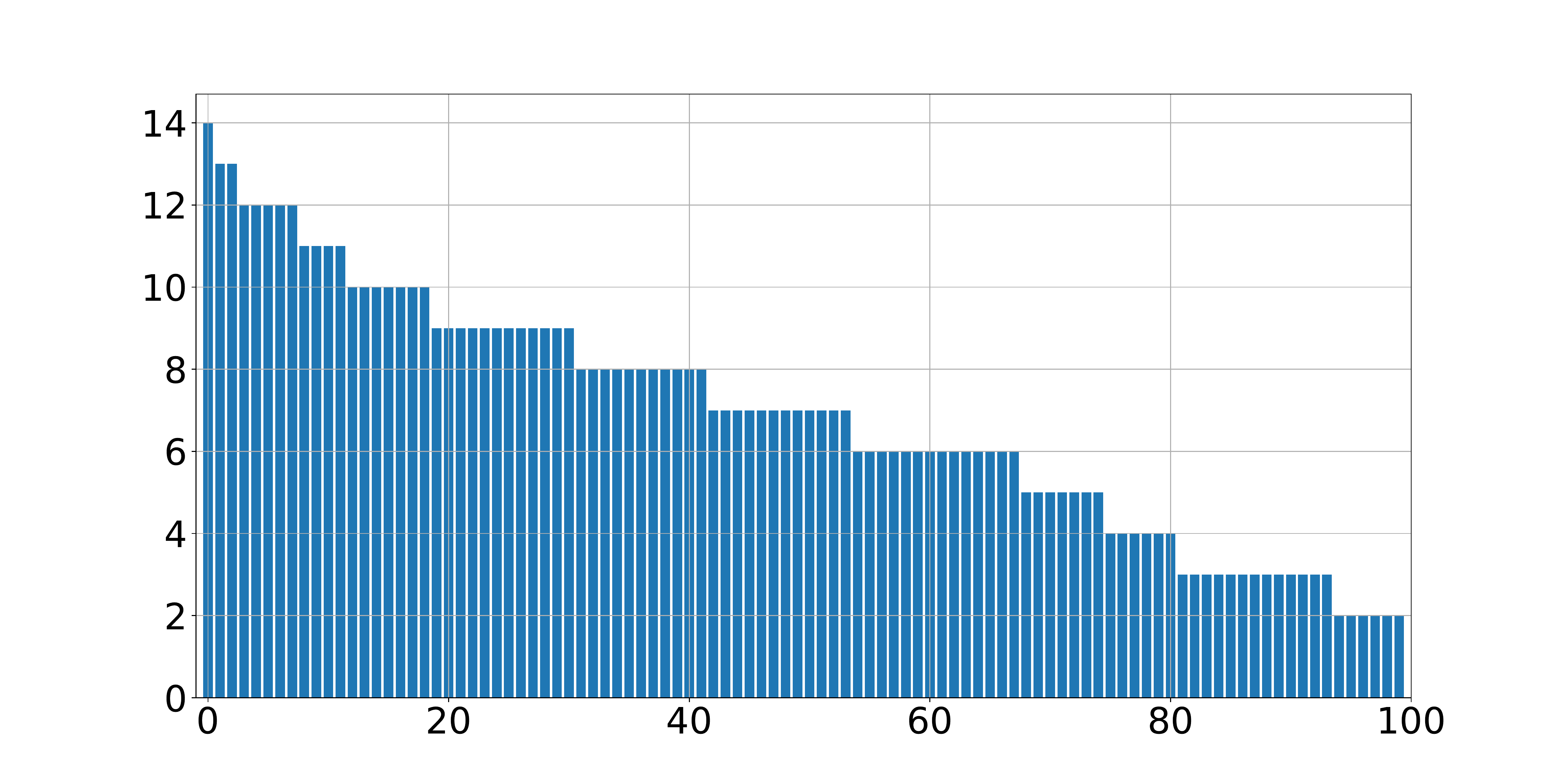} \hspace{0.03cm} &
   \rotatebox{90}{\footnotesize \quad \quad \quad \ \ \ \# clients}
   \includegraphics[trim=4cm 1cm 3.4cm 2.4cm, clip, width=\imgsizee]{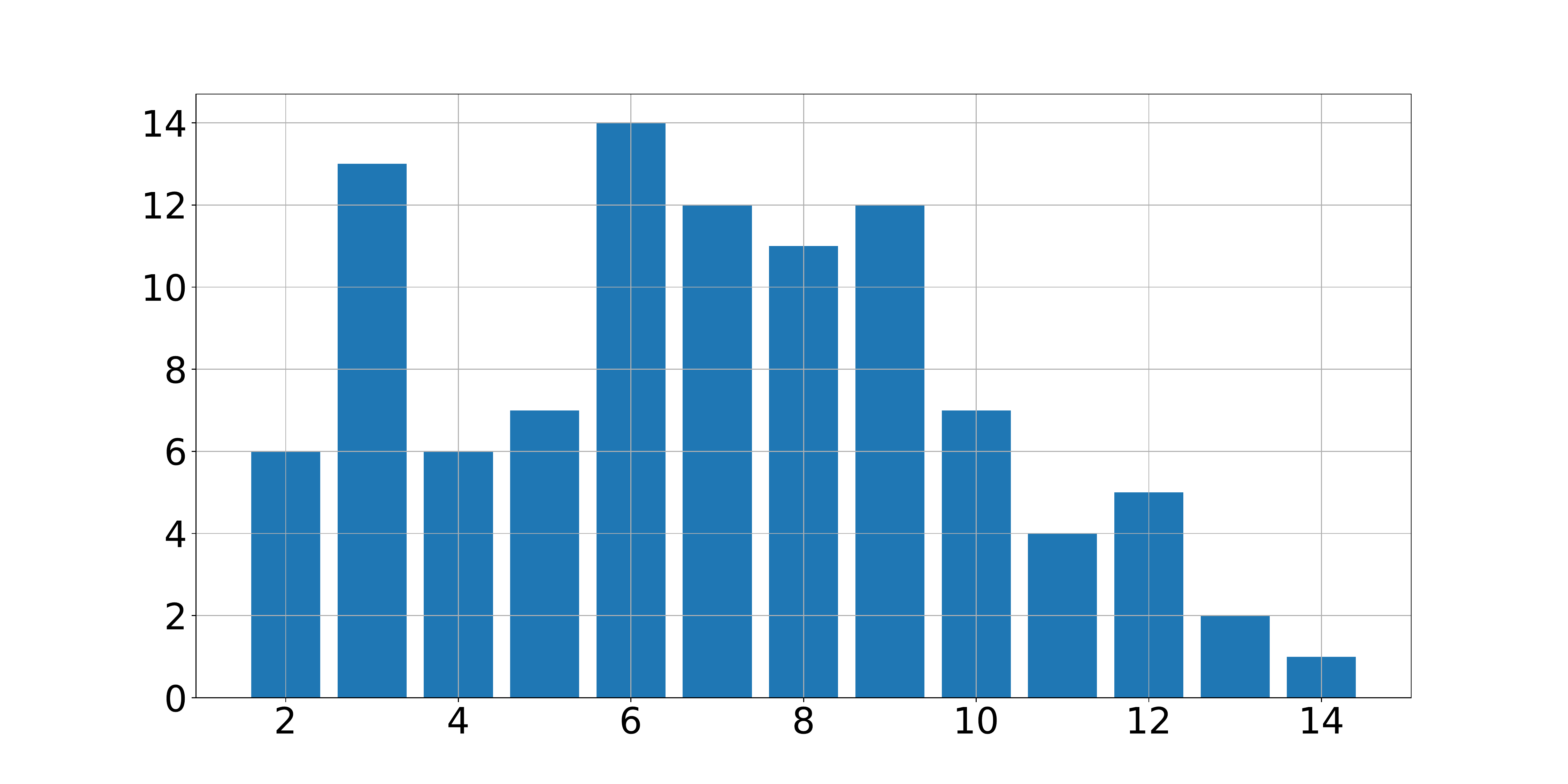} \hspace{0.03cm} &
   \rotatebox{90}{\footnotesize \quad \quad \quad \ \ client ID}
   \includegraphics[trim=4cm 1cm 3.4cm 2.4cm, clip, width=\imgsizee]{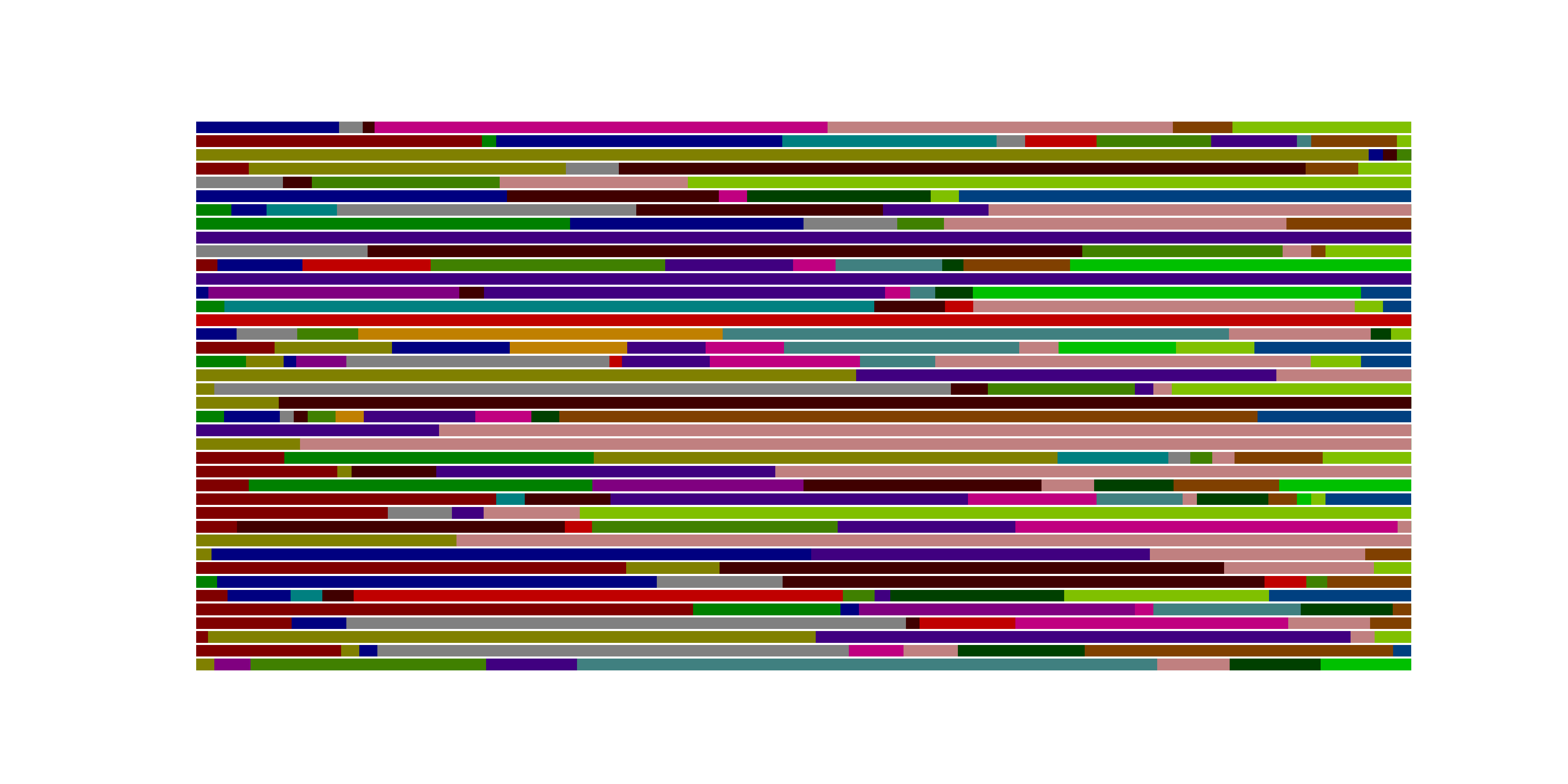} \\
   
      \rotatebox{90}{\quad \quad \ $\boldsymbol{\alpha=0.5}$} \hspace{0.05cm} &
   \rotatebox{90}{\footnotesize \quad \quad \quad \ \ \ \# classes}
   \includegraphics[trim=4cm 1cm 3.4cm 2.4cm, clip, width=\imgsizee]{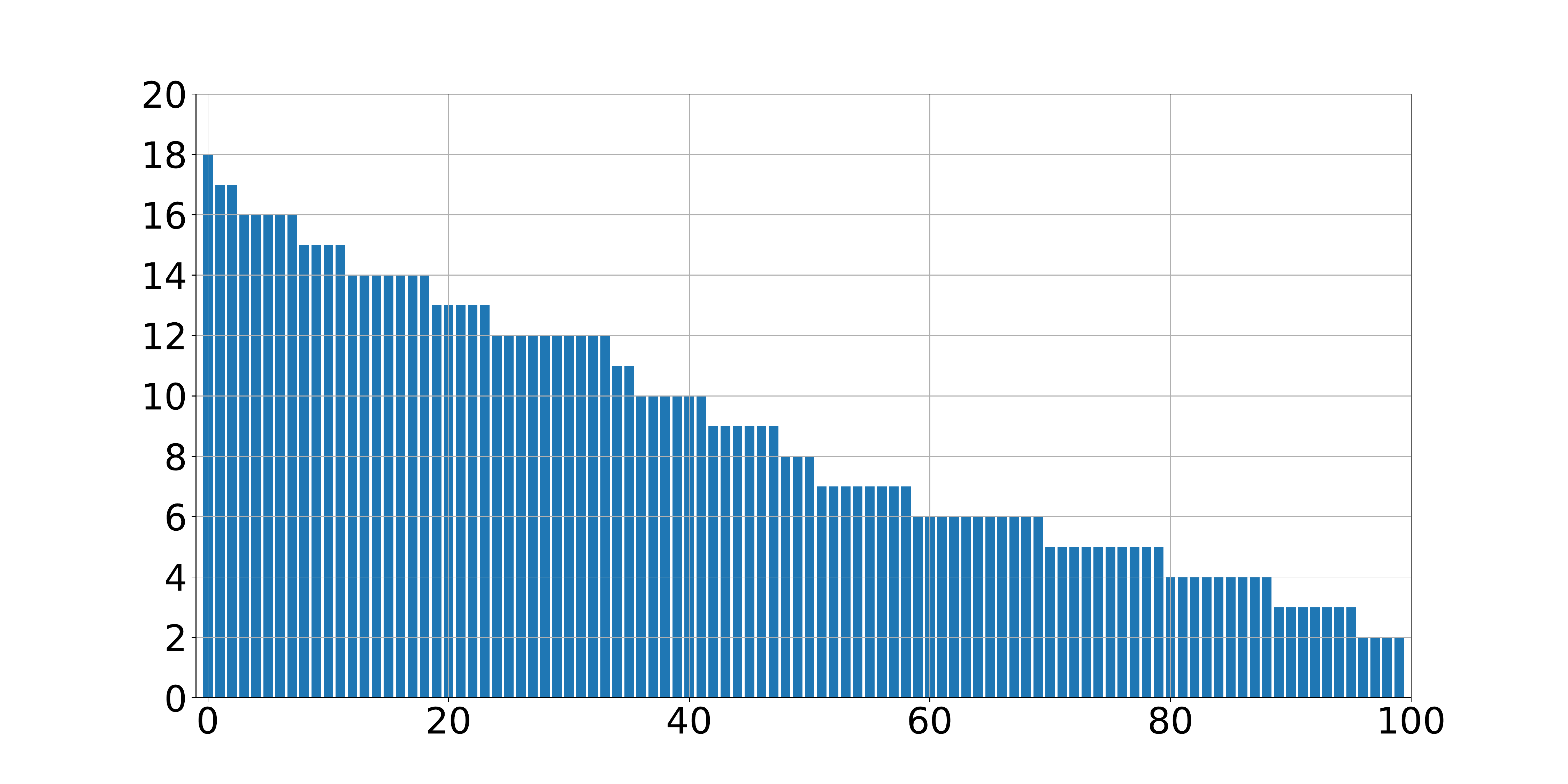} \hspace{0.03cm} &
   \rotatebox{90}{\footnotesize \quad \quad \quad \ \ \ \# clients}
   \includegraphics[trim=4cm 1cm 3.4cm 2.4cm, clip, width=\imgsizee]{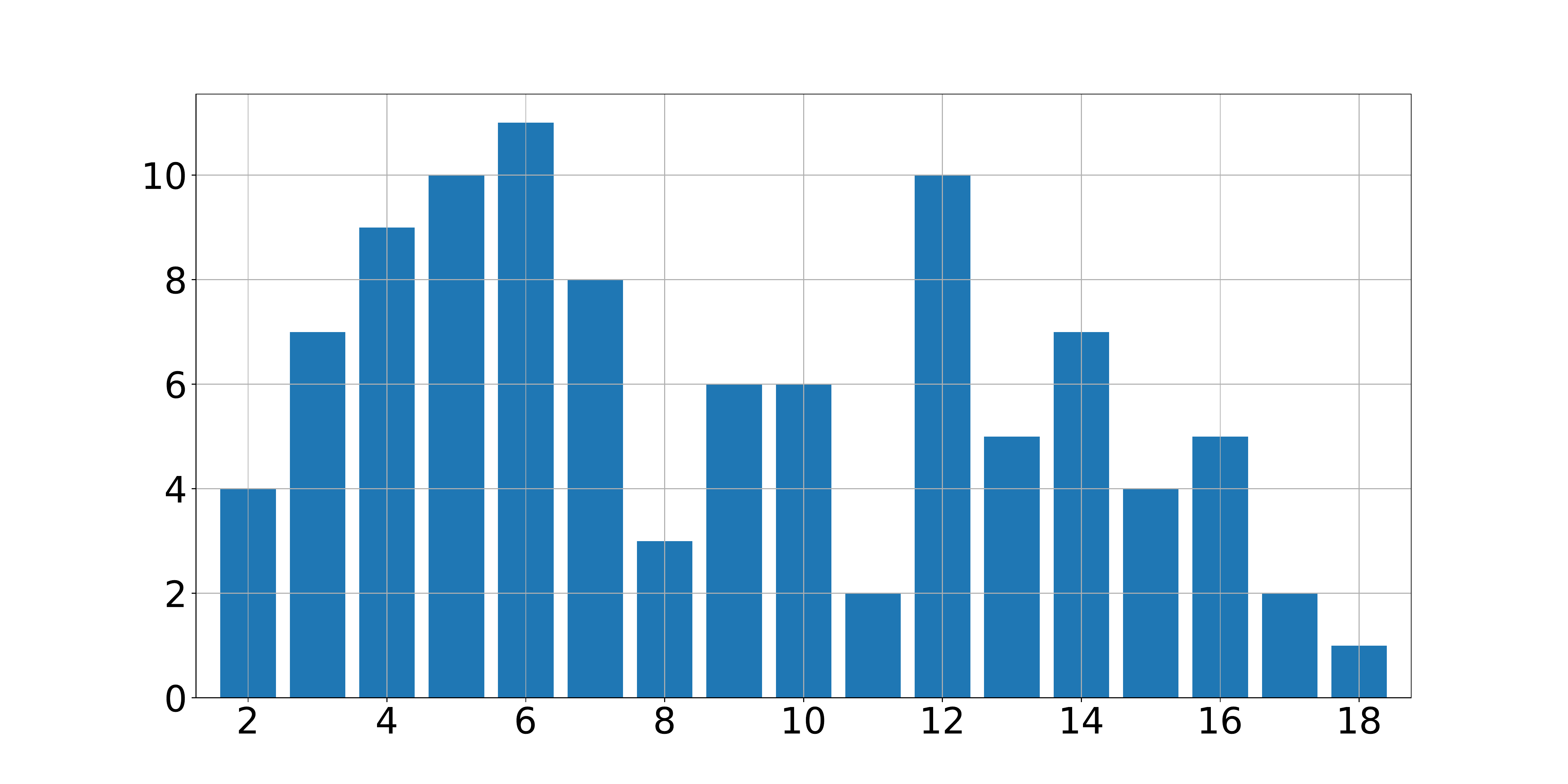} \hspace{0.03cm} &
   \rotatebox{90}{\footnotesize \quad \quad \quad \ \ client ID}
   \includegraphics[trim=4cm 1cm 3.4cm 2.4cm, clip, width=\imgsizee]{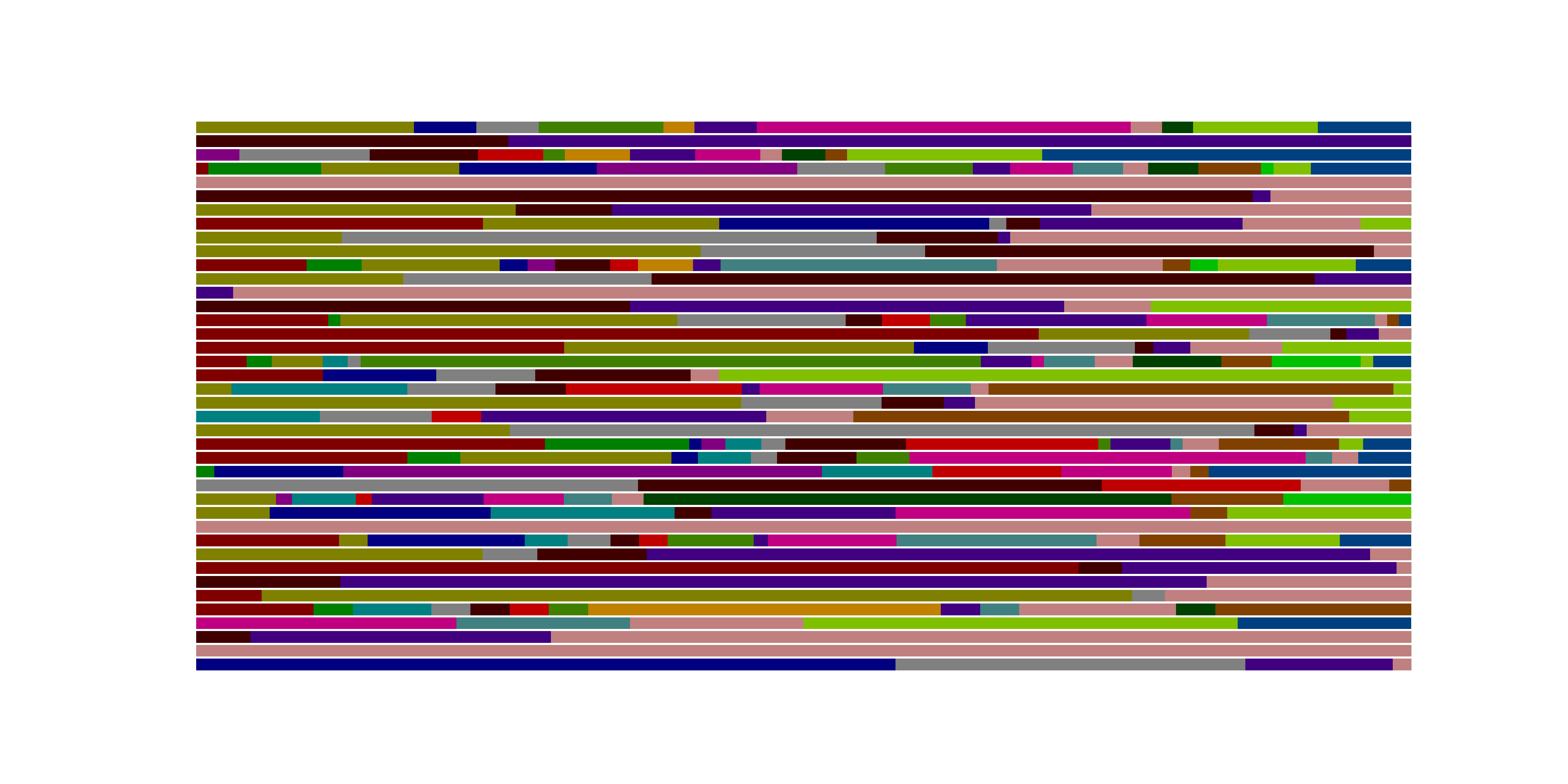} \\

   \rotatebox{90}{\quad \quad \ $\boldsymbol{\alpha=1}$} \hspace{0.05cm} &
   \rotatebox{90}{\footnotesize \quad \quad \quad \ \ \ \# classes}
   \includegraphics[trim=4cm 1cm 3.4cm 2.4cm, clip, width=\imgsizee]{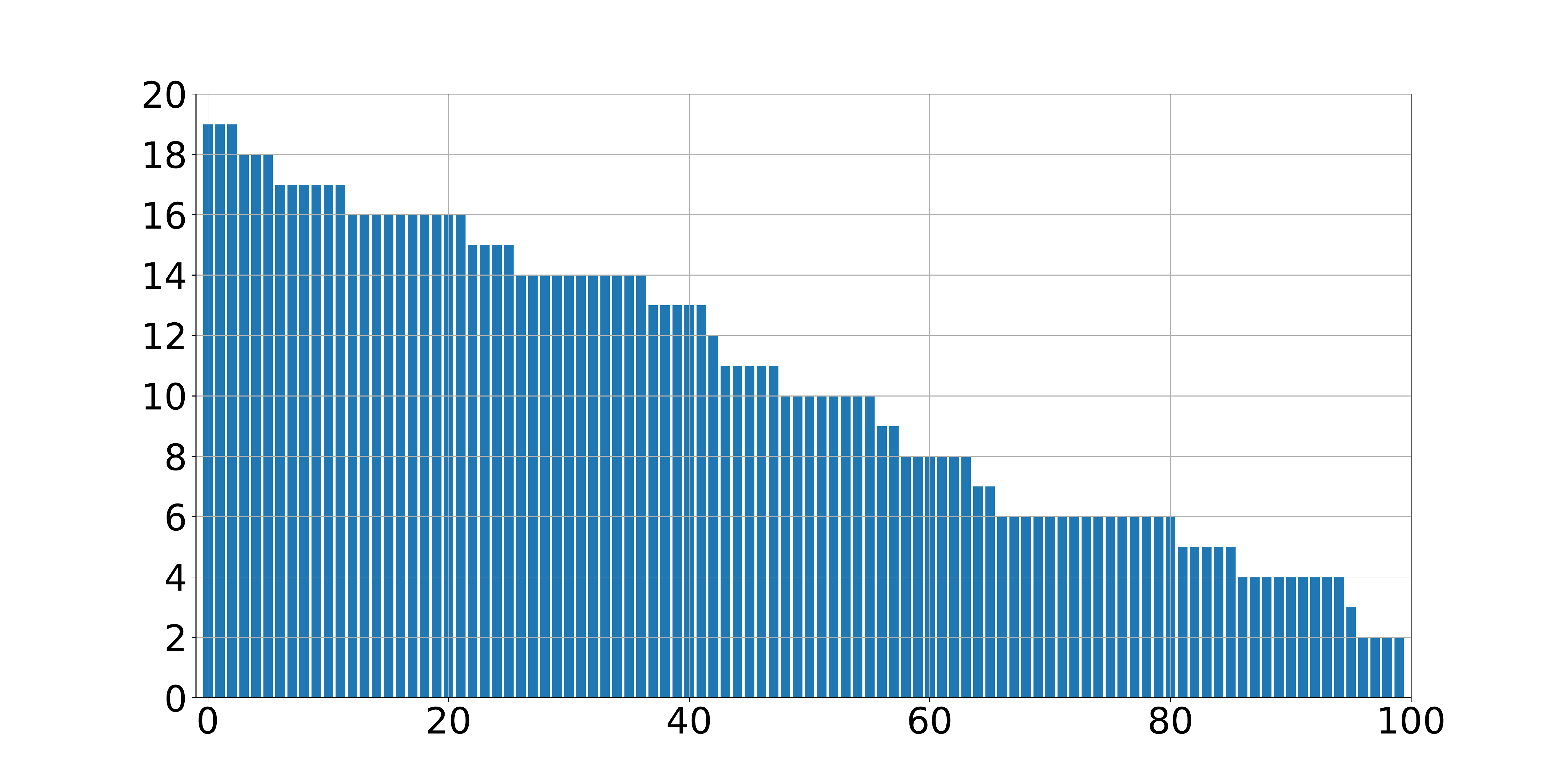} \hspace{0.03cm} &
   \rotatebox{90}{\footnotesize \quad \quad \quad \ \ \ \# clients}
   \includegraphics[trim=4cm 1cm 3.4cm 2.4cm, clip, width=\imgsizee]{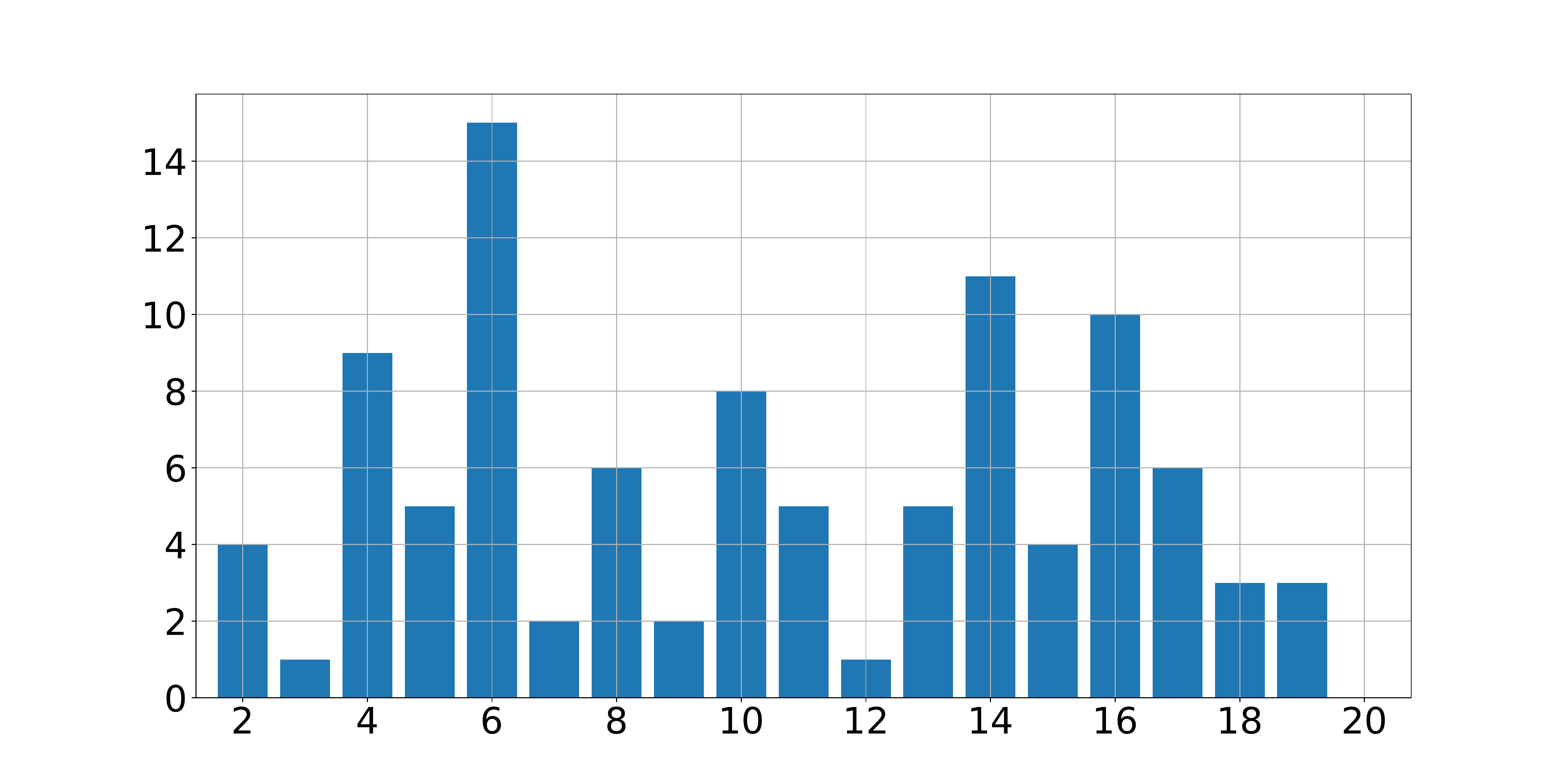} \hspace{0.03cm} &
   \rotatebox{90}{\footnotesize \quad \quad \quad \ \ client ID}
   \includegraphics[trim=4cm 1cm 3.4cm 2.4cm, clip, width=\imgsizee]{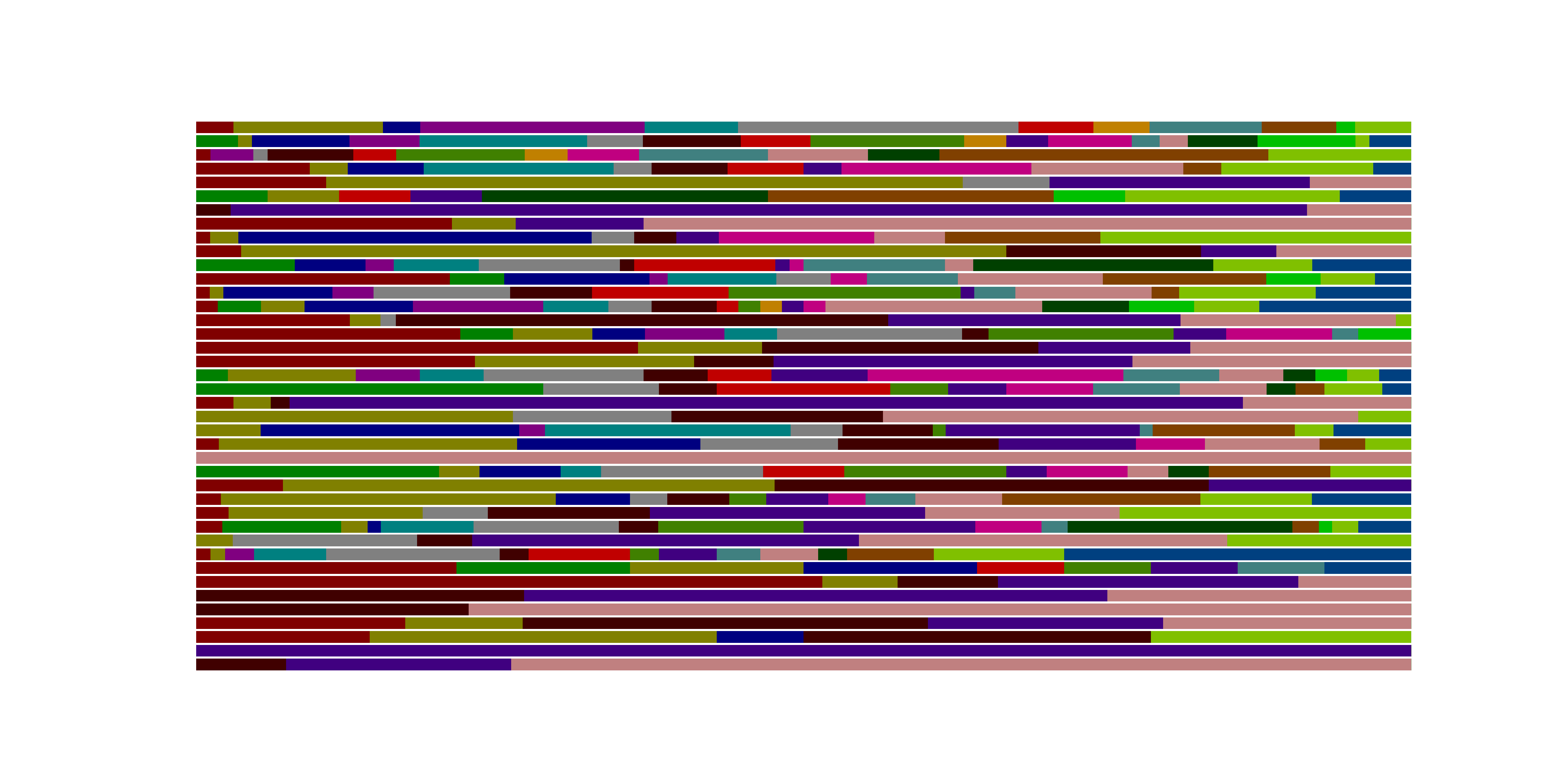} \\
  
   \rotatebox{90}{\quad \quad \ $\boldsymbol{\alpha=1000}$} \hspace{0.05cm} &
   \rotatebox{90}{\footnotesize \quad \quad \quad \ \ \ \# classes}
   \includegraphics[trim=4cm 1cm 3.4cm 2.4cm, clip, width=\imgsizee]{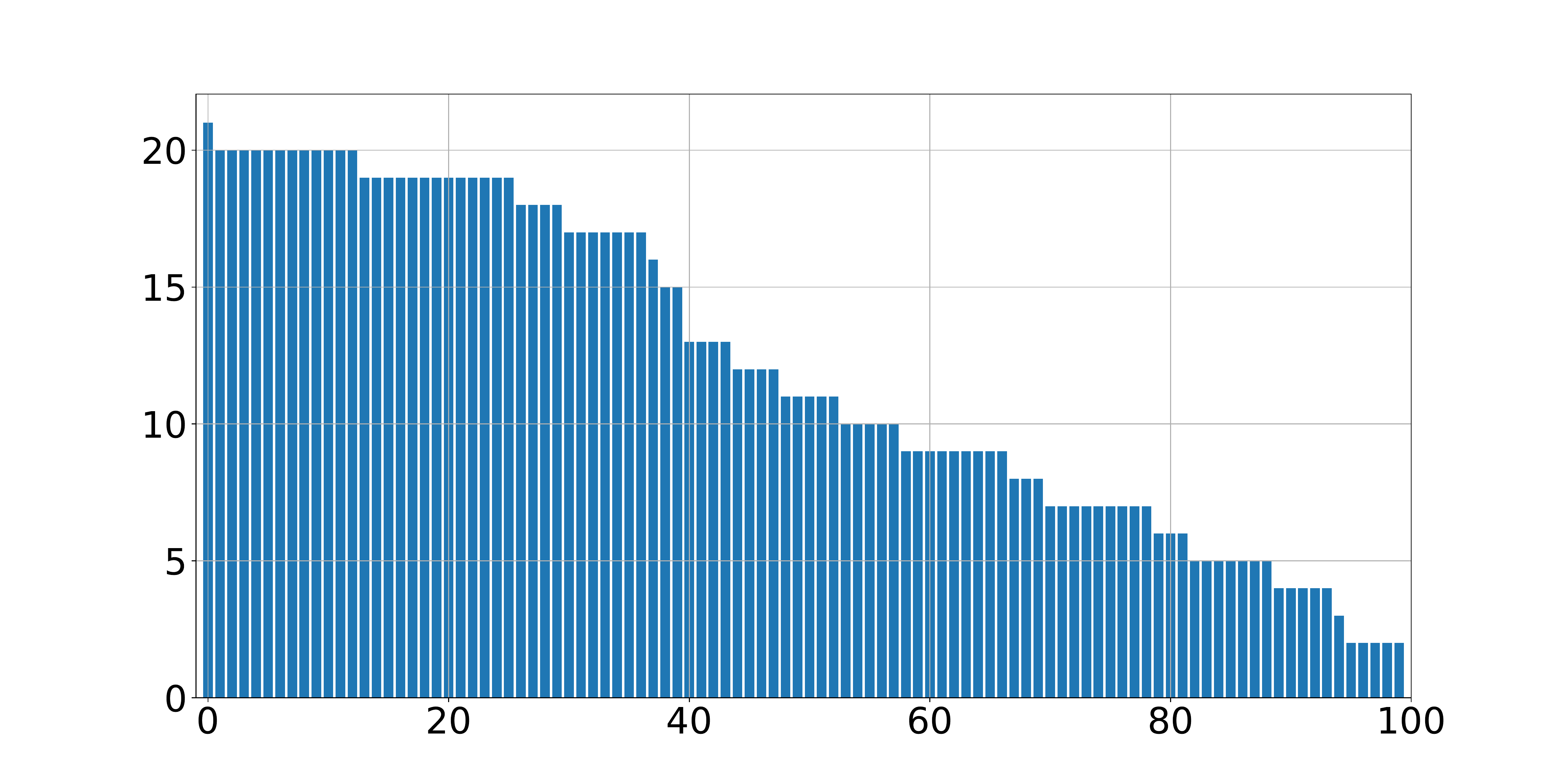} \hspace{0.03cm} &
   \rotatebox{90}{\footnotesize \quad \quad \quad \ \ \ \# clients}
   \includegraphics[trim=4cm 1cm 3.4cm 2.4cm, clip, width=\imgsizee]{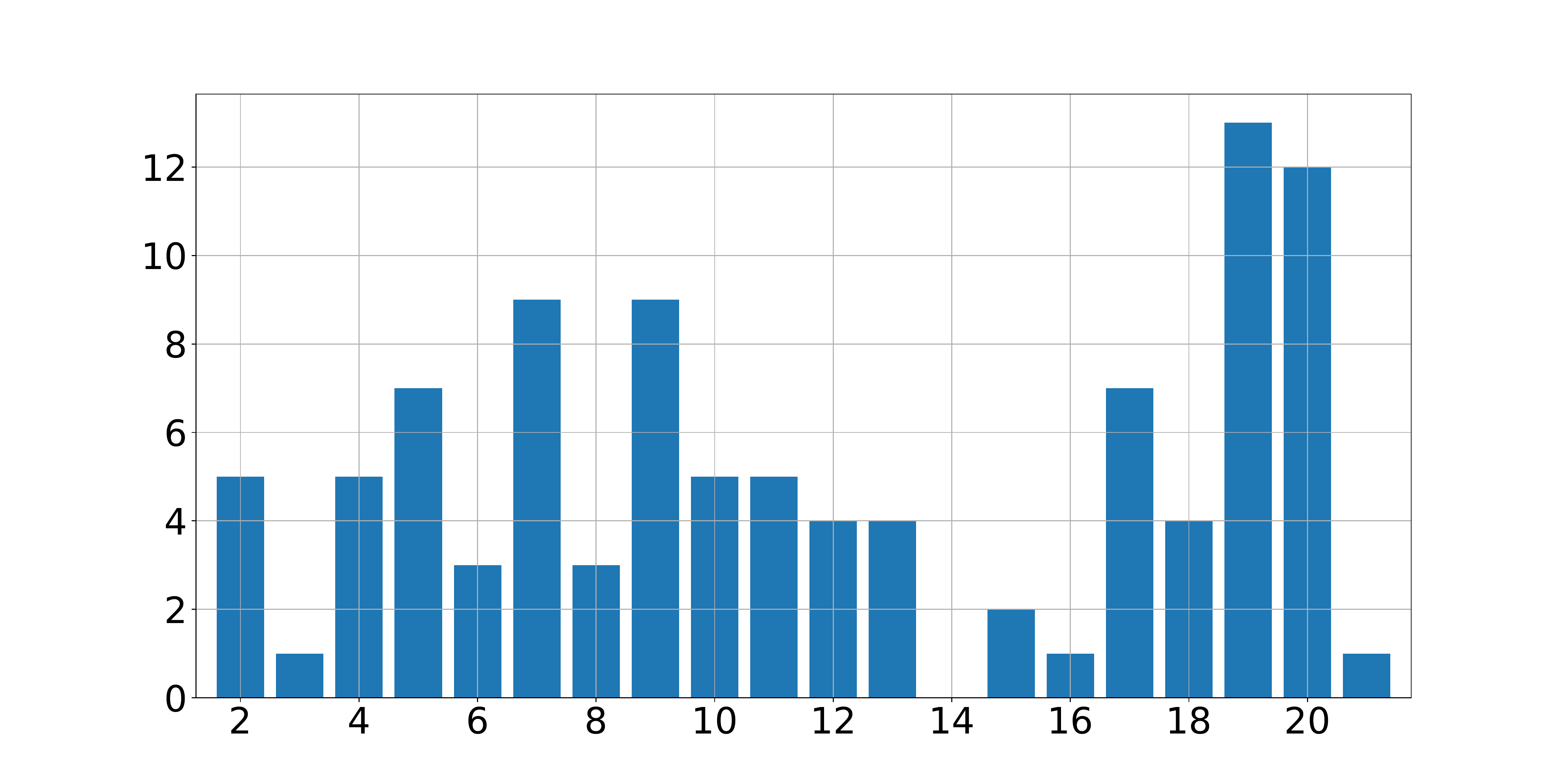} \hspace{0.03cm} &
   \rotatebox{90}{\footnotesize \quad \quad \quad \ \ client ID}
   \includegraphics[trim=4cm 1cm 3.4cm 2.4cm, clip, width=\imgsizee]{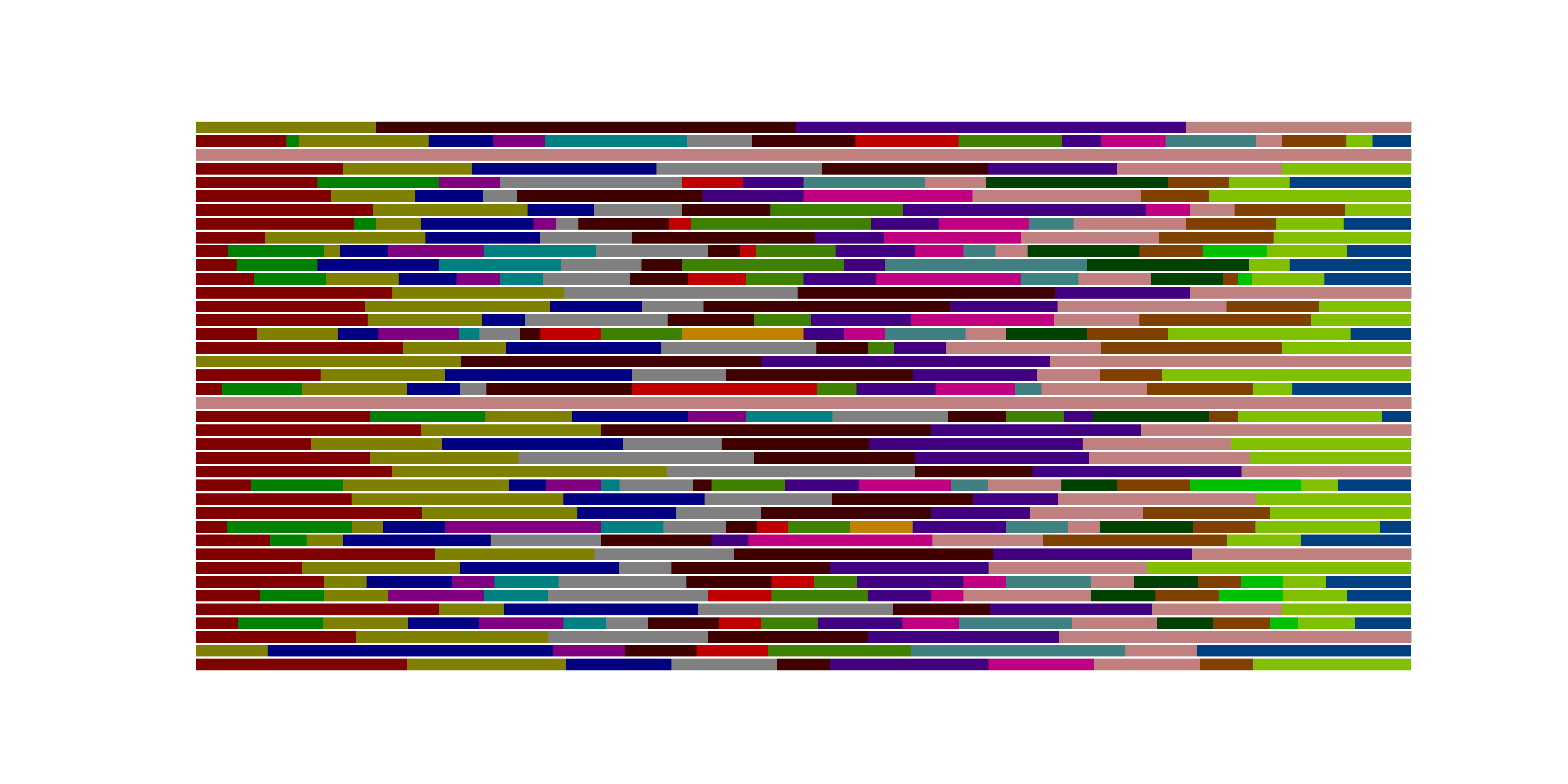} \\
   
  & \footnotesize client ID ($k$) & \footnotesize \# classes per client & \footnotesize class distribution
   
\end{tabular}
\vspace{0.05cm}
\caption{Dataset statistics of different data splitting schemes used by clients for the Pascal VOC2012 segmentation task. The first column reports the distribution of the number of classes among clients (note that the background is present in all the images). The second column shows the distribution of number of clients according to number of classes per client. The third column reports the per-client distribution of classes depicted with different colors, where  the client IDs are restricted to $30$ randomly sampled clients for visualization purposes (the background is not included in the visualization and the colors refer to the Pascal VOC2012 colormap). \textit{Best viewed in colors.}}
\label{suppl:fig:pascal_dataset_dirichlet}
\end{figure*}

\clearpage

\begin{figure}[htbp]
\centering
\setlength{\tabcolsep}{0.5pt} 
\renewcommand{\arraystretch}{0.4}
\centering
\begin{tabular}{cccc}
  
   & \textbf{Synthetic} & & \textbf{MNIST} \\
  
   \rotatebox{90}{\quad \quad \quad \quad Occurrences } &
   \includegraphics[trim=3.3cm 1.3cm 4.2cm 2.3cm, clip, width=0.46\linewidth]{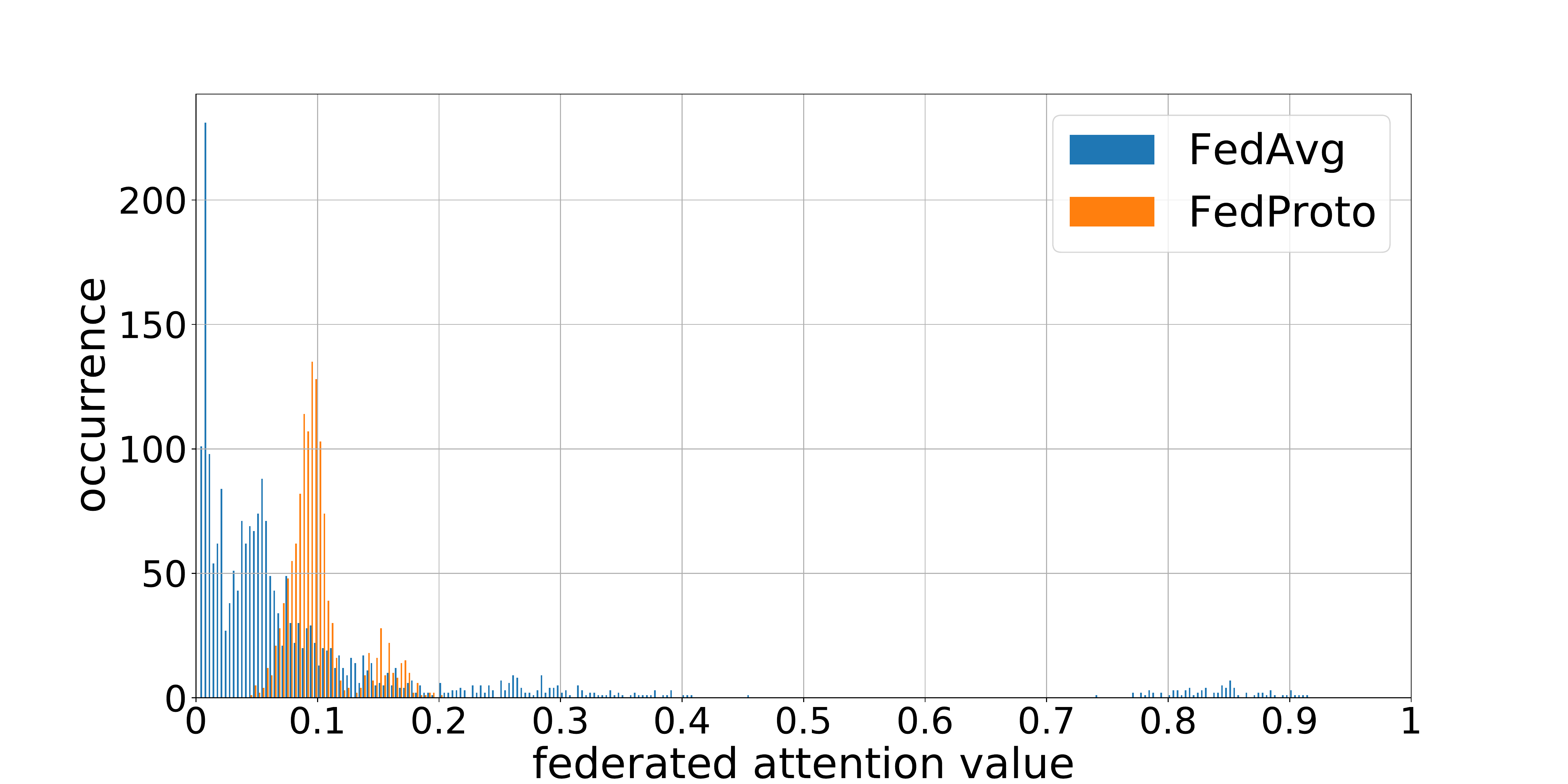} & \hspace{0.35cm} &
   \includegraphics[trim=3.3cm 1.3cm 4.2cm 2.3cm, clip, width=0.46\linewidth]{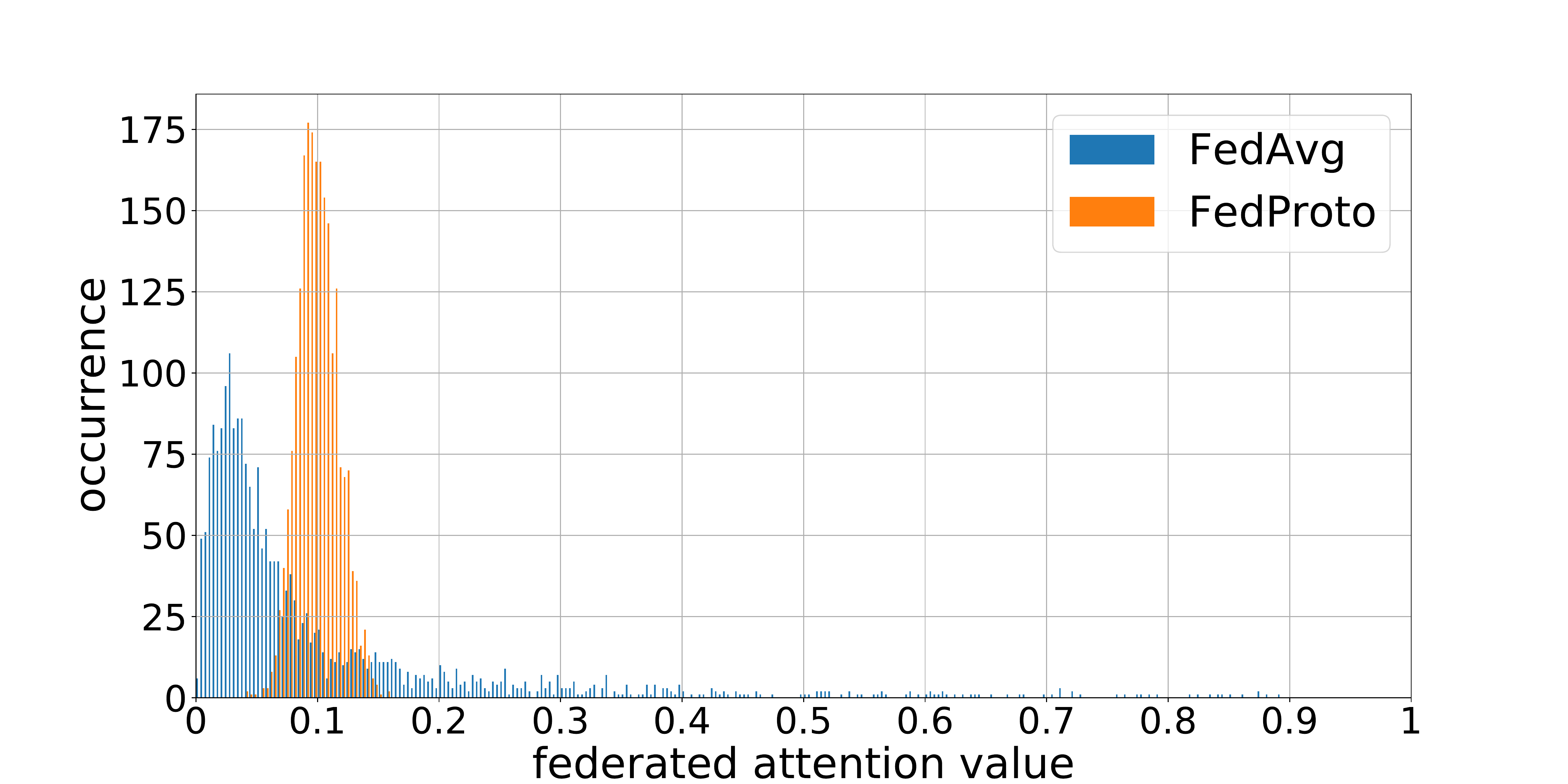} \\
   
   & \hspace{0.25cm} Federated Attention Value & & \hspace{0.25cm} Federated Attention Value \\
   & & & \\
   
      & \textbf{FEMNIST} & & \textbf{CelebA} \\
  
   \rotatebox{90}{\quad \quad \quad \quad Occurrences } &
   \includegraphics[trim=3.3cm 1.3cm 4.2cm 2.3cm, clip, width=0.46\linewidth]{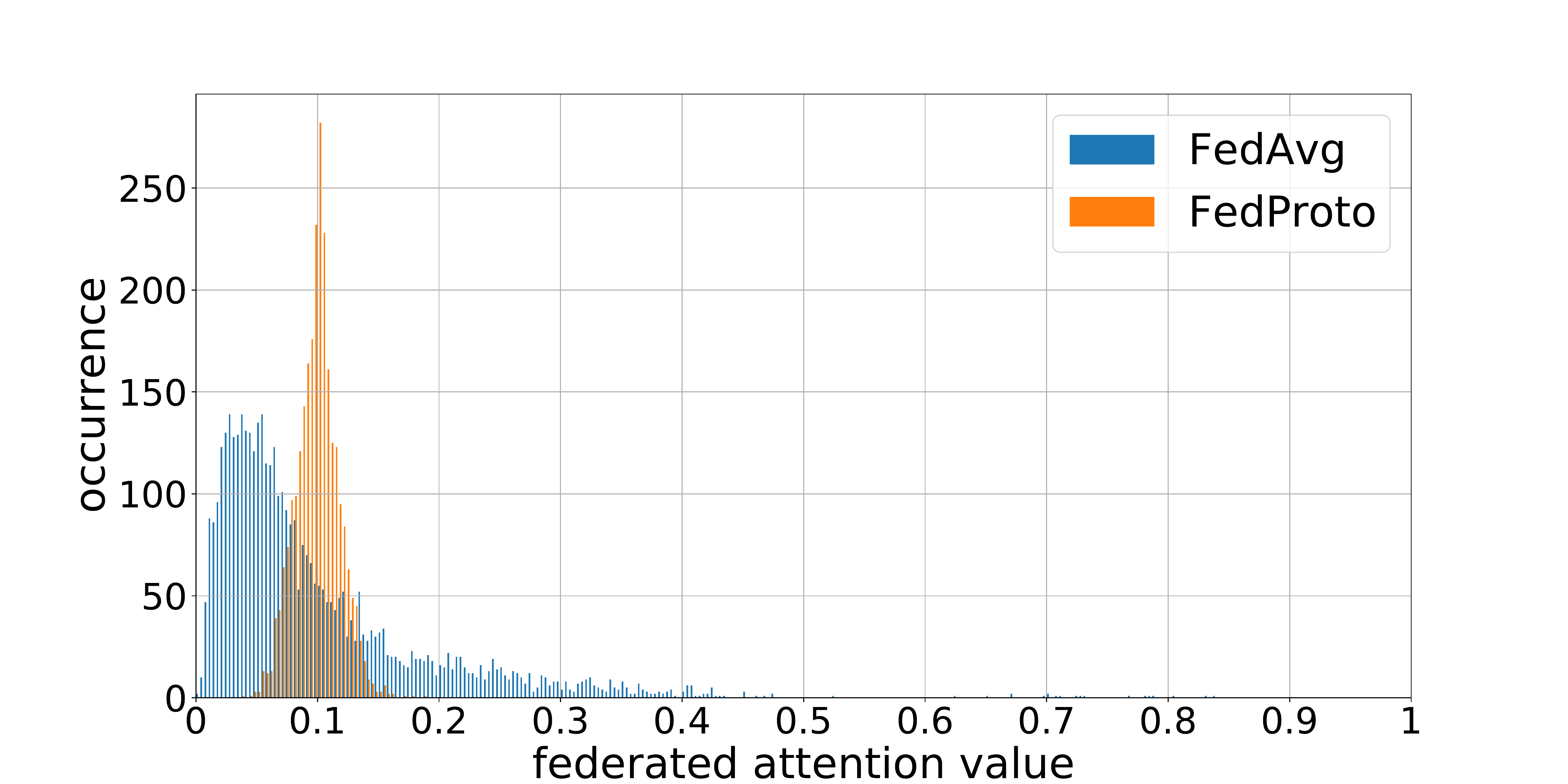} & \hspace{0.35cm} &
   \includegraphics[trim=3.3cm 1.3cm 4.2cm 2.3cm, clip, width=0.46\linewidth]{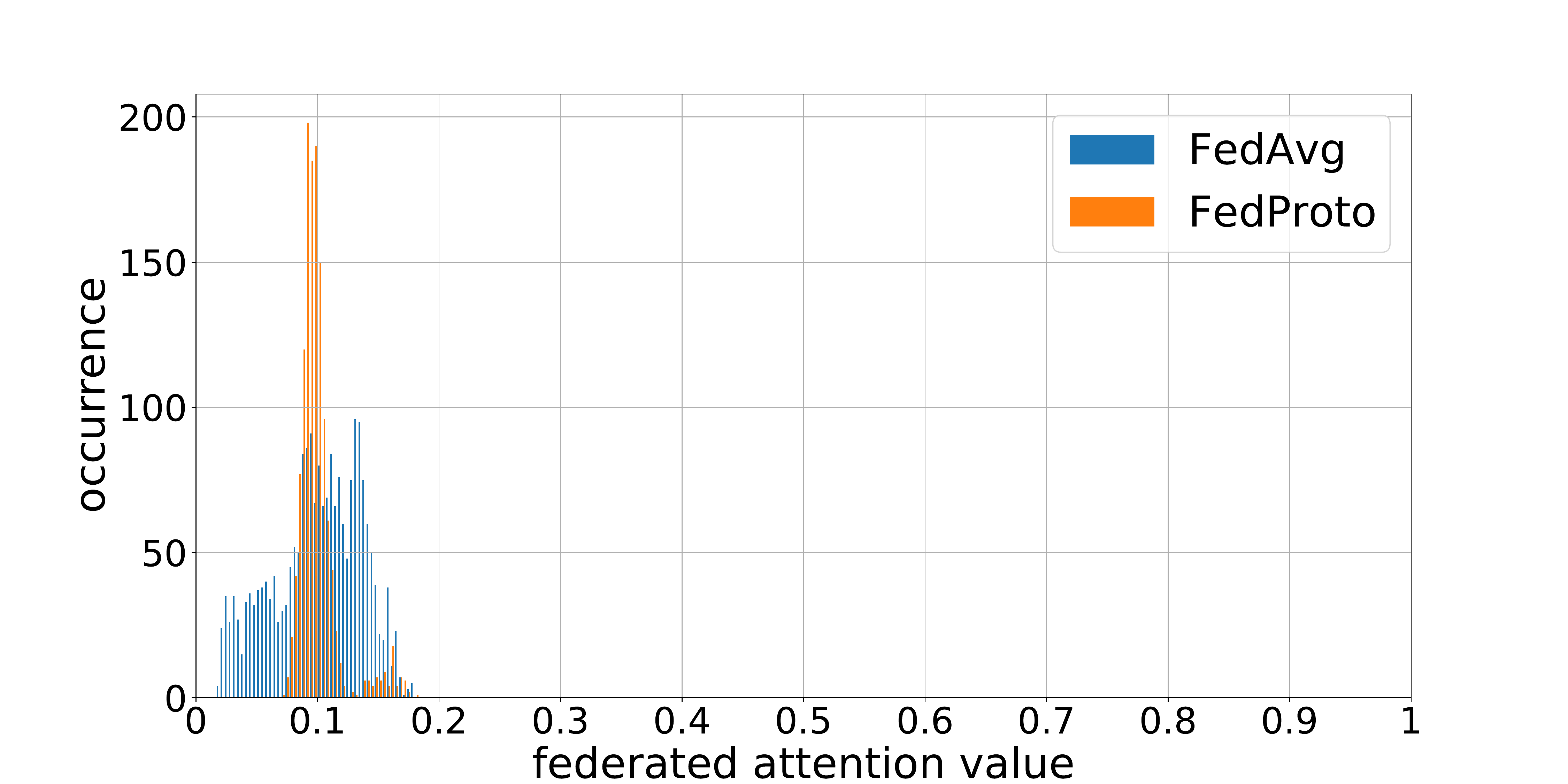} \\
   
   & \hspace{0.25cm} Federated Attention Value & & \hspace{0.25cm} Federated Attention Value \\
   & & & \\
  
\end{tabular}
\caption{Comparison of distributions of the federated attention vector $\mathbf{a}^t[k], \forall k, \forall t$, on classification datasets for FedAvg and our FedProto.}
\label{suppl:fig:att_vector_classification}
\end{figure}

\section{Additional Experimental Analyses}
\label{suppl:sec:results}

\subsection{Federated Attention Values in Image Classification}

In the main paper, we mentioned that our approach tends to produce federated attention values $\mathbf{a}^t[k]$ deviating less from a fairness policy (\ie, aggregation of weights $W_k^t$ by ${\mathbf{f}^t[k] = 1 / K'}, \forall k, \forall t$) than FedAvg, as also observed in other contemporary approaches \cite{wu2020fast,zhang2020hierarchically}. In Figure~\ref{suppl:fig:att_vector_classification}, we compare the distribution of federated attention values $\mathbf{a}^t[k], \forall k, \forall t,$ of FedAvg and of our approach. The results show that FedProto produces attention values having a much lower variance from the average value (we remark that $|\mathcal{K}| = 10$ is used) compared to FedAvg. In particular, FedAvg weights follow the same distribution of the number of samples, which could lead the framework to ignore clients with less samples during aggregation, regardless of the statistical distribution of local samples.
Moving from these considerations, in the main paper we reported some results employing a fairness policy, which we found to have comparable results to FedAvg and to be significantly surpassed by our approach.

\subsection{Additional Qualitative Results on Semantic Segmentation}

One of the main effects of our proposed FedProto is a class-conditional latent-level regularization, achieved via prototype guided federated optimization and margin maximization of the aggregate model during its distributed training. In order to visually represent the main effects, we report in Figure \ref{suppl:fig:pascal_tsne} the 2D t-SNE embeddings of the features of the final aggregate model \cite{van2008visualizing} for different values of $\alpha$. Here, the background class is not included and the colors refer to the Pascal VOC2012 colormap. Class membership for the low-resolution feature map is obtained with simple nearest neighbor downsampling of the full-resolution segmentation maps.
By visual inspection, we observe that t-SNE embeddings produced using the final aggregate model from FedProto are better subdivided into clusters (\ie, points of the same color). In particular, for $\alpha=0.01$, FedAvg confuses some animal classes (horse in pink, sheep in brown, cow in green, cat in dark red and dog in purple) into one mixed point cloud on the top right part of the plot, lacking class-discrimination at the feature level, as argued in the main paper.
FedProto, instead, produces a much clearer separation among these (and others) classes, being able to build class-discriminative clusters at the latent level (\ie, clusters points on the basis of their class membership). Similar discussion can be made also for the remaining scenarios, with a progressively smaller difference between t-SNE embeddings produced by FedAvg and FedProto as the data i.i.d.-ness increases.

\begin{figure*}[htbp]
\centering
\setlength{\tabcolsep}{0.5pt} 
\renewcommand{\arraystretch}{0.4}
\centering
\footnotesize
\begin{tabular}{cccccccccc}
    
   & &  \multicolumn{6}{c}{\includegraphics[trim=0cm 17.5cm 3.8cm 0cm, clip, width=0.5\linewidth]{img/arrows/arrow_v2.pdf}} & & \\
  
   & & \multicolumn{2}{c}{$\alpha=0.01$} & \multicolumn{2}{c}{$\alpha=0.1$} & \multicolumn{2}{c}{$\alpha=1$} & & \\
  
   \rotatebox{90}{\ \ Segmentation} &
   \includegraphics[width=\imgsize]{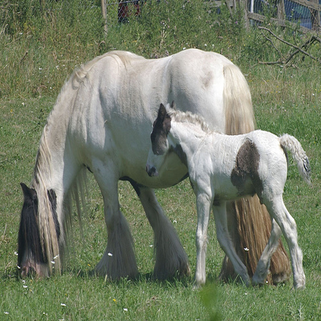} &
   \includegraphics[width=\imgsize]{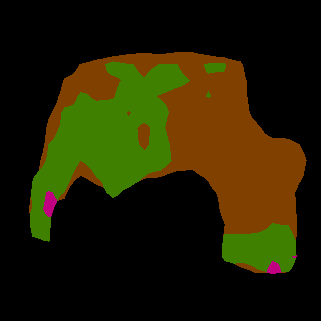} &
   \includegraphics[width=\imgsize]{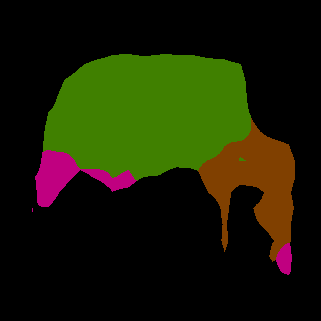} &   
   \includegraphics[width=\imgsize]{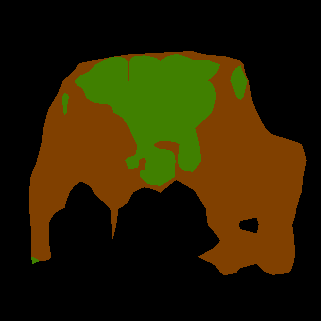} &   
   \includegraphics[width=\imgsize]{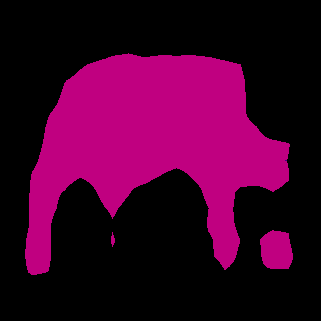} &   
   \includegraphics[width=\imgsize]{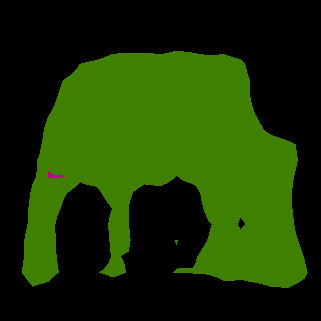} &   \includegraphics[width=\imgsize]{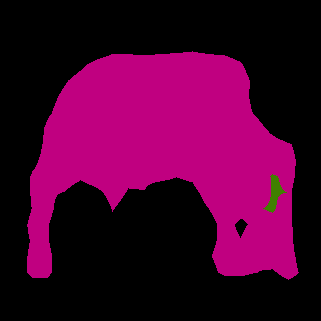} &
   \includegraphics[width=\imgsize]{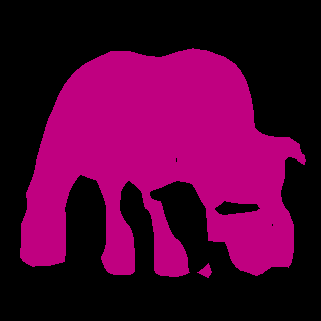} &
   \includegraphics[width=\imgsize]{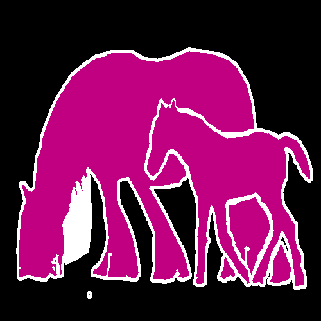} \\

   \rotatebox{90}{\ \ \ Soft.\ Entropy} &
   \includegraphics[width=\imgsize]{img/suppl/pascal_qualitative/img794_RGB.png} &
   \includegraphics[width=\imgsize]{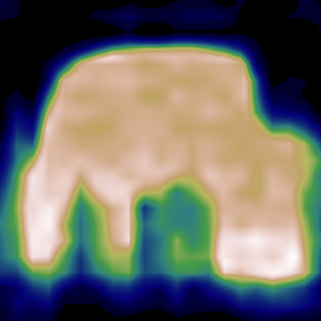} &
   \includegraphics[width=\imgsize]{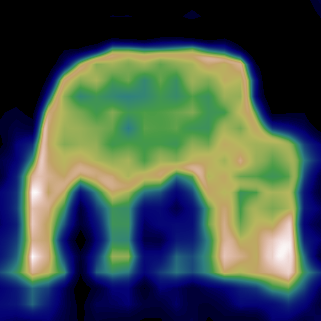} &   
   \includegraphics[width=\imgsize]{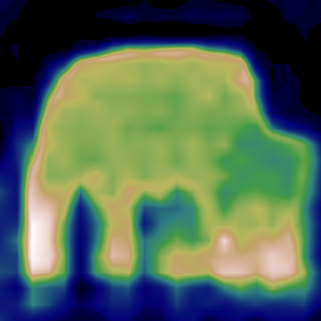} &   
   \includegraphics[width=\imgsize]{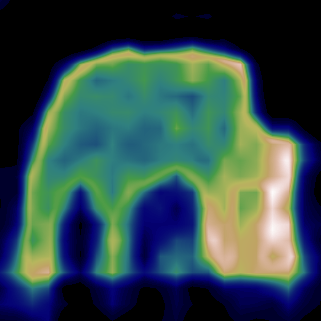} &   
   \includegraphics[width=\imgsize]{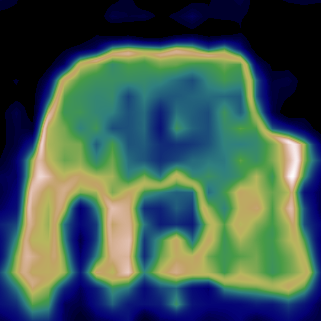} &   \includegraphics[width=\imgsize]{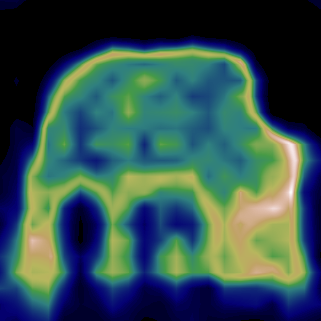} &
   \includegraphics[width=\imgsize]{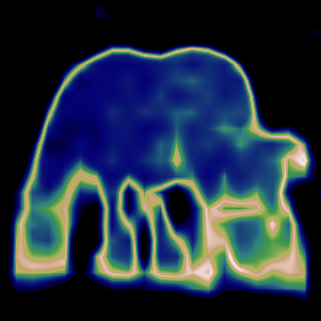} & 
   \multirow{2}{*}[17mm]{\includegraphics[width=4.7mm]{img/pascal_qualitative/colorbar.pdf}} \\
   
   \rotatebox{90}{\ \ Feat.\ Entropy} &
   \includegraphics[width=\imgsize]{img/suppl/pascal_qualitative/img794_RGB.png} &
   \includegraphics[width=\imgsize]{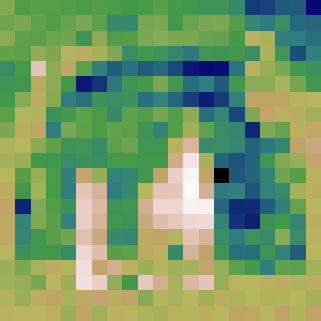} &
   \includegraphics[width=\imgsize]{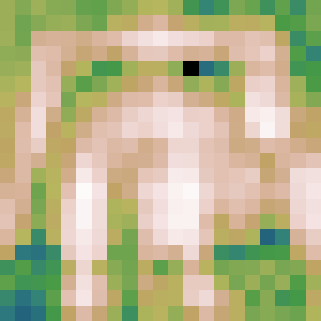} &   
   \includegraphics[width=\imgsize]{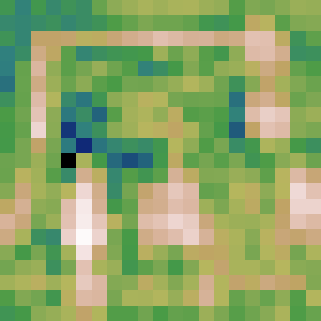} &   
   \includegraphics[width=\imgsize]{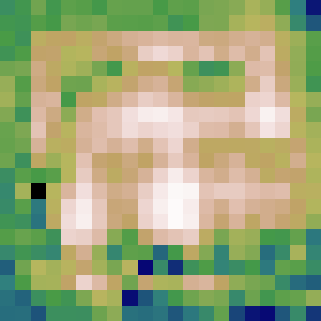} &   
   \includegraphics[width=\imgsize]{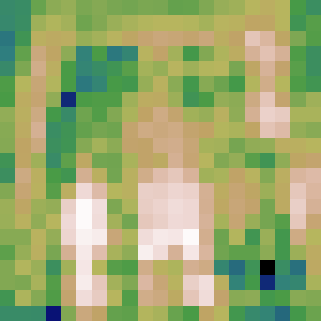} &   \includegraphics[width=\imgsize]{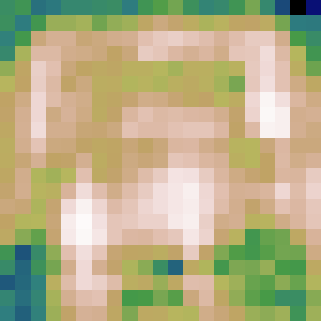} &
   \includegraphics[width=\imgsize]{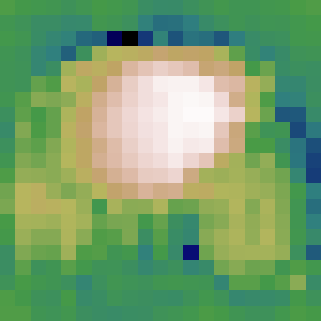} & \\
   
   
   \rotatebox{90}{\ \ Segmentation} &
   \includegraphics[width=\imgsize]{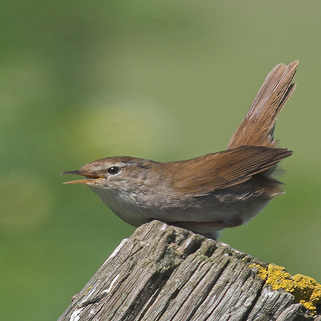} &
   \includegraphics[width=\imgsize]{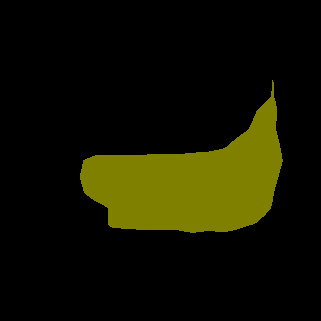} &
   \includegraphics[width=\imgsize]{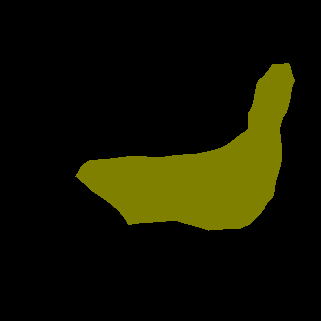} &   
   \includegraphics[width=\imgsize]{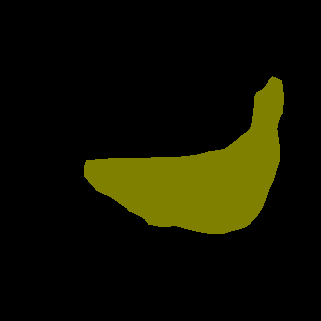} &   
   \includegraphics[width=\imgsize]{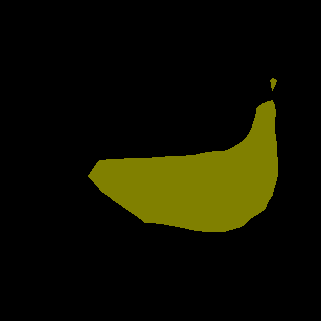} &   
   \includegraphics[width=\imgsize]{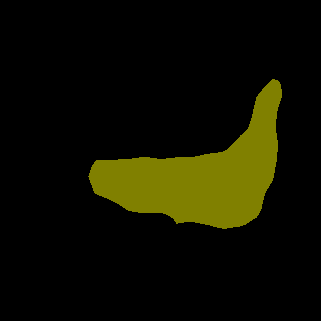} &   \includegraphics[width=\imgsize]{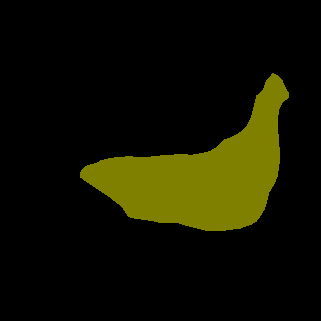} &
   \includegraphics[width=\imgsize]{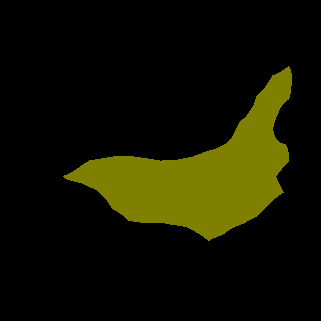} &
   \includegraphics[width=\imgsize]{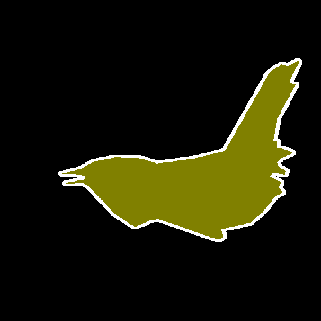} \\

   \rotatebox{90}{\ \ \ Soft.\ Entropy} &
   \includegraphics[width=\imgsize]{img/suppl/pascal_qualitative/img817_RGB.png} &
   \includegraphics[width=\imgsize]{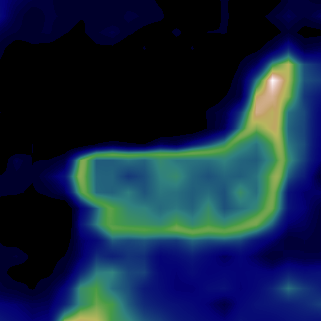} &
   \includegraphics[width=\imgsize]{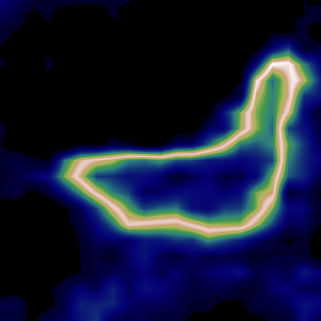} &   
   \includegraphics[width=\imgsize]{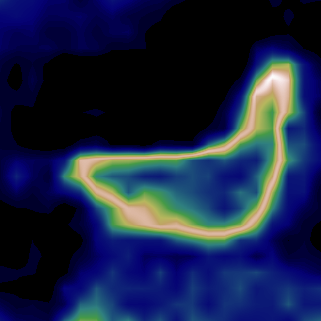} &   
   \includegraphics[width=\imgsize]{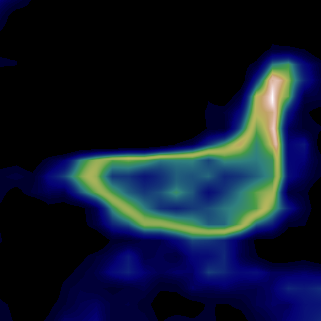} &   
   \includegraphics[width=\imgsize]{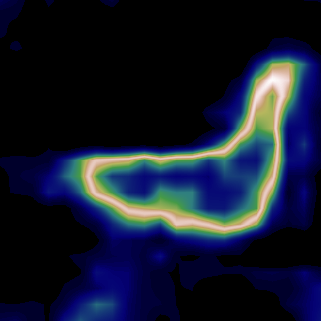} &   \includegraphics[width=\imgsize]{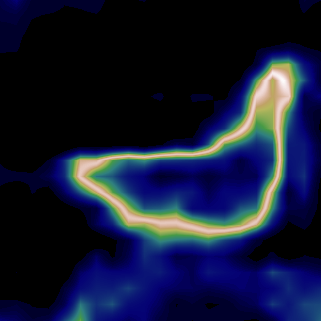} &
   \includegraphics[width=\imgsize]{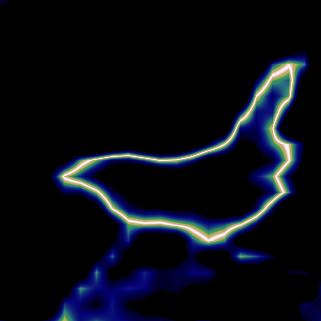} &
   \multirow{2}{*}[17mm]{\includegraphics[width=4.7mm]{img/pascal_qualitative/colorbar.pdf}} \\
   
   \rotatebox{90}{\ \ Feat.\ Entropy} &\includegraphics[width=\imgsize]{img/suppl/pascal_qualitative/img817_RGB.png} &
   \includegraphics[width=\imgsize]{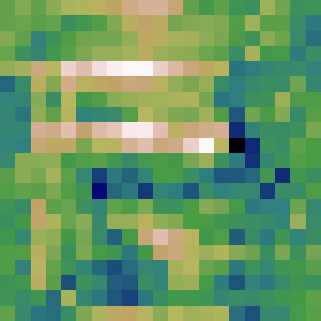} &
   \includegraphics[width=\imgsize]{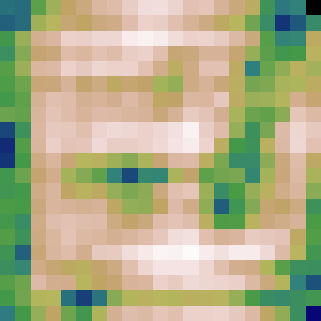} &   
   \includegraphics[width=\imgsize]{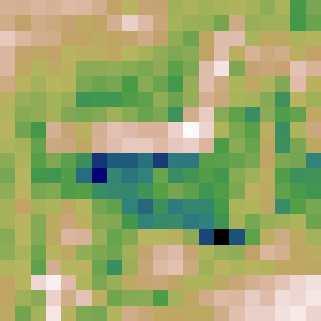} &   
   \includegraphics[width=\imgsize]{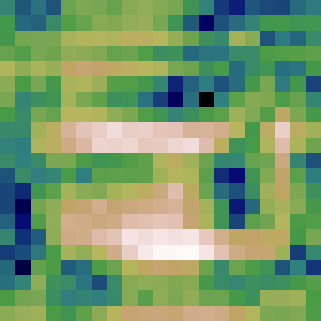} &   
   \includegraphics[width=\imgsize]{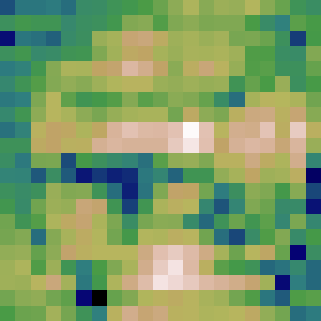} &   \includegraphics[width=\imgsize]{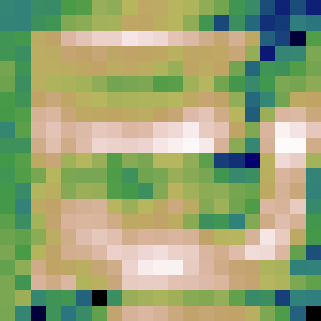} &
   \includegraphics[width=\imgsize]{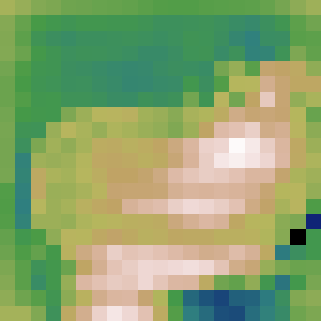} & \\
   
   
   \rotatebox{90}{\ \ Segmentation} &
   \includegraphics[width=\imgsize]{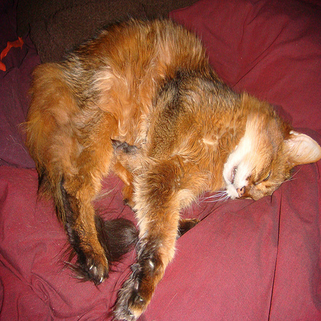} &
   \includegraphics[width=\imgsize]{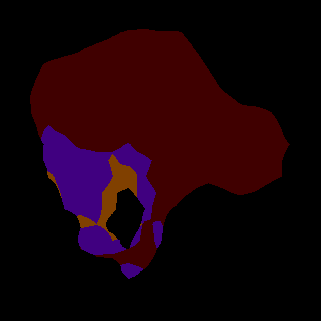} &
   \includegraphics[width=\imgsize]{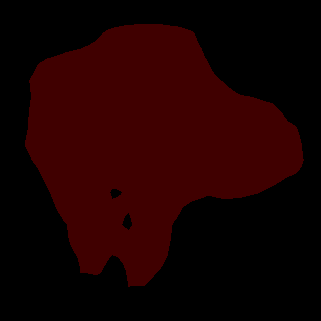} &   
   \includegraphics[width=\imgsize]{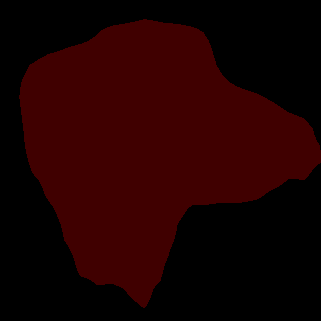} &   
   \includegraphics[width=\imgsize]{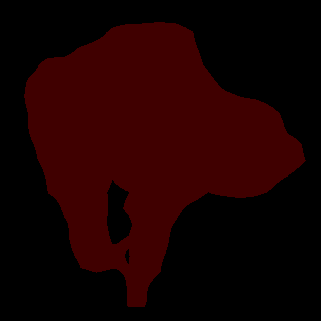} &   
   \includegraphics[width=\imgsize]{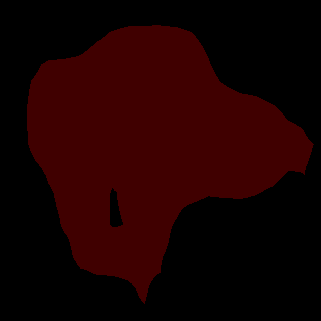} &   \includegraphics[width=\imgsize]{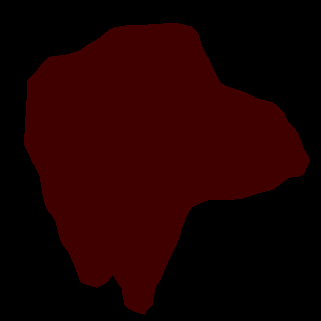} &
   \includegraphics[width=\imgsize]{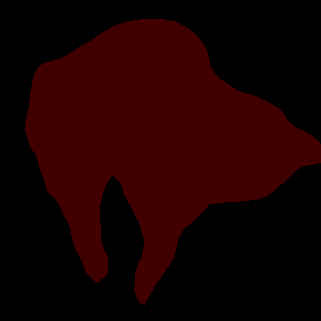} &
   \includegraphics[width=\imgsize]{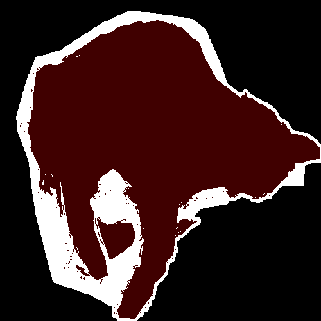} \\

   \rotatebox{90}{\ \ \ Soft.\ Entropy} &
   \includegraphics[width=\imgsize]{img/suppl/pascal_qualitative/img919_RGB.png} &
   \includegraphics[width=\imgsize]{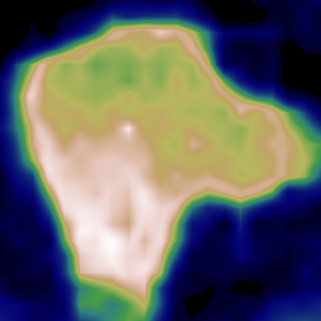} &
   \includegraphics[width=\imgsize]{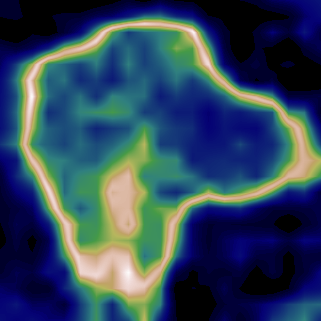} &   
   \includegraphics[width=\imgsize]{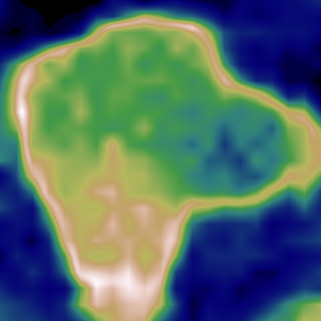} &   
   \includegraphics[width=\imgsize]{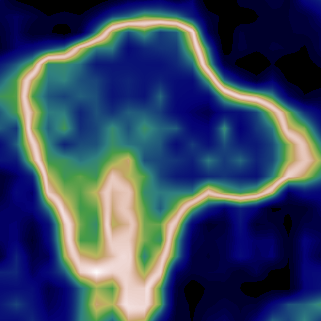} &   
   \includegraphics[width=\imgsize]{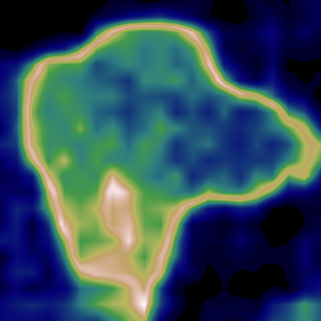} &   \includegraphics[width=\imgsize]{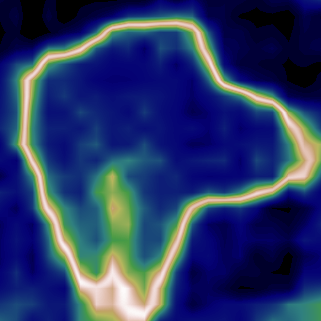} &
   \includegraphics[width=\imgsize]{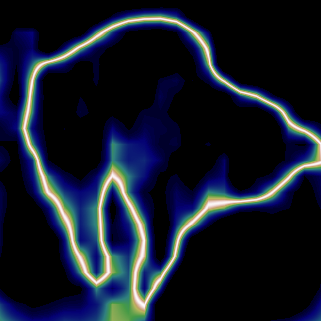} &
   \hspace{0cm} \multirow{2}{*}[17mm]{\includegraphics[trim=0cm 12.5cm 30.1cm 0cm, clip, width=18mm]{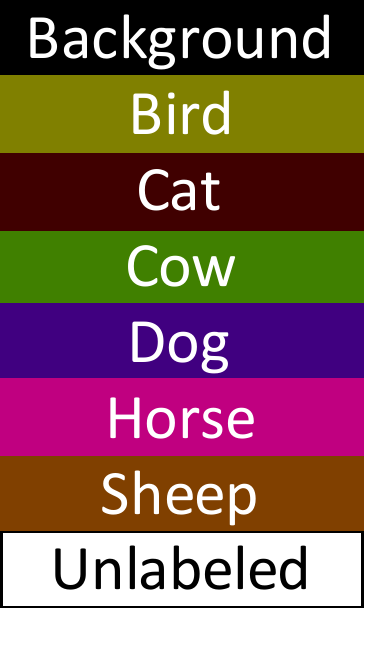}} \\
   
   \rotatebox{90}{\ \ Feat.\ Entropy} &\includegraphics[width=\imgsize]{img/suppl/pascal_qualitative/img919_RGB.png} &
   \includegraphics[width=\imgsize]{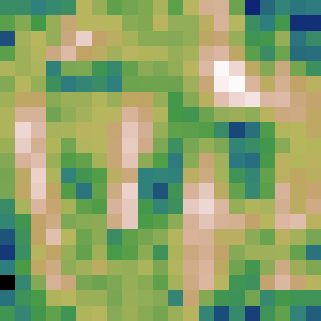} &
   \includegraphics[width=\imgsize]{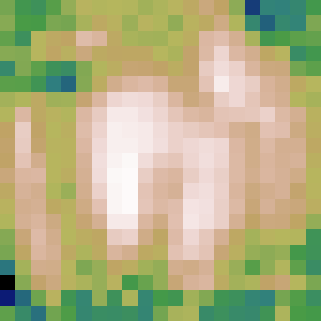} &   
   \includegraphics[width=\imgsize]{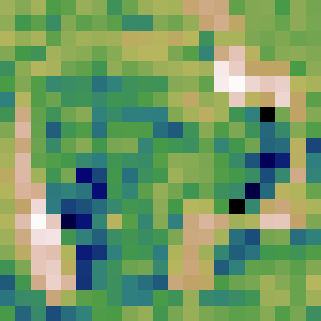} &   
   \includegraphics[width=\imgsize]{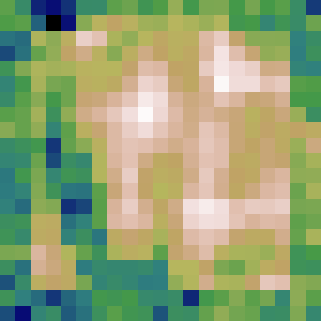} &   
   \includegraphics[width=\imgsize]{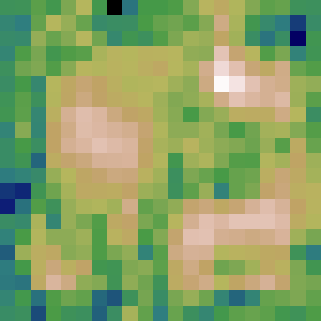} &   \includegraphics[width=\imgsize]{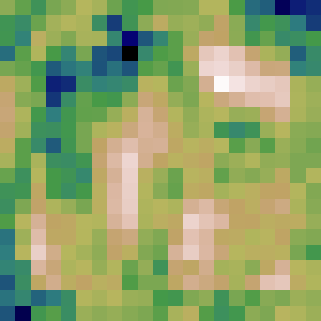} &
   \includegraphics[width=\imgsize]{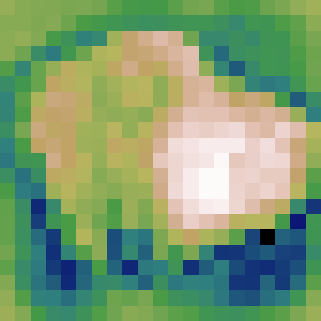} & \\
   
   & RGB &  FedAvg  &  FedProto  & FedAvg & FedProto & FedAvg  &  FedProto  &  Centralized & GT
  
\end{tabular}
\caption{Qualitative results for models trained using FedAvg and FedProto using three non-i.i.d.\ to i.i.d.\ configurations of Pascal VOC2012 dataset. For each of the three sample images, we depict; the output segmentation map (rows 1, 4 and 7), the softmax-level entropy map (rows 2, 5 and 8), and the feature-level entropy map (rows 3, 6 and 9). As a reference, output maps of models obtained using centralized training  are shown on the second last column. \textit{Best viewed in colors}.}
\label{suppl:fig:suppl_qual_res}
\end{figure*}

To conclude, we report in Figure~\ref{suppl:fig:suppl_qual_res} a qualitative analysis on segmentation and entropy maps of three sample images comparing the final aggregate models of FedAvg and FedProto on different data splitting configurations (\ie, for different values of ${\alpha \in \{0.01, 0.1, 1 \}}$). In particular, for each image, we show the predicted segmentation map (rows $1$, $4$ and $7$), the entropy map of the final softmax layer (rows $2$, $5$ and $8$) and the entropy map of the intermediate features (rows $3$, $6$ and $9$). 
Looking at the overall picture, we can appreciate a general improvement when going from more non-i.i.d.\ to more i.i.d.\ data, as the complexity of the optimization decreases.
Analogous considerations to those reported in the main paper also hold in this case. 

Our model generally achieves higher quality \textbf{segmentation maps}, which generally improve when data are more i.i.d., better resembling segmentation maps produced by centralized training. FedProto produces better segmentation maps for more non-i.i.d.\ data compared to FedAvg: more correct class identification in the first sample (horse, instead of cow or sheep) and of the third sample (in the non-i.i.d.\ case), and better objects shaping in the last two sample images. The ability to distinguish between class ambiguity is the direct consequence of a better latent space organization and regularization that FedProto achieves by maximizing prototype margin.

Furthermore, FedProto provides less uncertainty on the chosen classification labels than FedAvg, as shown by the \textbf{softmax-level entropy maps}. Here, pixel-wise uncertainty on the prediction of the network is measured via entropy levels: the lower the entropy (\ie, the darker the pixels) and the higher is the confidence of the prediction, being representative of a peaked distribution over the class probabilities.
 
Finally, we report the \textbf{feature-level entropy maps}, which measures how representative a feature is at each pixel location. Features corresponding to the desired class should be well activated so that the decoder can discriminate between them and assign the correct label: this is the case of centralized training, where features corresponding to (certain parts of) the object class are bright (\ie, high entropy denoting many activated patterns). Also in this case, we observe that FedProto produces feature-level entropy maps more similar to centralized training than the maps produced by FedAvg (particularly visible for low values of $\alpha$).


\newcommand{\imgsizeee}{23.9mm}
\begin{figure*}[t]
\centering
\setlength{\tabcolsep}{1pt} 
\renewcommand{\arraystretch}{2}
\begin{tabular}{cccccccc}
  
  &  \multicolumn{7}{c}{\hspace{0.75cm}\includegraphics[trim=0cm 18.15cm 4cm 0cm, clip, width=0.9\linewidth]{img/arrows/arrow_v2_slim.pdf}} \\
  
  & $\alpha=0.01$ & $\alpha=0.05$ & $\alpha=0.1$ & $\alpha=0.2$ & $\alpha=0.5$ & $\alpha=1$ & $\alpha=1000$ \\

   \rotatebox{90}{\quad FedAvg} &
   \includegraphics[trim=1.2cm 0.5cm 1.5cm 1.4cm, clip, width=\imgsizeee]{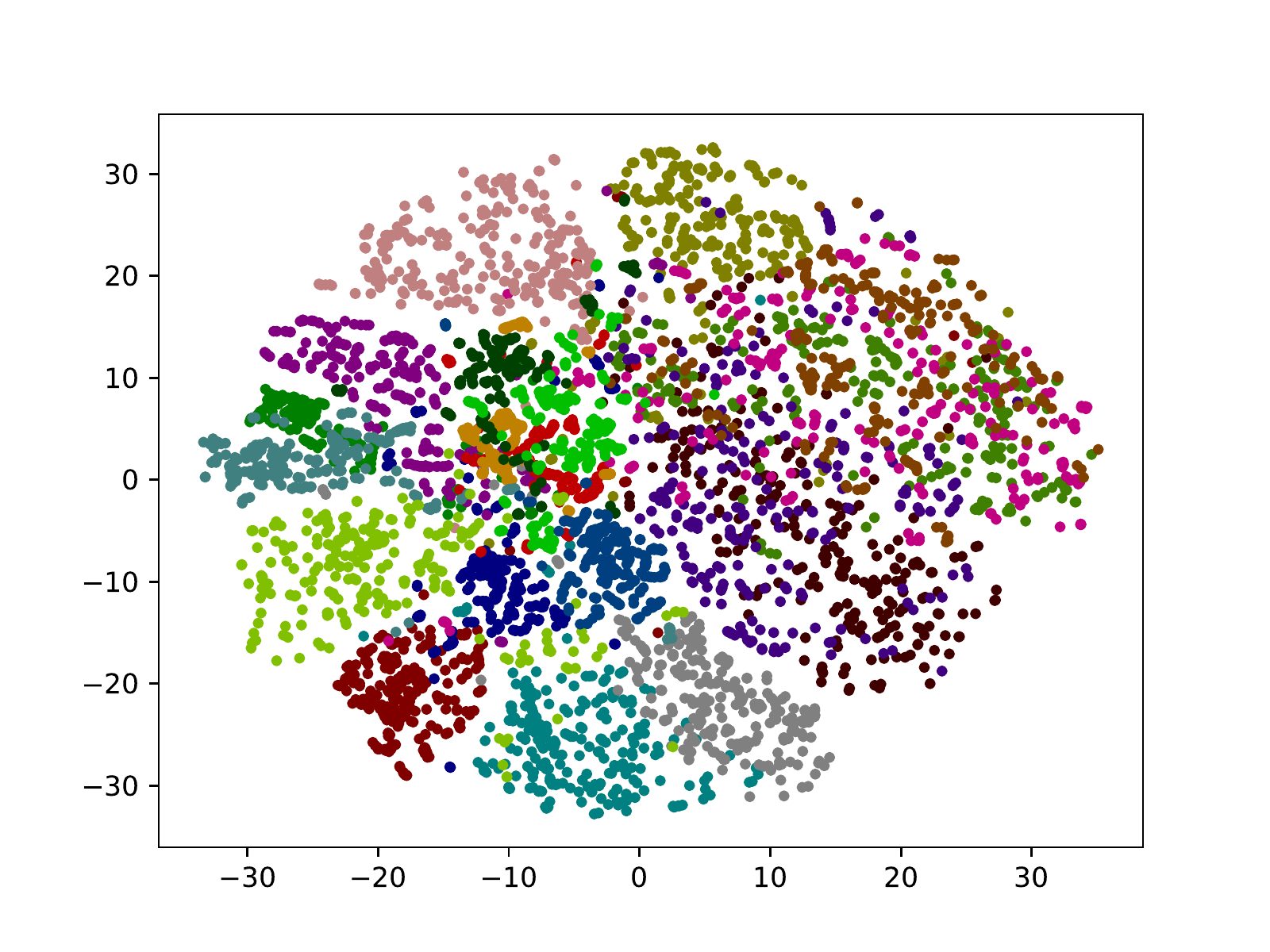} &
   \includegraphics[trim=1.2cm 0.5cm 1.5cm 1.4cm, clip, width=\imgsizeee]{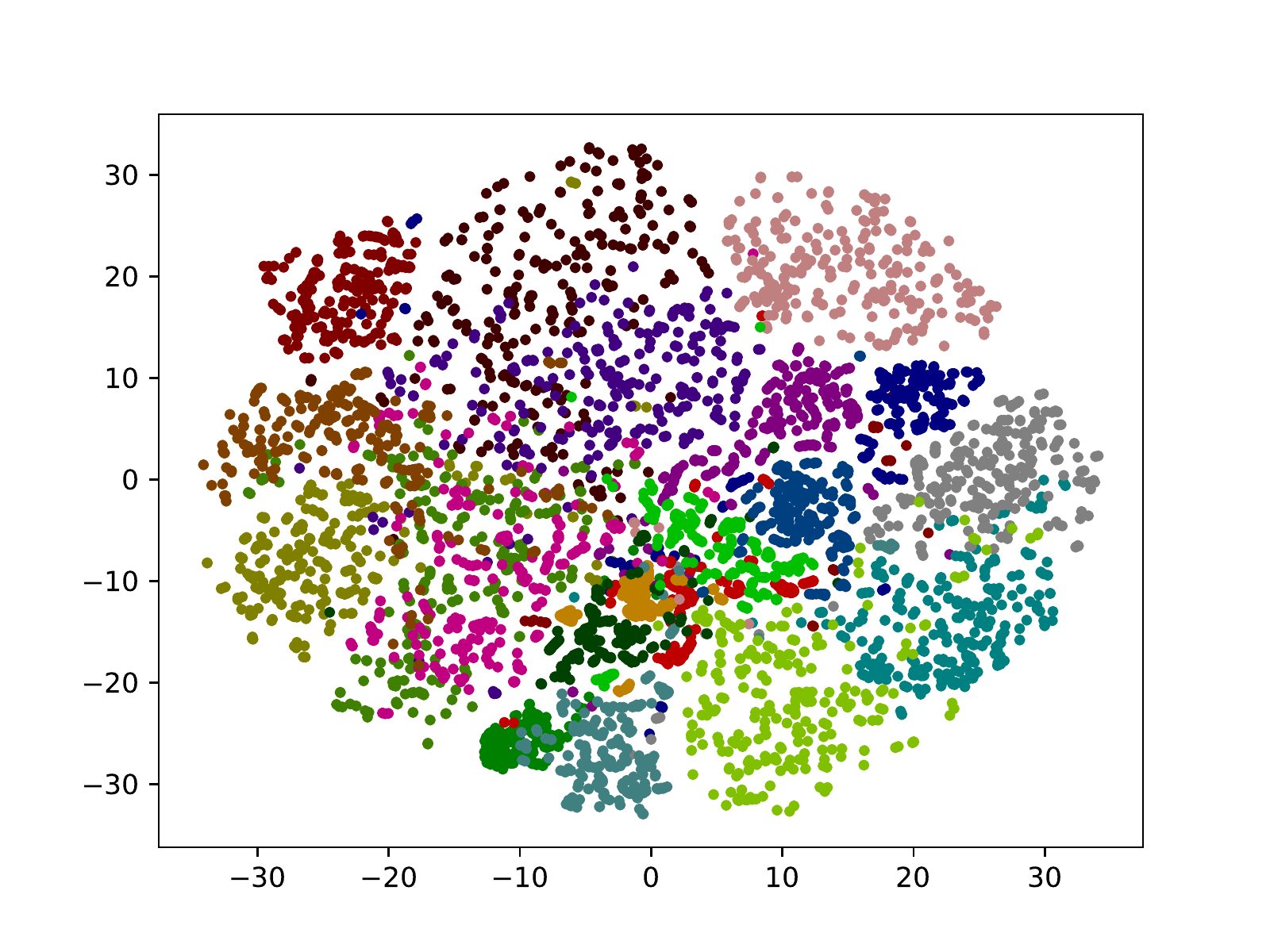} &
   \includegraphics[trim=1.2cm 0.5cm 1.5cm 1.4cm, clip, width=\imgsizeee]{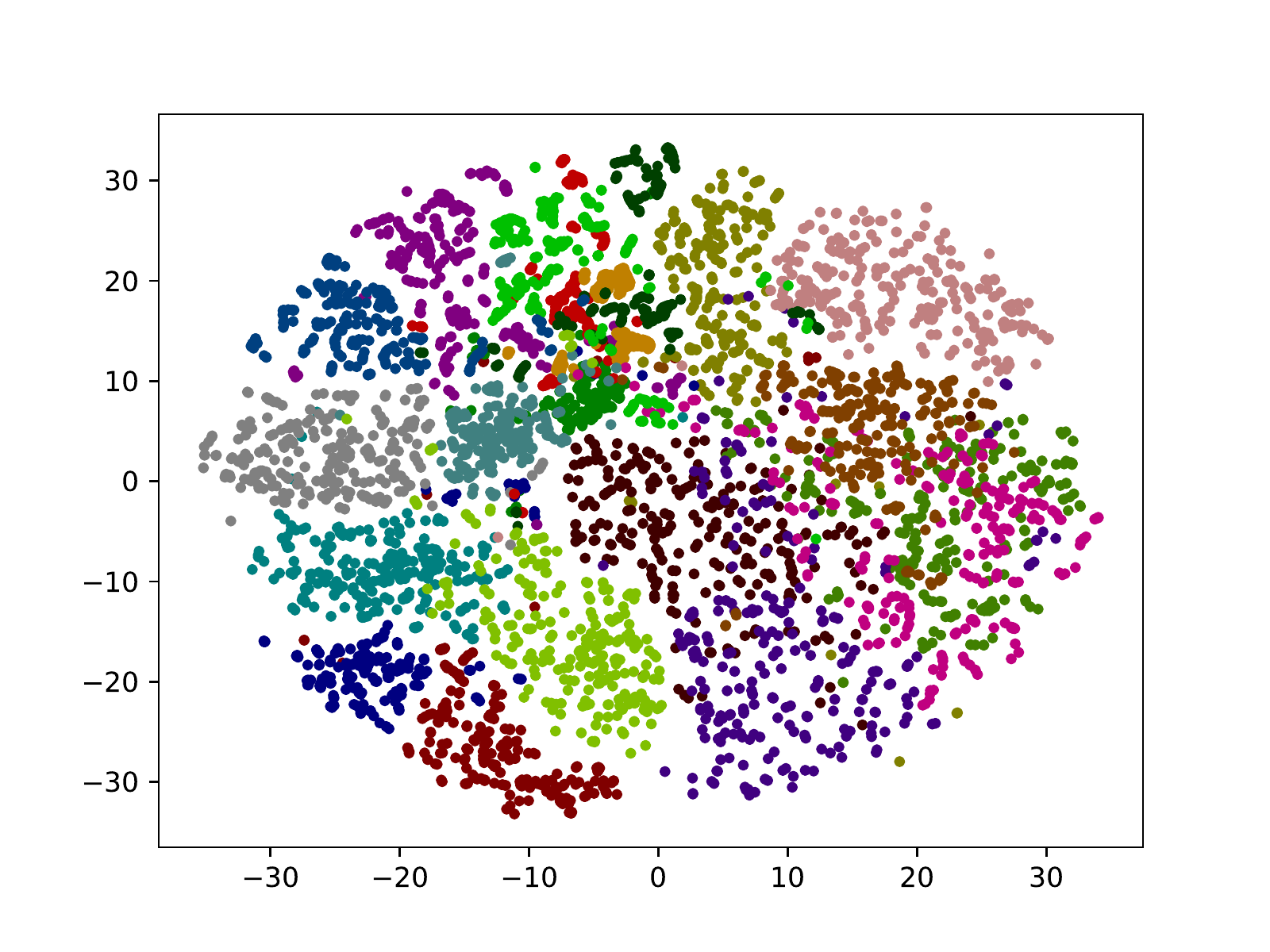} &
   \includegraphics[trim=1.2cm 0.5cm 1.5cm 1.4cm, clip, width=\imgsizeee]{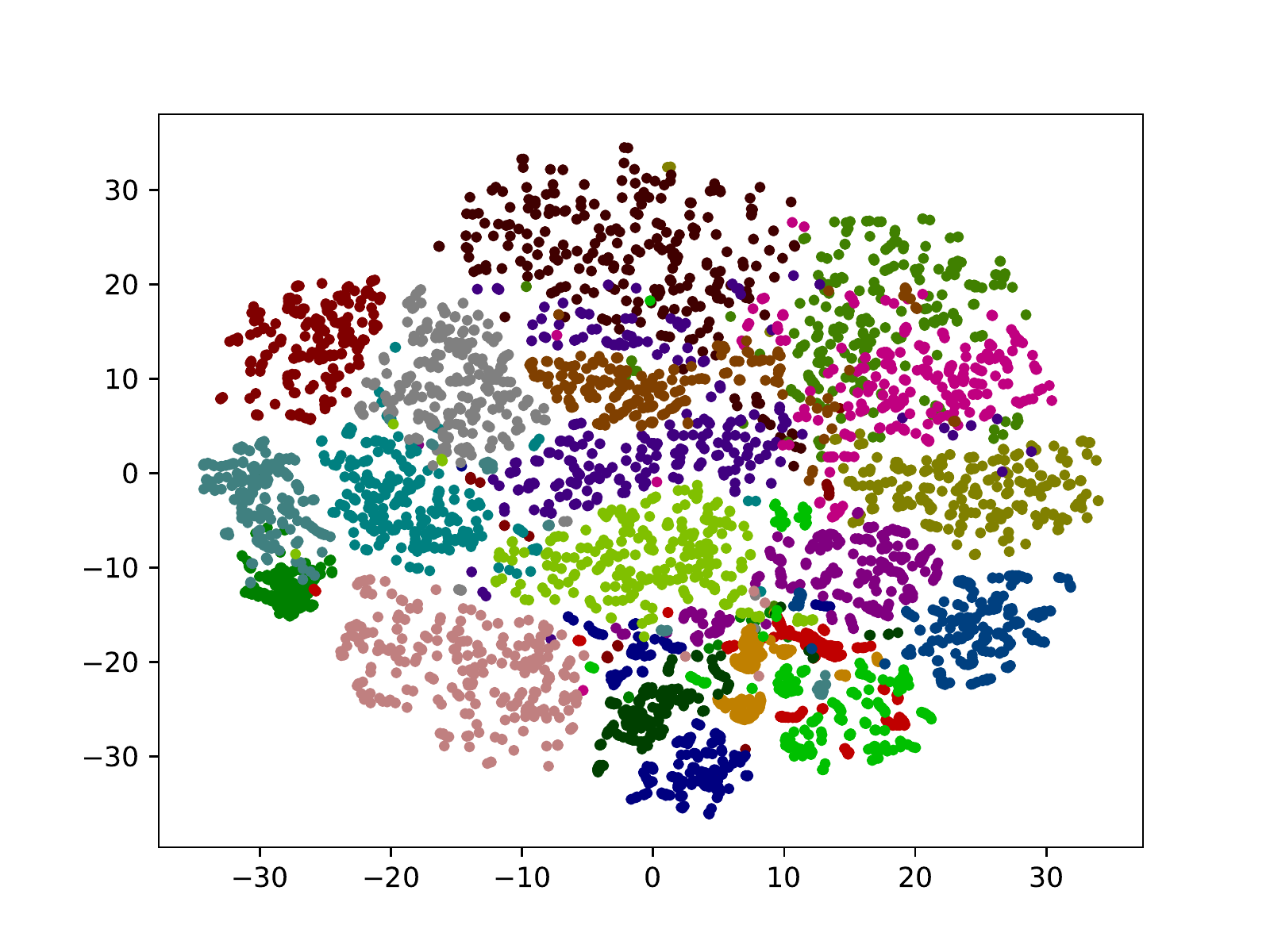} &
   \includegraphics[trim=1.2cm 0.5cm 1.5cm 1.4cm, clip, width=\imgsizeee]{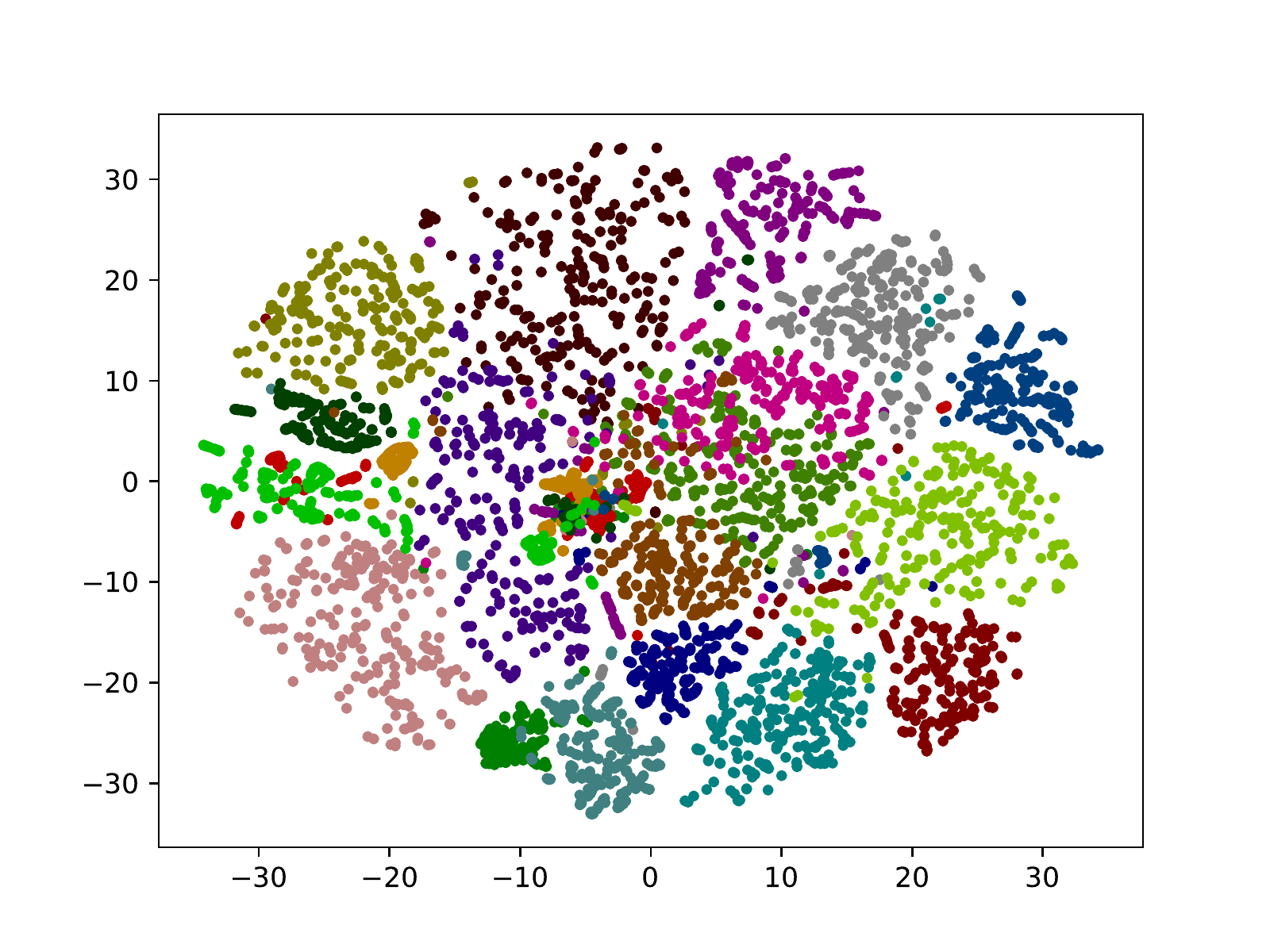} &
   \includegraphics[trim=1.2cm 0.5cm 1.5cm 1.4cm, clip, width=\imgsizeee]{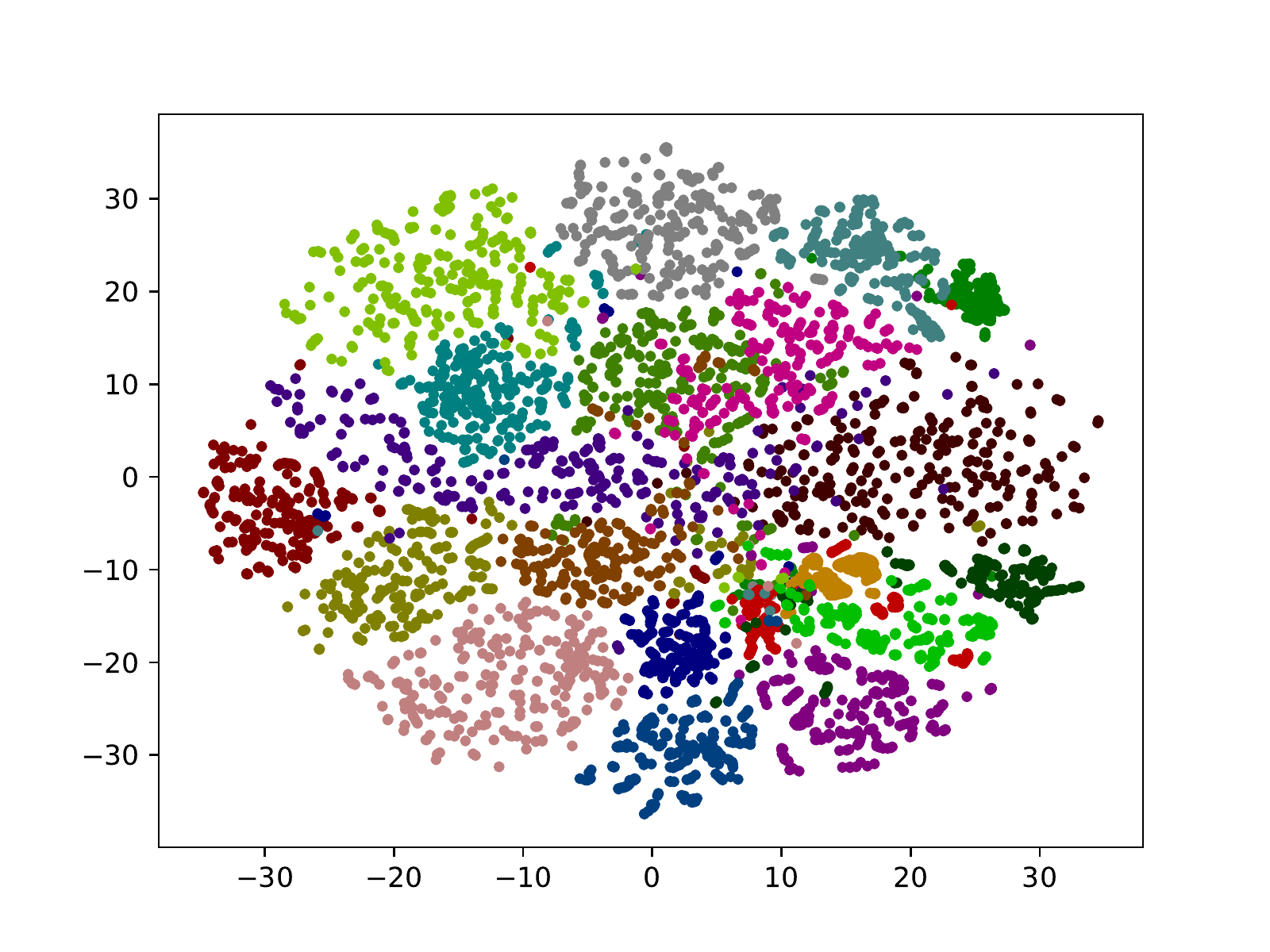} &
   \includegraphics[trim=1.2cm 0.5cm 1.5cm 1.4cm, clip, width=\imgsizeee]{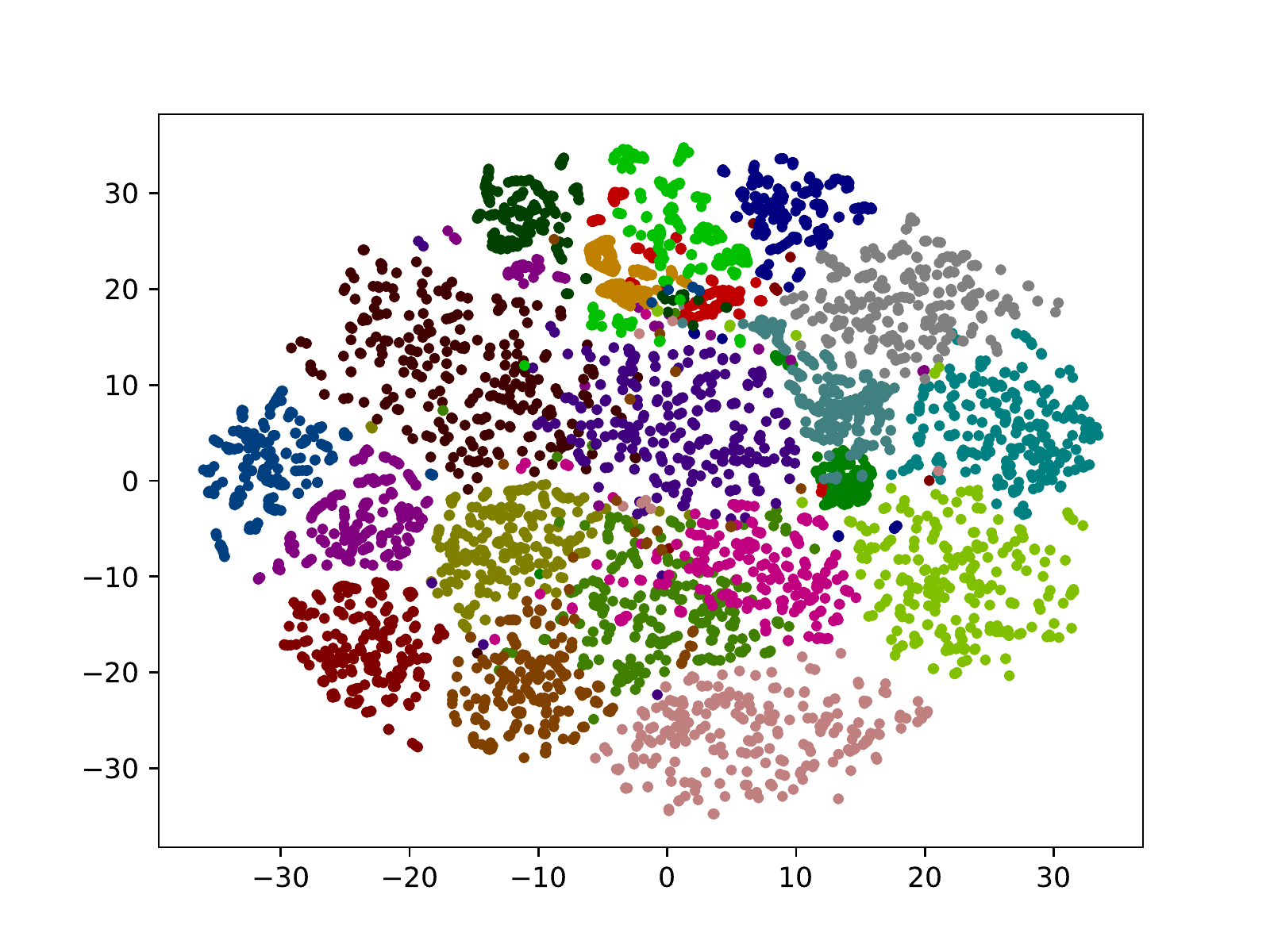}
   \\
   
      \rotatebox{90}{\quad FedProto} &
   \includegraphics[trim=1.2cm 0.5cm 1.5cm 1.4cm, clip, width=\imgsizeee]{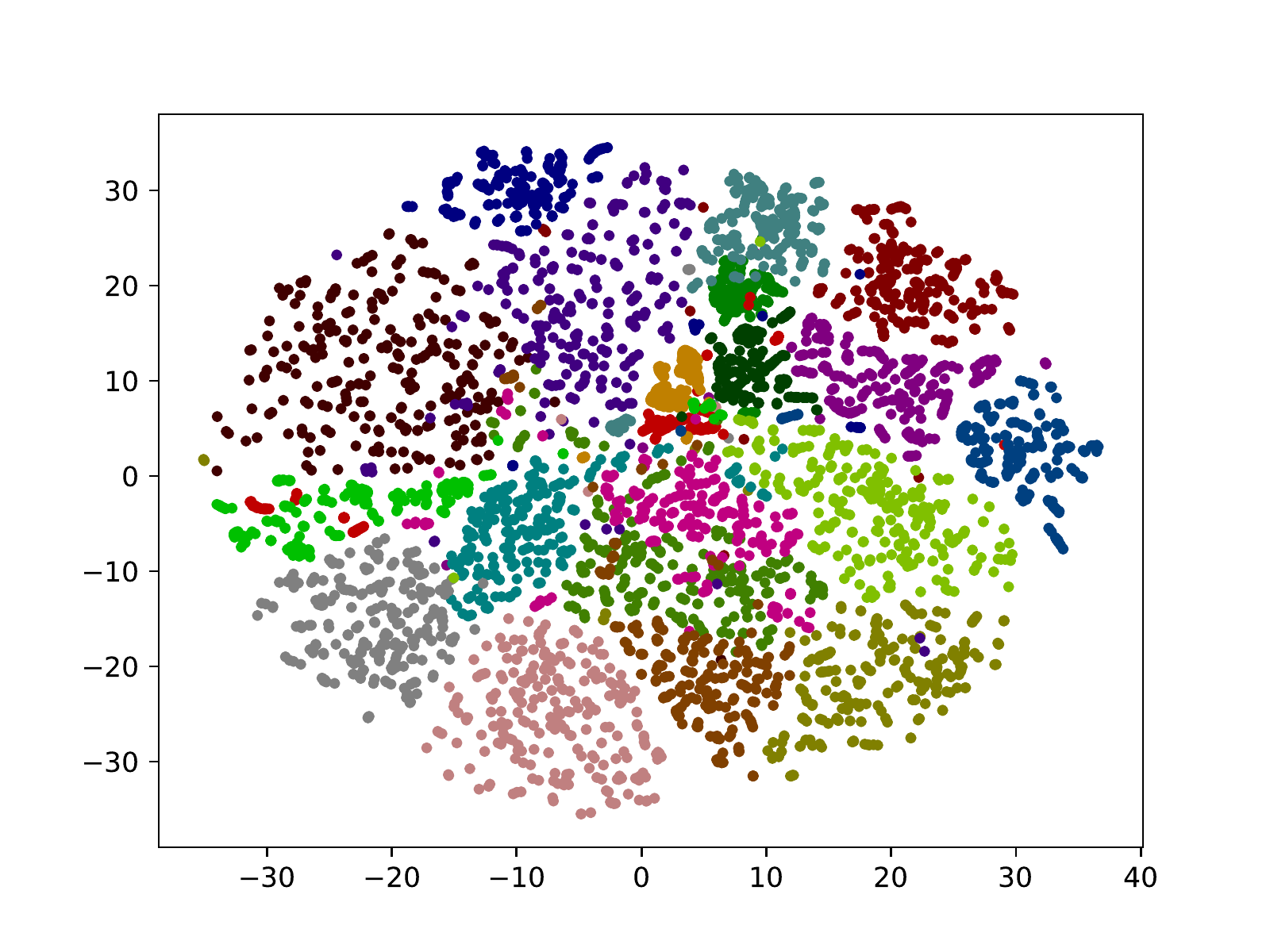} &
   \includegraphics[trim=1.2cm 0.5cm 1.5cm 1.4cm, clip, width=\imgsizeee]{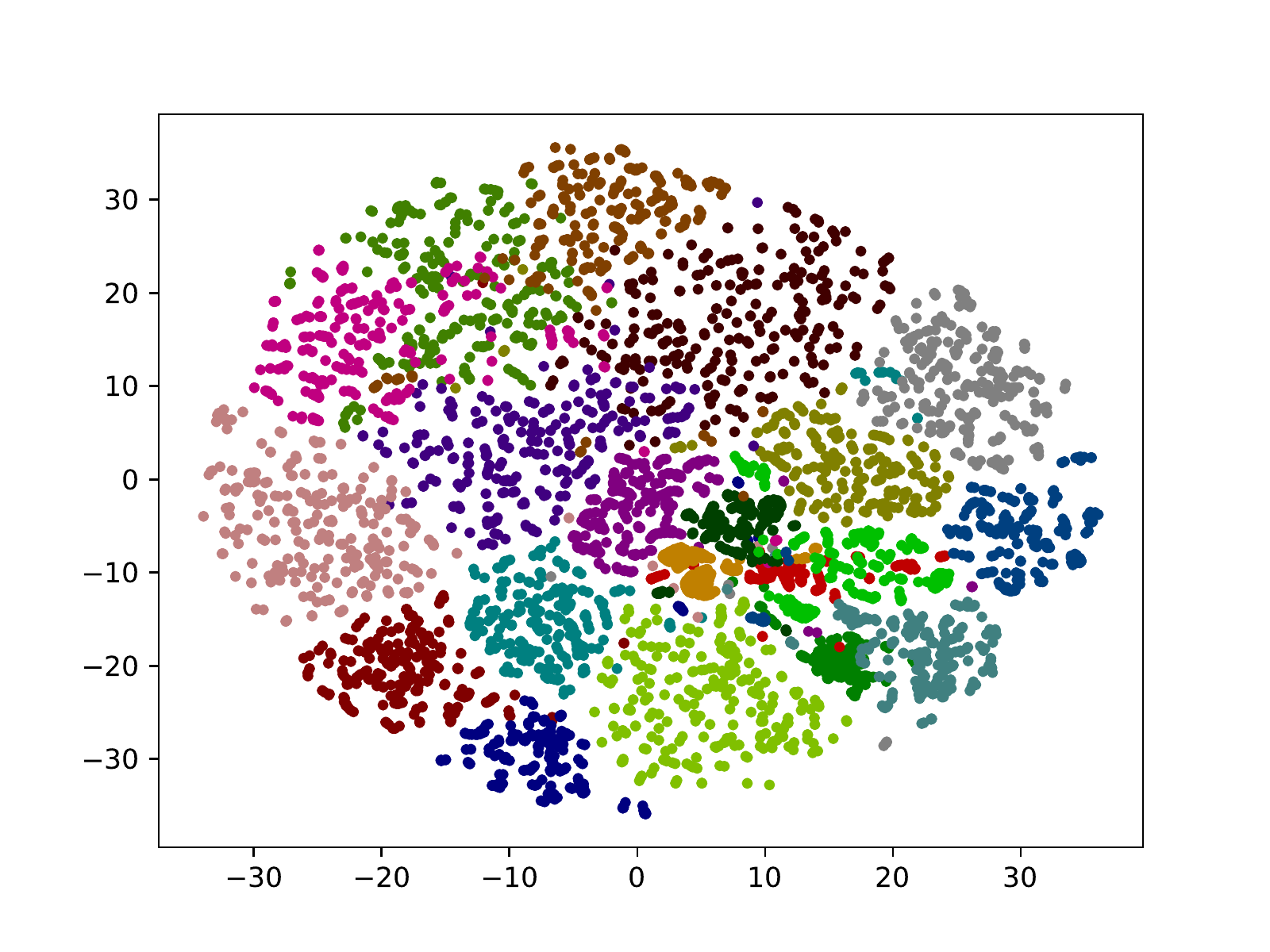} &
   \includegraphics[trim=1.2cm 0.5cm 1.5cm 1.4cm, clip, width=\imgsizeee]{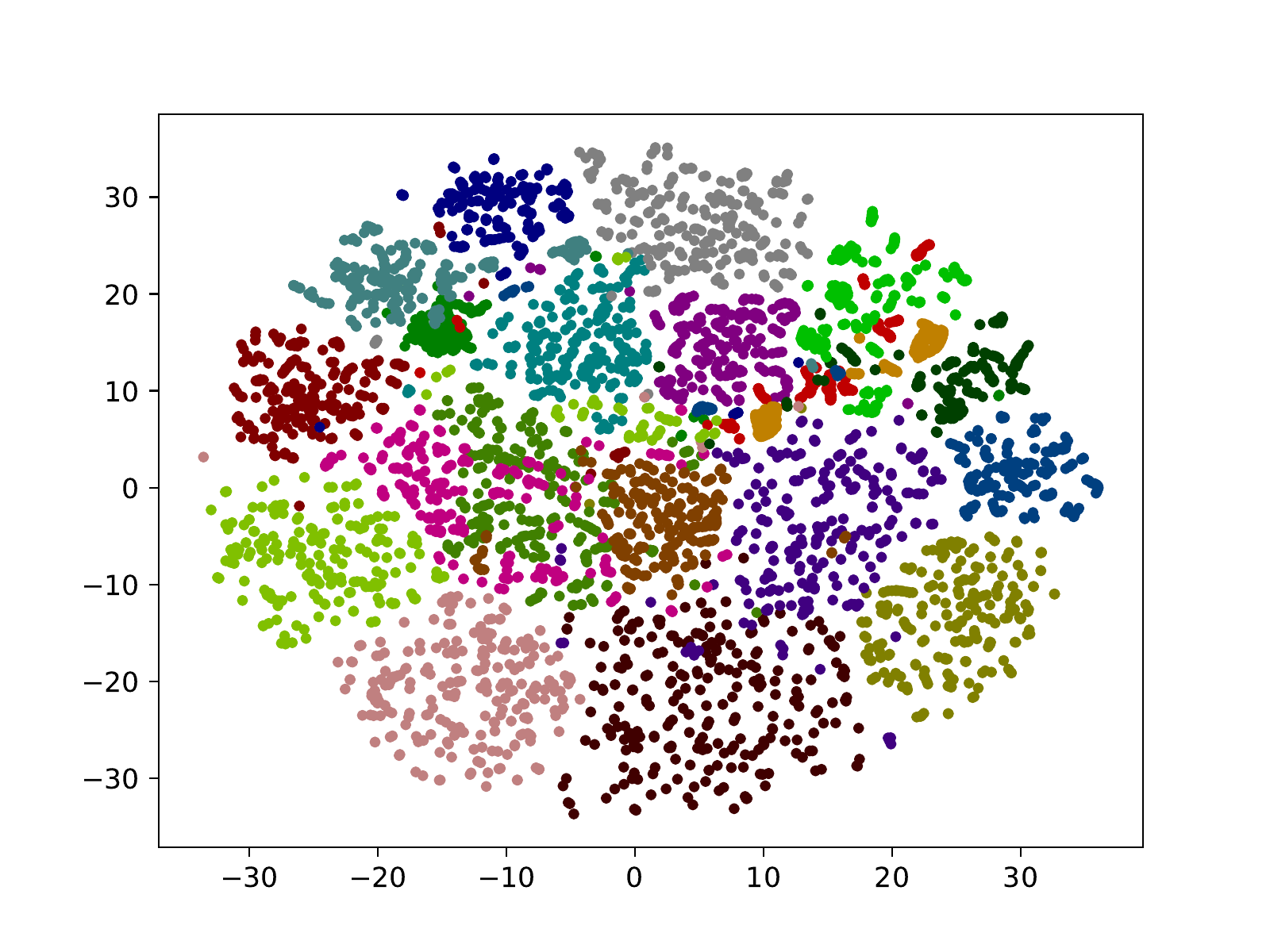} &
   \includegraphics[trim=1.2cm 0.5cm 1.5cm 1.4cm, clip, width=\imgsizeee]{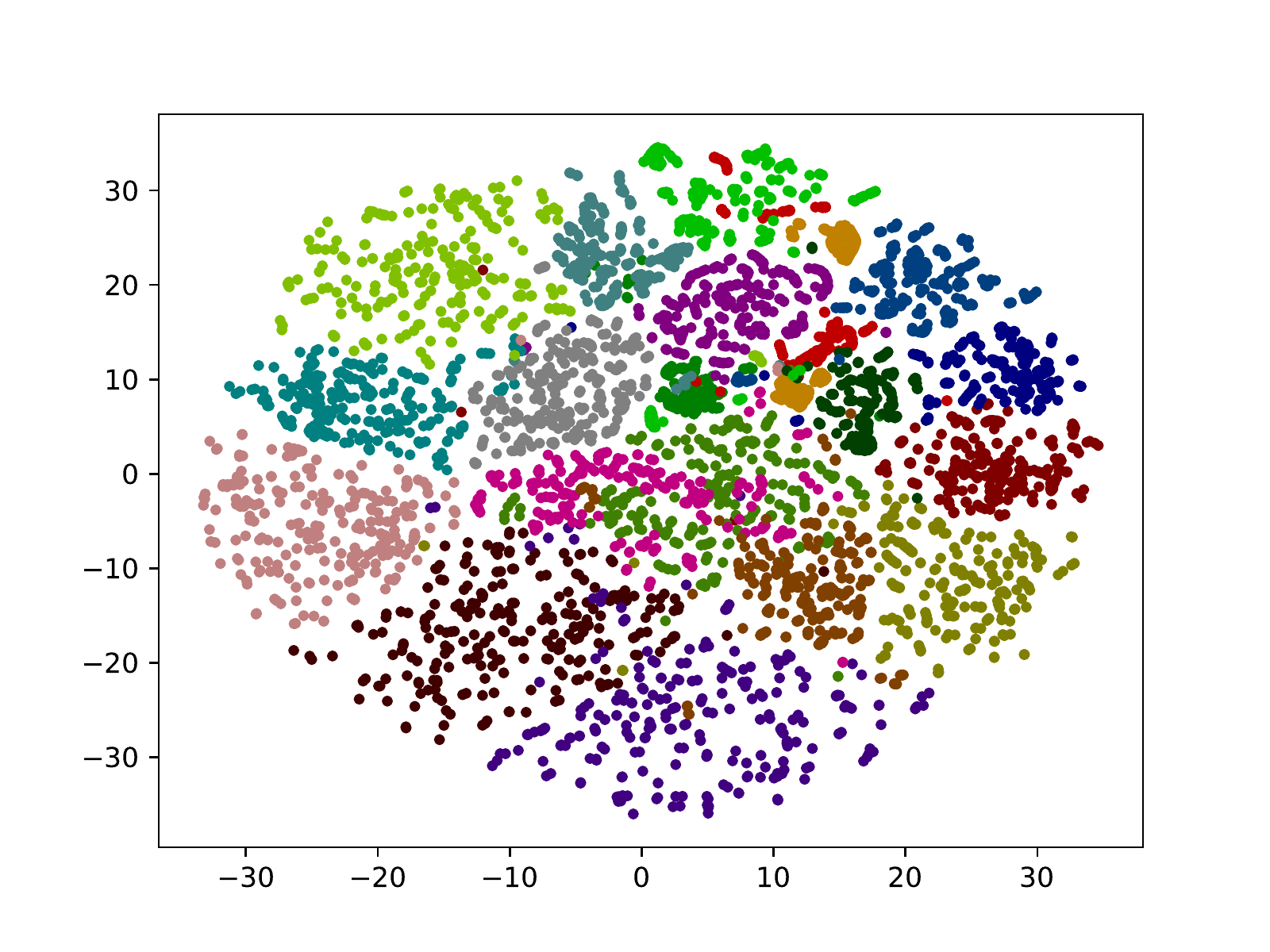} &
   \includegraphics[trim=1.2cm 0.5cm 1.5cm 1.4cm, clip, width=\imgsizeee]{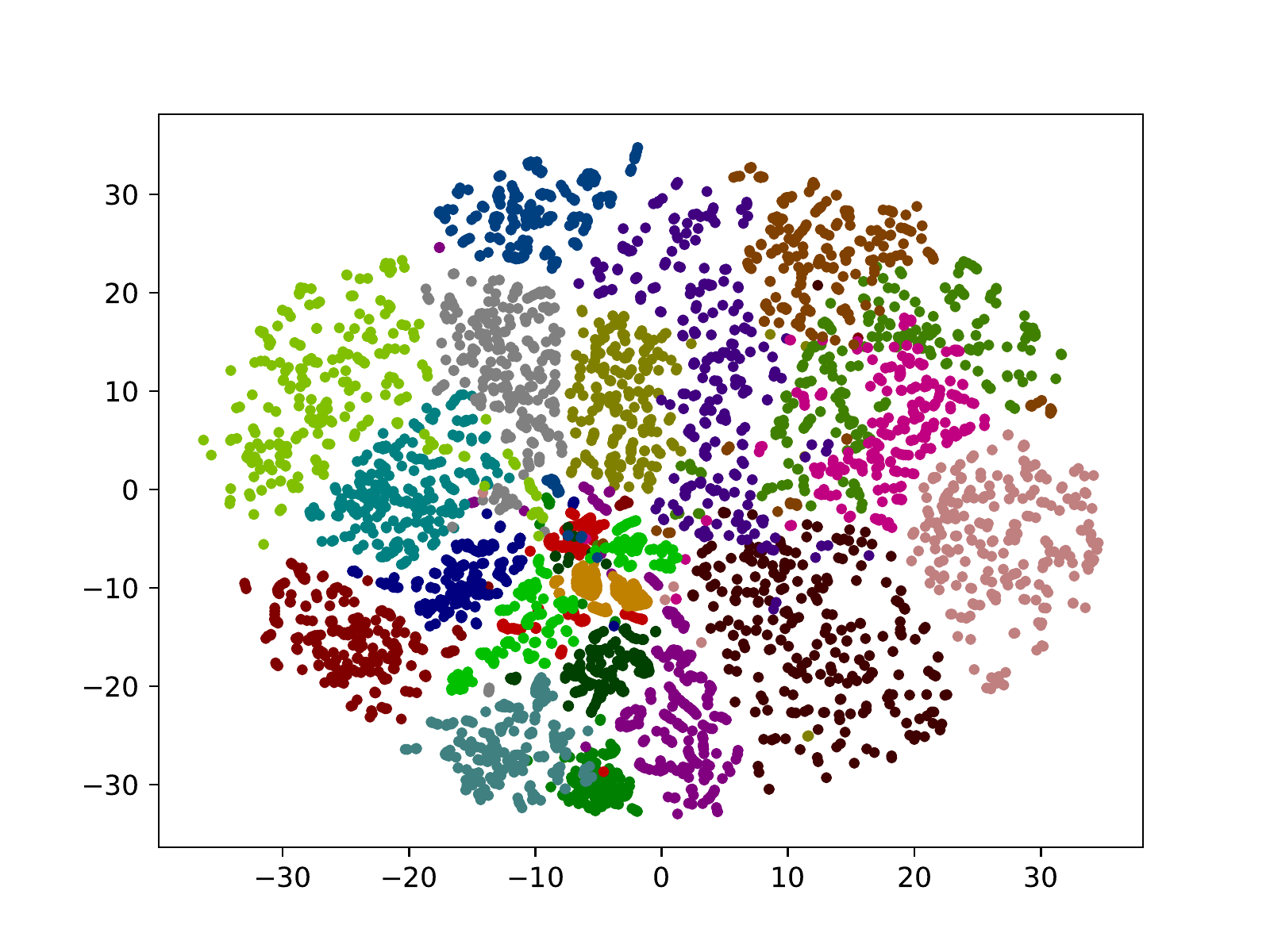} &
   \includegraphics[trim=1.2cm 0.5cm 1.5cm 1.4cm, clip, width=\imgsizeee]{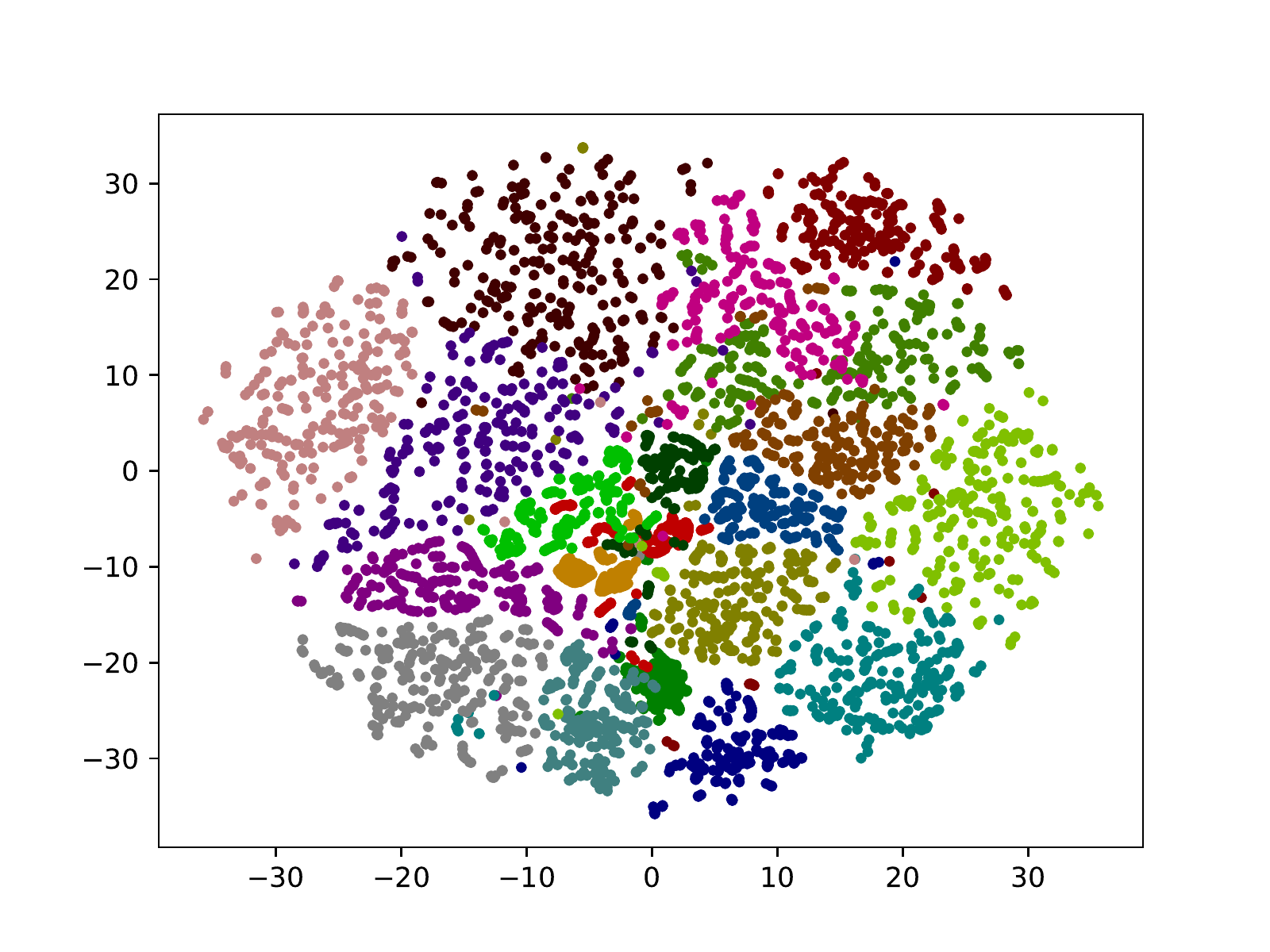} &
   \includegraphics[trim=1.2cm 0.5cm 1.5cm 1.4cm, clip, width=\imgsizeee]{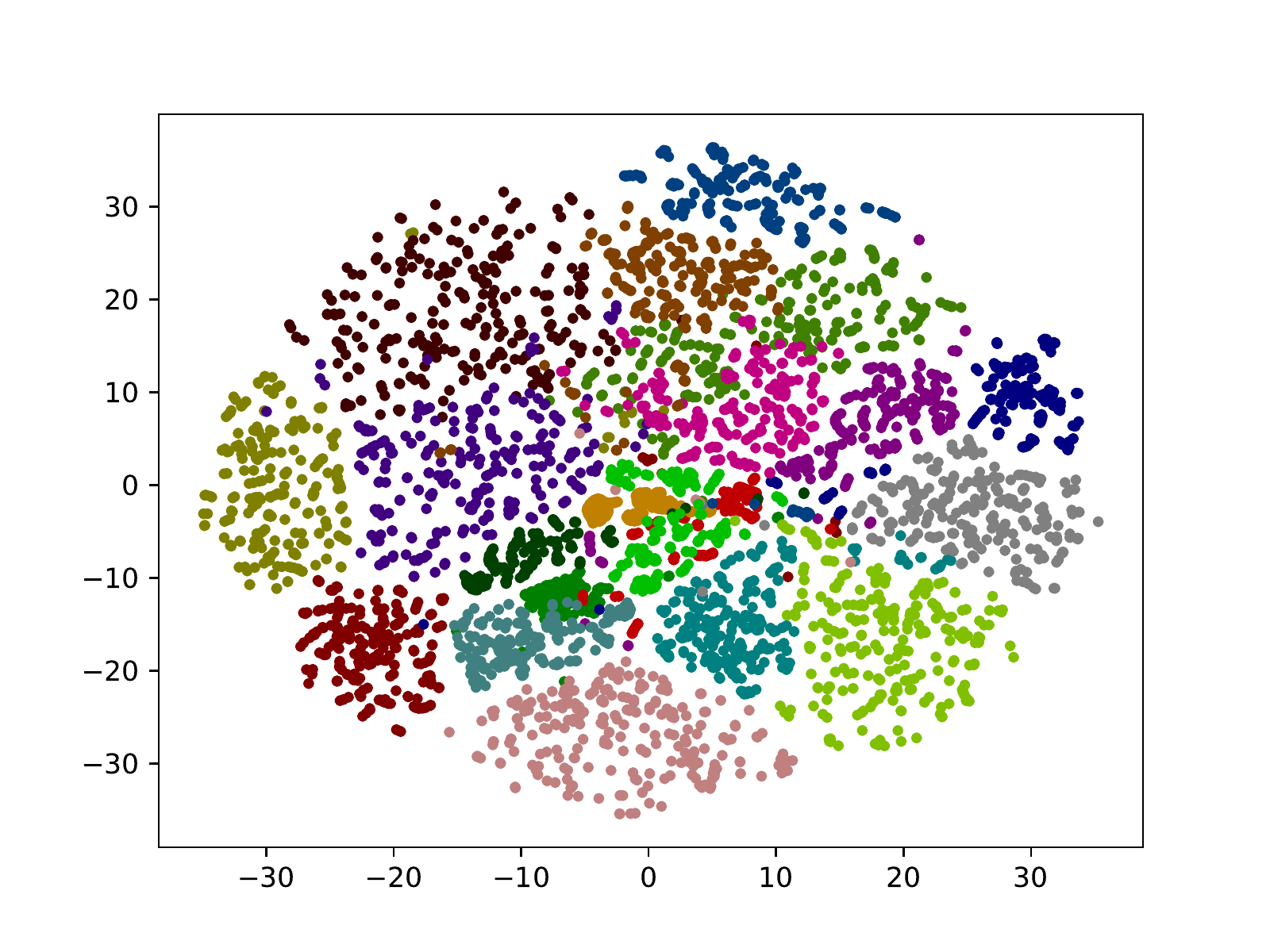}
   \\

\end{tabular}
\vspace{0.05cm}
\caption{Comparison of t-SNE embedding plots of feature representations learned  by FedAvg and by our FedProto, using the Pascal VOC 2012 segmentation benchmark with $20$ object level classes. Analyses are performed over different values of $\alpha$. The \textit{background} class is not included in the visualization, and the colors refer to the Pascal VOC2012 colormap. \textit{Best viewed in colors.}}
\label{suppl:fig:pascal_tsne}
\end{figure*}

\end{document}